\documentclass{article}
\usepackage[nonatbib, preprint]{neurips_2026}
\usepackage[utf8]{inputenc}
\usepackage[T1]{fontenc}

\usepackage{graphicx}
\usepackage{amssymb}
\usepackage{amsmath}
\usepackage{xcolor}

\usepackage{booktabs}
\usepackage{longtable}
\usepackage{tabularx}
\usepackage{array}
\usepackage{multirow}
\usepackage{enumitem}
\usepackage{float}
\usepackage{pdflscape}
\usepackage{placeins}
\usepackage{caption}   

\usepackage{microtype}

\usepackage{siunitx}
\DeclareSIUnit{\angstrom}{\textup{\AA}}
\usepackage[version=4]{mhchem}

\usepackage{algorithm}
\usepackage{algpseudocode}

\usepackage{listings}

\definecolor{lstbg}{gray}{0.93}
\DeclareUnicodeCharacter{2009}{\,}

\makeatletter
\@ifundefined{NBnumericcites}
  {\usepackage[style=authoryear,maxcitenames=2,maxbibnames=30,natbib=true,eprint=false]{biblatex}}
  {\usepackage[style=numeric-comp,sorting=none,maxbibnames=30,natbib=true,eprint=false]{biblatex}}
\makeatother
\AtEveryBibitem{\clearfield{issn}}
\renewbibmacro{in:}{%
  \ifentrytype{article}
    {\iffieldundef{journaltitle}
       {}
       {\printtext{\bibstring{in}\intitlepunct}}}
    {\printtext{\bibstring{in}\intitlepunct}}}

\usepackage{hyperref}
\usepackage{xparse}

\definecolor{clrprompt}{HTML}{8B0000}
\definecolor{clrthink}{HTML}{1a5276}
\definecolor{clrreason}{HTML}{6C3483}
\definecolor{clranswer}{HTML}{117A65}
\newcommand{\tprompt}[1]{{\small\ttfamily\bfseries\textcolor{clrprompt}{#1}}}

\definecolor{clrlabel}{HTML}{B45F06}

\definecolor{citegray}{gray}{0.25} 
\newcommand{\mycitestyle}[1]{\textcolor{citegray}{\small #1}}
\NewCommandCopy{\oldcite}{\cite}
\RenewDocumentCommand{\cite}{o o m}{%
  \mycitestyle{%
    \IfNoValueTF{#2}{\oldcite[#1]{#3}}{\oldcite[#1][#2]{#3}}%
  }%
}

\NewCommandCopy{\oldcitep}{\citep}
\RenewDocumentCommand{\citep}{o o m}{%
  \mycitestyle{%
    \IfNoValueTF{#1}{%
      \oldcitep{#3}
    }{%
      \IfNoValueTF{#2}{%
        \oldcitep[#1]{#3}
      }{%
        \oldcitep[#1][#2]{#3}
      }%
    }%
  }%
}

\NewCommandCopy{\oldcitet}{\citet}
\RenewDocumentCommand{\citet}{o o m}{%
  \mycitestyle{%
    \IfNoValueTF{#1}{%
      \oldcitet{#3}%
    }{%
      \IfNoValueTF{#2}{%
        \oldcitet[#1]{#3}%
      }{%
        \oldcitet[#1][#2]{#3}%
      }%
    }%
  }%
}

\NewCommandCopy{\oldcitealp}{\citealp}
\DeclareRobustCommand{\citealp}[2][]{%
  \mycitestyle{\oldcitealp[#1]{#2}}%
}   

\newsavebox{\mycaptionbox}  

\newif\iflandscapewide
\landscapewidefalse

\title{Frontier LLMs are effective batch optimizers: Assessing reasoning models in continuous and discrete settings}

\newcommand{\acktext}{We thank Derek Ma and the Converge computing team for help with the computing infrastructure necessary for conducting this work.}

\landscapewidetrue   

\author{%
  Frank Hu \quad Shriram Chennakesavalu \quad David Graff \\
  Prescient Design, Genentech \\
  South San Francisco, CA, USA \\ 
  \texttt{graff.david@gene.com}
}

\begin{document}

\maketitle
\begin{abstract}
Frontier large language models (LLMs) have become attractive priors for optimization due to their large-scale pretraining that enables them to navigate a variety of optimization settings. However, the effectiveness of modern reasoning LLMs in batch optimization settings remains underexplored.
Here we investigate the performance of the current generation of frontier LLMs as batch optimizers in both continuous and discrete settings. We find that while LLMs are competitive zero-shot batch optimizers for numerical test functions, their performance is brittle compared to classical non-LLM optimization approaches. However, LLM priors are significantly better in semantically rich settings, indicating that their batch optimization behavior is highly effective when navigating and reasoning over the discrete spaces most similar in structure to their pretraining data. 

\end{abstract}

\section{Introduction}
\label{sec:introduction}
Frontier large language models (LLMs) have become increasingly powerful with each new generation due to the continued development of pretraining corpora and reinforcement learning protocols, instilling these models with exceptional reasoning capabilities that enable them to tackle lengthier and more complex tasks. This has made them attractive for a variety of applications, most notably in programming~\citep{research_composer_2026}, mathematics~\citep{shao_deepseekmath_2024}, and more recently, scientific discovery~\citep{chennakesavalu_evaluating_2026,narayanan_training_2025}. 

While this continued evolution in LLM capabilities is remarkable, it remains unclear how effective these models have become for black-box optimization (BBO)~\citep{alarie_two_2021,audet_derivative-free_2026} settings where the evaluation of a target objective function is expensive, rendering brute force optimization intractable. This is a ubiquitous scenario in scientific settings like small-molecule drug discovery where candidate evaluations for a proposed small-molecule design require a high investment of time and resources, meaning that the evaluation budget is highly restricted against a combinatorially complex design space~\citep{ertl_cheminformatics_2003,ertl_estimation_2009}. Much work has been done building specialist methods for BBO problems in both continuous and discrete settings and some work has been done investigating earlier iterations of LLMs on similar classes of problems~\citep{song_reinforced_2024,meindlZeroShotOptZeroShotPretrained2025,liuLargeLanguageModels2024,yangLargeLanguageModels2024}, but it remains unclear how well the current generation of reasoning LLMs perform on batch optimization for both continuous and discrete settings. Gauging these capabilities would help better characterize the optimization behavior of policies based on these models while underscoring failure modes and directions for improvement.

In this work we investigate the performance of the Anthropic family of models (Sonnet 4.6~\citep{anthropic_introducing_2026}, Opus 4.8~\citep{_introducing_op48}, and Opus 5~\citep{_introducing_a}) as batch optimizers for a collection of continuous and discrete BBO tasks. For evaluation over a continuous search space, we use a set of well-known optimization functions of varying dimensionality, taking care to disguise their domains and ranges to minimize memorization exploitation from the model. For the discrete setting, we choose molecular optimization against the set of oracles detailed in the practical molecular optimization (PMO)~\citep{gaoSampleEfficiencyMatters2022} benchmark. Molecular optimization lends itself easily to natural-language reasoning via the usage of SMILES~\citep{weininger_smiles_1988} strings as the representation, and the deterministic oracles enable straightforward verification.

Overall, we find that frontier reasoning LLMs can be effective but brittle numerical optimizers, with performance that rivals classical Bayesian Optimization with a Gaussian process surrogate and an expected improvement acquisition function, but the general performance depends heavily on task transformations, dimensionality, and batch size. However, we show that in the discrete molecular optimization setting, the frontier reasoning LLMs are both highly performant and sample efficient, frequently beating specialist methods with only a fraction of the oracle sampling budget. 

\section{Related work}
\label{sec:related-work}

{\noindent
\textbf{Bayesian Optimization}
Bayesian Optimization (BO) is a global optimization method for black-box optimization \citep{frazierTutorialBayesianOptimization2018}. It is an active learning method that employs a probabilistic surrogate model, 
$\hat f$, over the input domain $\mathcal H$ to characterize the posterior predictive distribution of the objective, together with an acquisition function, $\alpha(h; \hat f, \mathcal D)$, that quantifies the utility of evaluating an input $h \in \mathcal H$ conditioned on a fitted surrogate model $\hat f$ and measured dataset $\mathcal D$ to guide the selection of the next point. Gaussian processes (GPs) \citep{Rasmussen2006Gaussian} are a typical choice of surrogate model given the intuitive prior defined via the kernel and analytically tractable posterior inference \citep{snoekPracticalBayesianOptimization2012}. Common acquisition functions include both upper confidence bound (UCB) and expected improvement (EI), which balance exploration of the landscape and exploitation of promising regions. BO is a popular choice to optimize black-box functions across many domains, such as materials science~\citep{chitturi_targeted_2024}, chemistry~\citep{nambiar_bayesian_2022}, biology~\citep{li_biobo_2025}, and robotics~\citep{martinez-cantin_funneled_2019}.

{\noindent
\textbf{Pretrained models for optimization}
Prior work has explored the application of pretrained models to optimization tasks. Broadly, these works can be classified into two regimes: (1) pretraining transformer models to mimic tokenized optimization trajectories on synthetic tasks \citep{song_reinforced_2024,meindlZeroShotOptZeroShotPretrained2025} and (2) leveraging pretrained LLMs to enhance individual components of the BO loop~\citep{liuLargeLanguageModels2024,ramosBayesianOptimizationCatalysis2025}. Most similar to our work are \citet{yangLargeLanguageModels2024} and \citet{huangExploringTruePotential2024}, both of which utilize a prompt-based approach as a stand-in for a full optimization policy. But these methods do not explore batched LLM optimization policies applied to standard BO test functions nor do they apply these approaches to molecular optimization. Most importantly, LLM capabilities have substantially improved since these prior studies were conducted. Hence, it is important to revisit these methodologies using significantly more capable frontier models.
}

{\noindent
\textbf{Molecular optimization}
Optimizing molecular properties is a key challenge in the development of novel functional materials, such as pharmaceuticals, photovoltaics, and batteries. Broadly, there are two strategies to improve desired properties: (1) forward search in a predefined library, where molecules are \emph{screened} according to the property itself (or some surrogate thereof) and (2) inverse design, where molecules are \emph{generated} based on a specification of the desired properties \citep{sanchez-lengelingInverseMolecularDesign2018}. LLMs have been used in a variety of ways for molecular optimization: as generators of property-optimized molecules \citep{navratilGPMoLFormerSimTestTime2025,bagalMolGPTMolecularGeneration2021,kopfSampleEfficientGenerative2026,rankovicLargeLanguageModels2026}, molecular representation learners for use in property prediction \citep{freyNeuralScalingDeep2023,ahmadChemBERTa2ChemicalFoundation2022}, property predictors themselves, and augmentations to individual components of closed-loop optimization protocols \citep{ramosBayesianOptimizationCatalysis2025,ranExLLMExperienceEnhancedLLM2025,wangEfficientEvolutionarySearch2025}}.

\section{Background}
\label{sec:background}
\subsection{Black-box optimization}\label{subsec:bbo}

We consider the setting of black-box optimization, where we are provided with an objective function $f:\mathcal{H} \rightarrow \mathbb{R}$ and $\mathcal{H}$ is the search space that can take on different forms, e.g., $\mathcal{H}\subset \mathbb{R}^d$ for a $d$-dimensional search space or $\mathcal{H}$ can be a discrete space as in the case of language modeling. We seek to maximize the objective function by finding the optimal $h^*\in\mathcal{H}$ where $h^*=\arg \max_{h\in\mathcal{H}} f(h)$. In cases where the optimal value is known \textit{a priori} for a particular problem, we can measure how close a proposed point is to the optimum via the instantaneous regret, defined as $r_i = f(h^\ast) - f(h_i)$. We focus here on optimizing $f$ in an iterative fashion, where the specific optimization policy emits a trajectory of points across a budget of $N/q$ turns, where $N$ is the total budget for the objective function and at each turn the policy can suggest either a single $h \in \mathcal{H}$ or a batch $\{h_i\}_{i=1}^q\subset\mathcal{H}$ to assess using the objective function. In this setting, we can also consider the simple regret over the trajectory $r = \min_i r_i$, i.e., the minimum of the instantaneous regret. Lower regret is better.

\subsection{LLMs for black-box optimization}

For this work we restrict our scope to the family of frontier autoregressive LLMs from Anthropic, comparing the evolution in performance across Sonnet 4.6, Opus 4.8, and Opus 5. An autoregressive LLM can be treated as a policy $\pi$ from which we can sample responses $y\in\mathcal{Y}$ provided prompts $x\in\mathcal{X}$, with $\mathcal{X}$ and $\mathcal{Y}$ both being discrete spaces. For the case of BBO, the prompt $x\in\mathcal{X}$ consists of information relevant to the optimization problem, such as previously selected points and their values under the objective function and the remaining turn budget. The responses that we sample $y \sim \pi(\cdot \mid x)$ are naturally structured as a trajectory of individual responses, where each response consists of a single proposed point for evaluation or a batch of points. We also assess the effect of using the LLM in single- versus multi-turn settings: in a single-turn setting the prompt $x_i$ at turn $i$ consists of the previously selected points and their values under the objective function $f$ and the response is sampled as $y_i\sim\pi(\cdot|x_i)$ whereas in a multi-turn setting, the model maintains one conversation, enabling reflection on previous reasoning that is emitted during the optimization process. While previous work has examined the efficacy of LLMs in different optimization settings~\citep{liuLargeLanguageModels2024,yangLargeLanguageModels2024,ramosBayesianOptimizationCatalysis2025}, we are interested in the effectiveness of current frontier reasoning LLMs in these optimization settings.

\section{Experimental design}
\subsection{Synthetic optimization tasks}
We first examine the performance of the frontier LLMs on a set of synthetic optimization tasks using classical test functions with known optima: Branin, Hartmann-3, Hartmann-6, Ackley, and Rastrigin. For a summary of these synthetic test functions and their optima, refer to Appendix~\ref{app:synthetic-functions}. For each of these functions, we normalize the domain to $[0,1]^d$ and perform any necessary negations such that every task can be framed as a maximization. We then evaluate each LLM in both single- and multi-turn settings with batch sizes $q=1,2,4$ where the model emits a collection of points $\{h_i\}_{i=1}^q\subset[0,1]^d$ at each trajectory step. To ensure that the overall optimization budget is matched, we fix a total run budget of $N=120$ with $2d$ initial points based on the dimensionality of the problem, and where the length of each trajectory for each batch size is computed as $N/q$. In addition to testing on the original version of these functions, we also test on a disguised variation of these test functions by applying a bijection on the domain $\mathcal{T}:[0,1]^d \rightarrow [0,1]^d$. Specifically, the bijection is defined as a composition of three per-axis transformations:
\begin{equation}
    \mathcal{T} = \mathcal{P} \circ \mathcal{F} \circ \mathcal{W},
\end{equation}
where $\mathcal{P}$ is a permutation over the axes of $h$: $\mathcal P(h) = \sigma (h)$; $\mathcal{F}$ is an independent Bernoulli flip over each axis of $h$: $\mathcal F(h_i) = (1-B)h_i + B(1 - h_i)$, where $B \sim \operatorname{Bern}(0.5)$; and $\mathcal W$ is a power warp transformation for each axis of $h$: $\mathcal W(h_i) = h_i^a$, where $a\sim\mathcal{U}(0.5, 2.0)$. This bijection on the domain of each function transforms the landscape of each test function and shifts the optimal point $h^*$ without changing the value of the objective function $f(h^*)$.
We further apply an affine transformation $\mathcal S$ to the output $y = f(h)$ of each function: $\mathcal S(y) = ay+b$, where $a \sim \operatorname{LogUniform}\left(\frac{1}{3},3\right)$ and $b\sim\mathcal{U}(-10,10)$. 

We compare the LLM-based optimization policies to both a BO policy with a GP surrogate and EI acquisition function (``GP-EI'') as well as a random baseline. Each method is evaluated with identical batch sizes and total sample budgets and we evaluate simple regret as a function of the number of turns and the lowest obtained regret based on the best proposed point across five random seeds. 


\subsection{Molecular optimization tasks}

For a semantically richer optimization task, we assess the LLMs on the suite of molecular optimization tasks described in PMO~\citep{gaoSampleEfficiencyMatters2022} which tests the ability of different methods for maximizing small-molecule oracles in a sample-efficient manner. Unlike the synthetic test functions, these optimization problems naturally admit a discrete, natural-language representation via SMILES~\citep{weininger_smiles_1988} strings, and there is no analytic optimum for these oracles. We choose this molecular optimization task as opposed to other natural-language benchmarks because of the availability of cheap and deterministic oracle functions, meaning there is no ambiguity with regard to assessing improvement on specific tasks or performance against specific oracles. 

We evaluate all three frontier LLMs on all 23 PMO oracles across five seeds, again testing both single-turn and multi-turn settings. We compare the performance of the frontier LLMs against all of the other methods included in the original PMO paper and some additional ones. Unlike the 10000 oracle call budget that was used in the original study, we restrict our LLM evaluations to a very low-budget setting consisting of 10 initial molecules followed by 20 batches of 10 for a total oracle budget of 210. For comparing across all the methods, we focus on two metrics for our evaluation: Top-10 AUC and the Top-10 mean, where the former is especially important for gauging a method's sample efficiency. Where applicable, we use early stopping in a manner consistent with the method's original publication.

\section{Results}

\subsection{Optimization on synthetic test functions}\label{sec:synth_tasks}

\begin{figure}[htbp]
    \centering
    \includegraphics[width=\textwidth]{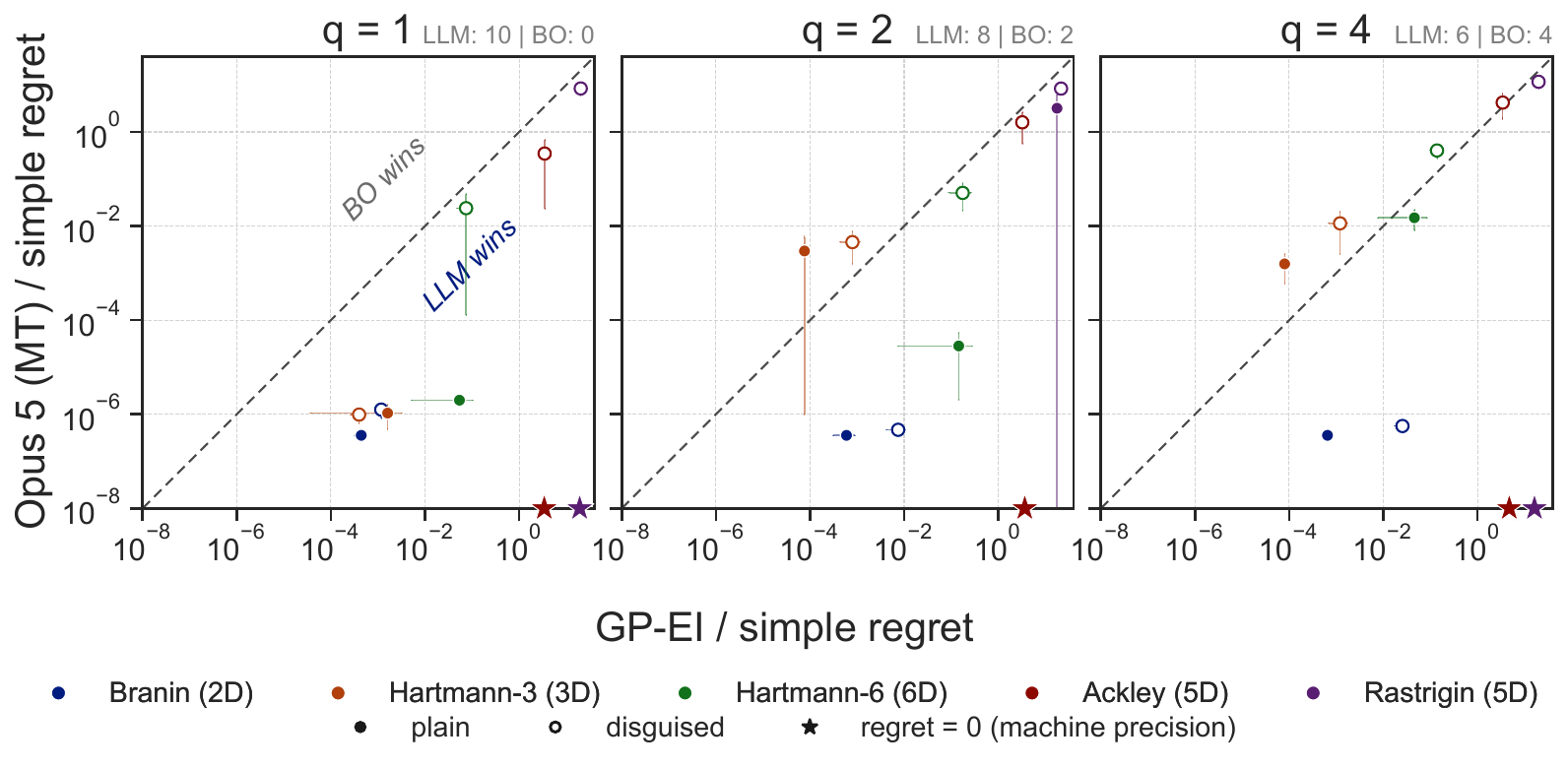}
    \caption{A comparison of the simple regret of Opus 5 multi-turn against GP-EI for batch sizes $q=1,2,4$ across all five synthetic tasks and both disguised and plain variants. Stars indicate cases where the policy achieved a regret that is truly zero up to machine precision. Bars are $\pm$1 standard error (SE) computed across five seeds.}
    \label{fig:opus_synthetic_comparison}
\end{figure}

Figure~\ref{fig:opus_synthetic_comparison} compares Opus 5 multi-turn against GP-EI on the full panel of synthetic test functions for both the plain and disguised variations and batch sizes $q=1,2,4$. We choose Opus 5 multi-turn because it is the strongest-performing LLM condition when compared to its own single-turn variation and the performance of Opus 4.8 and Sonnet 4.6 (for a detailed breakdown of performance, see Appendix~\ref{app:synth_optim}). A policy only obtains a regret value of 0 and is marked with a star if its regret value is 0 up to machine precision. We see that across all batch sizes, Opus 5 multi-turn outperforms GP-EI on a majority of the 10 optimization tasks, with the LLM regret slightly degrading with increasing batch size as evidenced by the upward vertical drift observed for the tasks other than the plain Ackley function and the two Branin variations. This vertical shift is most noticeable for the Hartmann-3 task, where the plain (disguised) average simple regret is $1.00\times 10^{-6}$ ($1.00\times 10^{-6}$) for $q=1$, $2.96\times 10^{-3}$ ($4.57\times 10^{-3}$) for $q=2$, and $1.57\times 10^{-3}$ ($1.14\times 10^{-2}$) for $q=4$. In contrast, the performance of GP-EI remains consistent across batch sizes for both the disguised and plain variations of the tasks, indicating that batch size does not have a significant positive effect on the final attained regret for the non-LLM method.

Comparing across the disguised and plain variations of the tasks, we see that for Opus 5, the model consistently performs worse on the disguised variation compared to the plain variation of each task whereas GP-EI again maintains a consistent level of performance. The discrepancy for the LLM is most noticeable in the cases of the Ackley and Rastrigin tasks, where the model attains an average simple regret of exactly zero for the $q=1$ and $q=4$ plain cases but not for the disguised variants. To understand this discrepancy, we examine the optimization trajectory of the Opus 5 $q=1$ scan across all 10 variations grouped by task as shown in Figure~\ref{fig:plain_disguised_task_comp}. We see that for the lower-dimensional Branin and Hartmann-3 tasks, the average simple regret eventually reaches the low regret floor over the course of the optimization for both the disguised and plain tasks, with the LLM performing comparably well on the disguised task compared to the plain variations. However, for the higher-dimensional tasks Hartmann-6, Ackley, and Rastrigin, disguising both the domain and the range leads to a significant loss in performance, with the model no longer able to attain the zero regret optimum.

\begin{figure}[h]
    \centering
    \includegraphics[width=\textwidth]{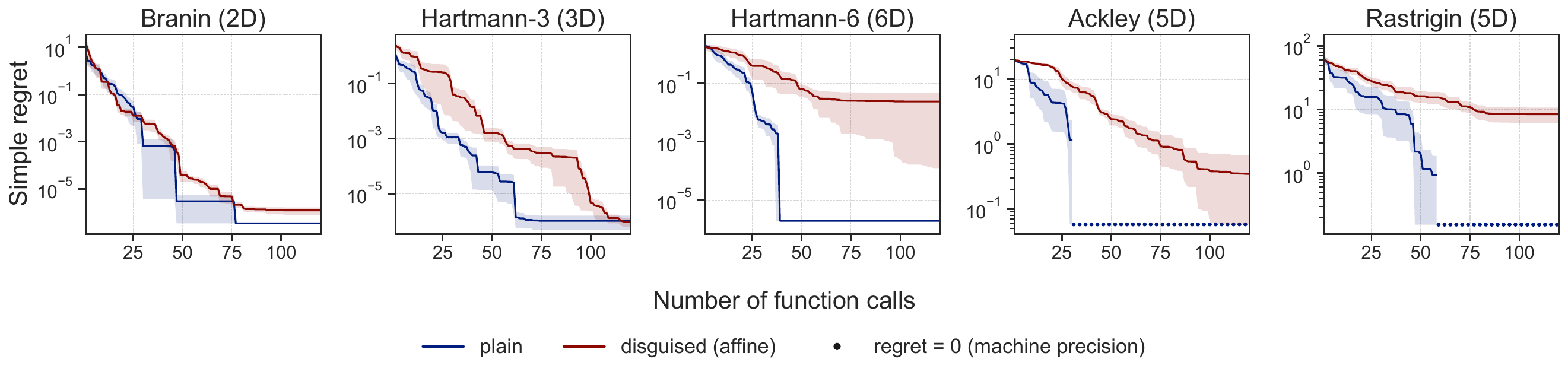}
    \caption{Simple regret versus number of function evaluations for Opus 5 in the multi-turn setting with $q=1$ for all five tasks, with both variations shown. Solid lines represent the mean across five seeds and shaded bands represent $\pm$1 SE. A dotted line corresponds to 0 regret (up to machine precision).}
    \label{fig:plain_disguised_task_comp}
\end{figure}

In addition to the discrepancies between the disguised and plain variants of the tasks and the decreased optimization performance as a function of problem dimensionality, the shapes of the trajectories are also informative about the model's behavior, where the optimization trajectories for the plain tasks show large, rapid decreases in the regret that are not observed for the disguised tasks. We find that this corresponds with a recognition behavior of the model, where the model recognizes the topology of the function it is navigating by reasoning over the evaluated points and then uses that knowledge to refine its optimization. For simpler tasks such as Branin (which has a well-documented shape with three minima) and Hartmann-3, we observe that even when disguising both the domain and range, the model is still able to recall some information about the shape of the function and its optima, resulting in the model solving the disguised and plain tasks equally well. However, without the ability to fingerprint the functions based on their optimal values and optima locations, the LLM performs worse when tasked with the disguised higher-dimensional optimization functions, leading to the larger regret discrepancy. We provide a more detailed memorization analysis in Appendix~\ref{app:synth_mem_analysis}.

Taken together, these results indicate that frontier reasoning LLMs have strong batch optimization capabilities at levels of performance comparable to or better than a strong non-LLM method in BO. However, their performance as batch numerical optimizers is brittle, with variability upon transformations of function domains and ranges, problem dimensionality, and batch size. Furthermore, care needs to be taken when evaluating them on synthetic, well-known test functions that have likely appeared in their pretraining corpus, as even with both domain and range disguises, models are still able to obtain some notion of their identity from their topology.

\subsection{Molecular optimization on PMO}
The experiments using synthetic optimization functions suggest that while frontier reasoning LLMs can function as batch numerical optimizers, their performance is not robust when dealing with arbitrary transformations of the optimization landscape and changes in batch size and search space dimensionality. We now turn to evaluating these reasoning LLMs on their performance in discrete optimization, a more natural setting given that autoregressive LLMs are trained to navigate discrete spaces such as natural language. We use the PMO molecular optimization benchmark which focuses on maximization under a series of deterministic black-box small-molecule oracles with comparisons across a number of different methods.

\begin{figure}[h]
    \centering
    \includegraphics[width=\textwidth]{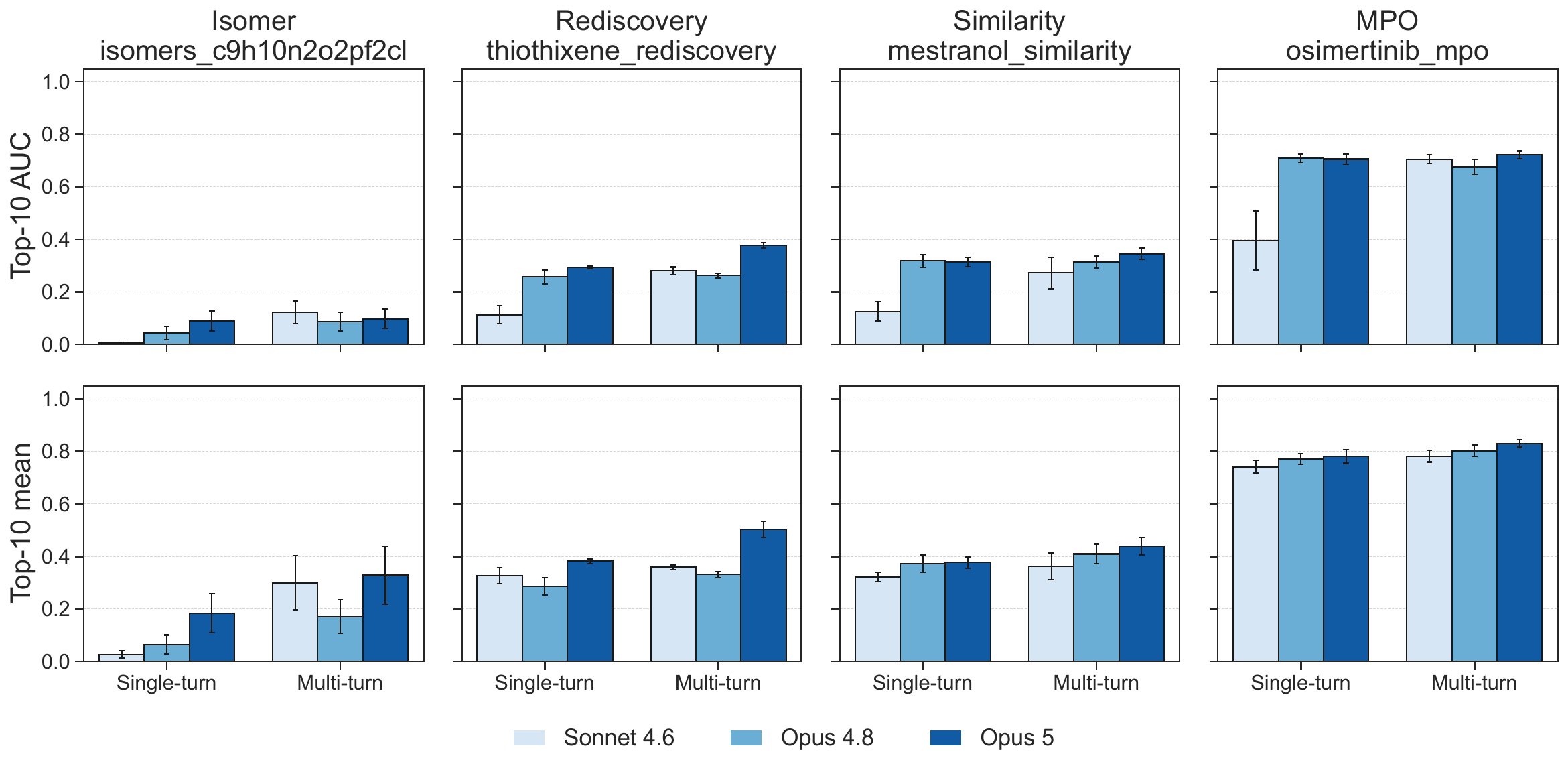}
    \caption{A comparison of Sonnet 4.6, Opus 4.8, and Opus 5 on a selected set of isomer, rediscovery, similarity, and MPO tasks, with all three models evaluated in both single- and multi-turn settings. The top row shows the Top-10 AUC score and the bottom row shows the Top-10 mean score, with error bars corresponding to $\pm$1 SE over five seeds.}
    \label{fig:inter_llm_comp_pmo}
\end{figure}

We restrict our LLM budget to only 210 oracle evaluations, corresponding to an initial set of 10 molecules randomly sampled from the ZINC 250K dataset \citep{tingle_zinc22xe5f8a_2023} followed by 20 batches of 10. Figure~\ref{fig:inter_llm_comp_pmo} compares the Top-10 AUC and the Top-10 mean across Sonnet 4.6, Opus 4.8, and Opus 5 on an example of an isomer task, a rediscovery task, a similarity task, and a multiproperty optimization (MPO) task for clarity, with the full LLM results across all oracles in Appendix~\ref{app:inter_llm_on_pmo}. On the selected tasks, model performance increases with newer generations of frontier reasoning models and multi-turn performance exceeds the corresponding single-turn performance, resulting in Opus 5 multi-turn once again being the best-performing condition. Indeed, on the full set of 23 oracles, Opus 5 is the best by Top-10 AUC in 17 out of the 23 cases, and the best by Top-10 mean in 20 out of the 23 cases. This highlights both the continual improvement of frontier reasoning models and the necessity of stateful, persistent reasoning for navigating complex search spaces.

\begin{figure}[h]
    \centering
    \includegraphics[width=0.95\textwidth]{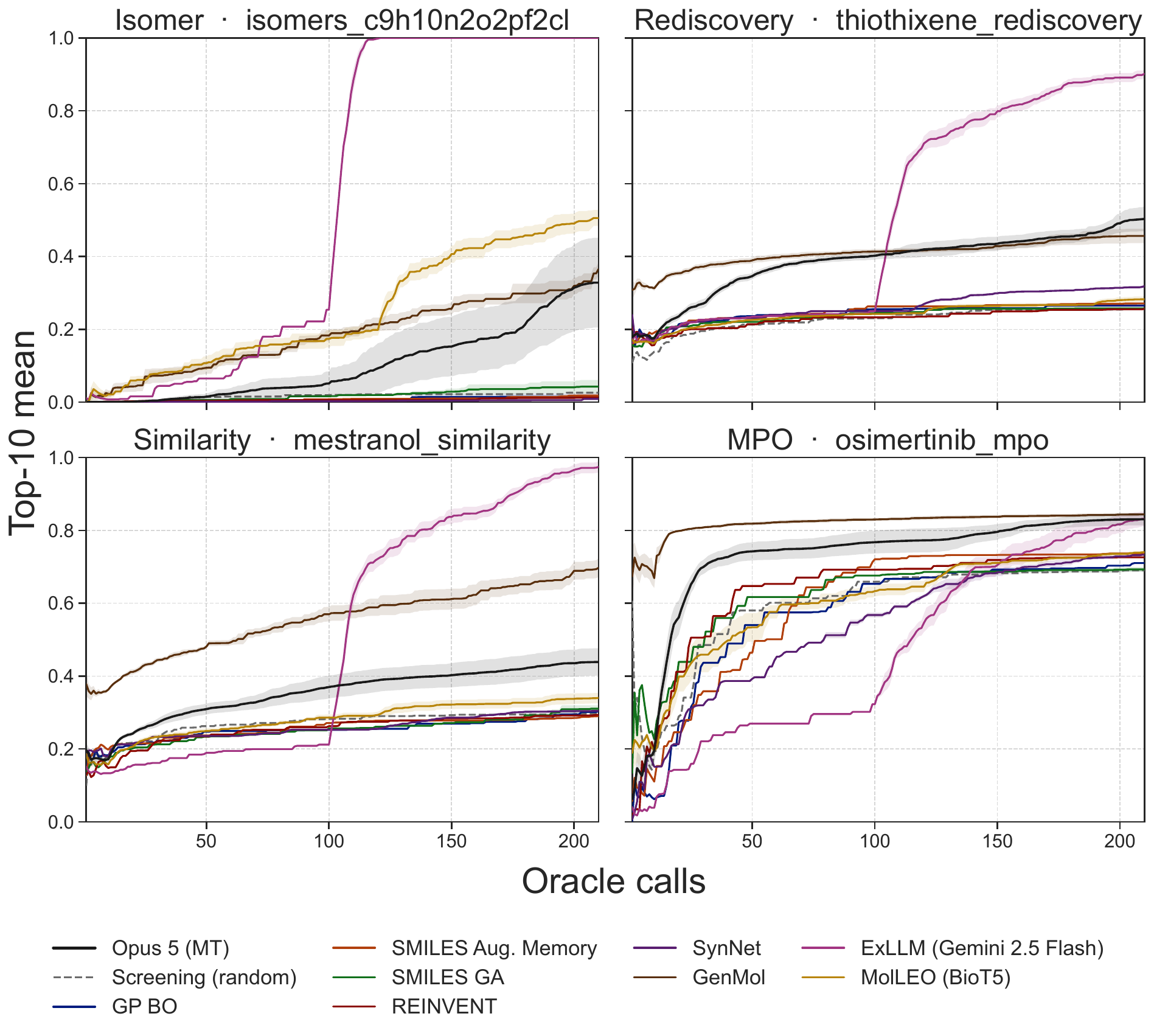}
    \caption{Top-10 mean reward curves for the selected isomer, rediscovery, similarity, and MPO tasks, comparing Opus 5 multi-turn against nine selected specialist methods for each oracle across a budget of 210 oracle evaluations. The shaded regions correspond to $\pm$1 SE over the five seeds.}
    \label{fig:llm_v_specialist_pmo}
\end{figure}

Continuing with Opus 5 multi-turn, we next compare its performance against a subset of nine specialist methods on the same tasks, only showing a subset for clarity. This includes a mix of methods tested in the original PMO paper and methods that we have reproduced, such as SMILES Augmented Memory~\citep{guoAugmentedMemorySampleEfficient2024}, GenMol~\citep{lee_genmol_2025}, ExLLM~\citep{ranExLLMExperienceEnhancedLLM2025}, and MolLEO~\citep{wangEfficientEvolutionarySearch2025}. Some additional methods such as LICO~\citep{nguyenLICOLargeLanguage2025} and GP-MoLFormer-SIM~\citep{navratilGPMoLFormerSimTestTime2025} also benchmark on the PMO task set, but these methods do not provide any code or optimization trajectories necessary for our analysis; as such, we compare our best results against their methods in Appendix~\ref{app:non-repro-comparison}. Figure~\ref{fig:llm_v_specialist_pmo} shows the Top-10 mean as a function of the oracle call number for each of the tasks, ranging from 1 to the maximum budget of 210. We see that in the low-budget regime, the reasoning LLM is a competitive optimizer, exceeding the Top-10 mean of some bespoke deep-learning-based methods like REINVENT~\citep{blaschke_reinvent_2020} and SynNet~\citep{gao_amortized_2022} as well as BO-based methods and search methods. When compared to more specialized methods like GenMol, MolLEO, and ExLLM, Opus 5 maintains comparable performance for oracles such as thiothixene rediscovery and osimertinib MPO, but falls behind specialist methods on the selected isomer and mestranol similarity tasks.

While the differences in performance between Opus 5 and these alternative methods can appear especially large, it is worth noting that these approaches use either a carefully trained specialist architecture already tuned on drug-like molecules (as in the case of GenMol) or an LLM to drive the optimization with significantly more scaffolding around the problem (as in the case of MolLEO and ExLLM). GenMol, especially, has an advantage in molecular optimization tasks because it was trained on an extensive dataset of small molecules from the SAFE dataset~\citep{noutahi_gotta_2023}, meaning it is more efficient at generating drug-like compounds off the shelf. This is a possible source of superior performance when compared to the LLM-based methods where it wins 13 of 23 oracles by Top-10 AUC under an oracle call budget of 210. However, we note that in terms of Top-10 mean at a budget of 210, GenMol is comparable to Opus 5 multi-turn (0.590 average across oracles for GenMol versus 0.596 for Opus 5) with ExLLM being the best-performing method with an average of 0.738 across oracles (see Appendix~\ref{app:molec_optim}). In contrast to these methods, Opus 5 is used directly without any additional tuning or scaffolding, so the fact that the frontier reasoning LLM maintains comparable performance is especially impressive and indicative of the evolving capabilities of these models. Additionally, comparing the LLM performance in the discrete setting to the performance in the continuous setting in Section~\ref{sec:synth_tasks}, it is clear that the reasoning LLM functions as a better batch optimizer in discrete settings.

\begin{figure}[h]
    \centering
    \includegraphics[width=0.95\textwidth]{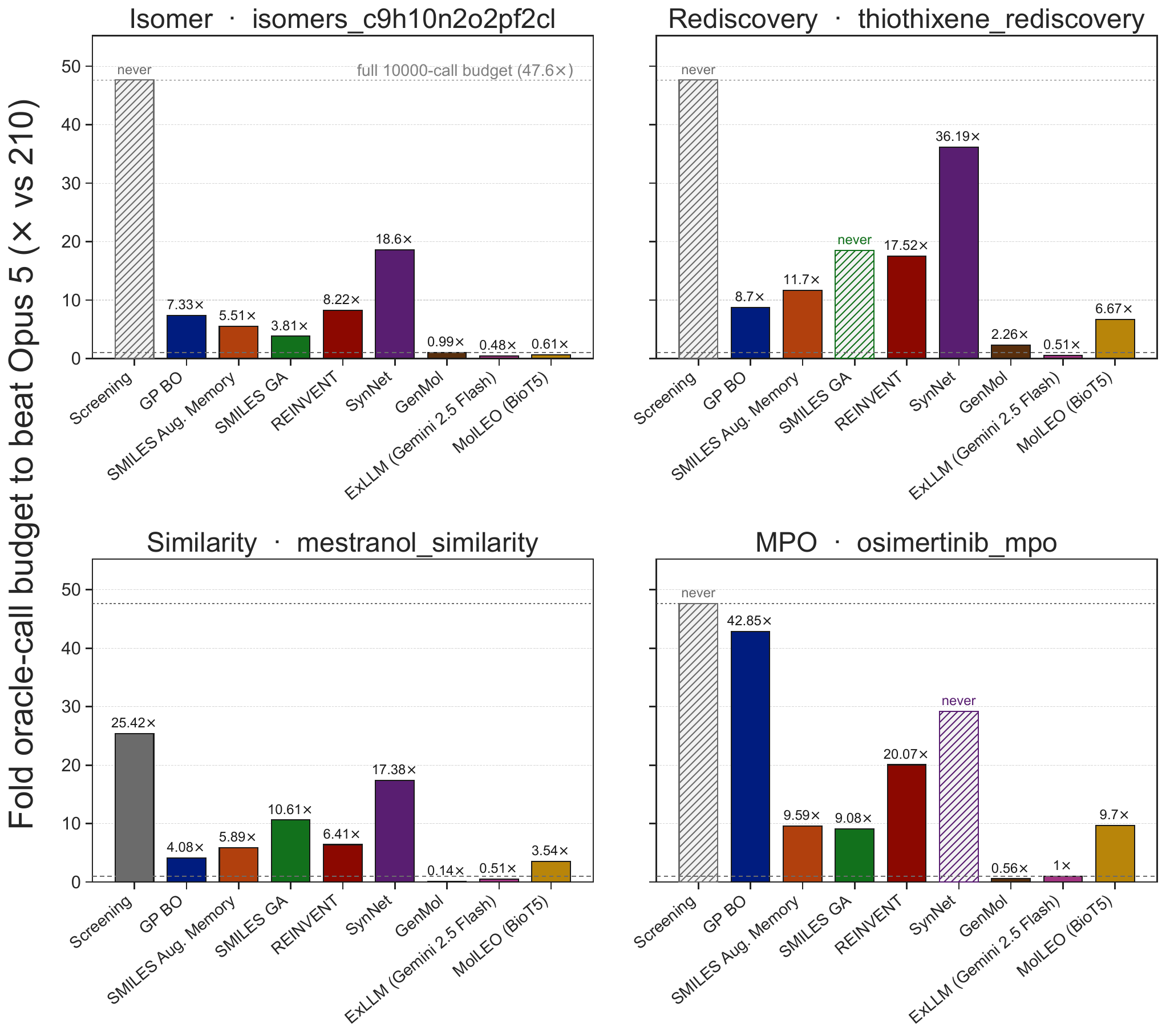}
    \caption{Comparison of the number of additional oracle calls required for specialist methods' Top-10 mean to reach LLM performance across the selected isomer, rediscovery, similarity, and MPO tasks. The fold value is computed using the median of the first step to reach LLM performance, taken over the set of seeds that reached LLM-level performance. Methods that never reach LLM performance within their optimization are represented as hatched bars.}
    \label{fig:llm_v_specialist_fold}
\end{figure}

Given that Opus 5 multi-turn remains competitive against other methods in the low-budget regime, we can extend our analysis by measuring the performance improvement when the full oracle budget is allowed. Doing this allows us to quantify how many additional (or fewer) oracle calls are required before a given method achieves performance comparable to that of the LLM, resulting in a budget multiplier where a lower multiplier indicates a more sample-efficient method. Figure~\ref{fig:llm_v_specialist_fold} shows the comparison for a selection of tasks, where we have used the same set of methods and tasks. As expected based on the reward curves, we see that GenMol, ExLLM, and MolLEO are more sample efficient than Opus 5, all of them taking fewer than 210 oracle calls to match the performance in the isomer task. ExLLM is especially consistent, achieving a multiplier of $\leq1\times$ across the four selected tasks, with GenMol as a close second where it was only less efficient than Opus 5 on the thiothixene rediscovery task. For the other selected methods, we see that even the most sample-efficient of this group requires at least three times as many oracle calls to achieve comparable performance, with many methods requiring 20--40 times as many oracle calls. Certain methods also never reach LLM performance throughout the course of their optimization, as shown by the performance of screening in three of the four tasks, SMILES GA in thiothixene rediscovery, and SynNet in osimertinib MPO. Having compared across both the raw performance and sample efficiency axes, we see that while frontier reasoning LLMs still underperform carefully trained and highly scaffolded specialist approaches, their zero-shot performance is still competitive with a range of optimization methods while maintaining sample efficiency. This highlights their utility as batch optimizers in discrete settings, with future iterations of frontier models likely leading to continued improvements in this regard. For full results on the PMO tasks, see Appendix~\ref{app:molec_optim}. 

\section{Conclusion}

Here we present an analysis of frontier reasoning LLMs as batch optimizers for both continuous and discrete BBO tasks. We show that while LLMs are competitive with GP-EI on synthetic test functions, their performance is brittle and dependent on factors such as batch size, task dimensionality, and extent of transformation applied. However, in the discrete setting of molecular optimization, we show that LLMs are both highly performant and sample-efficient optimizers, beating a host of bespoke methods without additional tuning or problem scaffolding and underperforming only when compared to specially trained or highly structured strategies. In the context of BBO-type problems, our results indicate that frontier reasoning LLMs are powerful priors for batch discrete optimization even without any additional tuning, with potentially wide-reaching utility. Immediate future directions for our work include extending our analysis to new numerical and discrete settings to evaluate LLM performance on additional BBO problem formulations and designing mid- and post-training protocols for enhancing open-weight reasoning LLM performance on BBO tasks.

\begin{ack}
\acktext
\end{ack}

\clearpage
\printbibliography

\clearpage
\appendix
\renewcommand{\thetable}{A\arabic{table}}
\setcounter{table}{0}
\renewcommand{\thefigure}{A\arabic{figure}}
\setcounter{figure}{0}

\section{Results on synthetic test functions}\label{app:synthetic_results}

\subsection{Description of synthetic test functions}
\label{app:synthetic-functions}

Table~\ref{tab:synth-functions} contains details for the five synthetic test functions used in our analysis. Optimizers are listed in each function's native coordinates. Runs search the normalized unit cube $[0,1]^{d}$, so the effective optimizer is $(\mathbf{x}^\star-\mathbf{l})/(\mathbf{u}-\mathbf{l})$: Hartmann is already defined on $[0,1]^{d}$; for Ackley and Rastrigin the origin maps to the cube center $[0.5]^{d}$. Branin has three equivalent global minimizers, which map to $\approx(0.124,0.818),\,(0.543,0.152),\,(0.962,0.165)$ in the unit cube.

\begin{table}[!h]
  \centering
  \caption{Classical synthetic optimization test functions used in the
    budget-matched benchmark. Each function is evaluated in a \emph{plain} form
    and a \emph{disguised} twin (a random bijection of the
    input cube that hides the optimizer while leaving the dimensionality $d$ and
    the optimal value $f(\mathbf{x}^\star)$ unchanged); both share the row below.
    Runs \emph{maximize} $-f$, so the target value used for simple regret is
    $f^\star=-f(\mathbf{x}^\star)$.}
  \label{tab:synth-functions}
  \small
  \begin{tabular*}{\textwidth}{@{\extracolsep{\fill}}l r r r r@{}}
    \toprule
    Function & $d$ & Search domain & Global optimizer(s) $\mathbf{x}^\star$ & $f(\mathbf{x}^\star)$ \\
    \midrule
    Branin & $2$ & $[-5,10]\times[0,15]$ &
    \begin{tabular}[c]{@{}r@{}}$(-\pi,\,12.275)$\\ $(\pi,\,2.275)$\\ $(9.42478,\,2.475)$\end{tabular} &
      $0.397887$ \\
    \addlinespace
    Hartmann-3 & $3$ & $[0,1]^{3}$ &
      $(0.1146,\,0.5556,\,0.8525)$ & $-3.86278$ \\
    \addlinespace
    Hartmann-6 & $6$ & $[0,1]^{6}$ &
      \begin{tabular}[c]{@{}l@{}}$(0.2017,\,0.1500,\,0.4769,$\\ $\phantom{(}0.2753,\,0.3117,\,0.6573)$\end{tabular} &
      $-3.32237$ \\
    \addlinespace
    Ackley & $5$ & $[-32.768,\,32.768]^{5}$ &
      $\mathbf{0}$ \; (origin) & $0$ \\
    \addlinespace
    Rastrigin & $5$ & $[-5.12,\,5.12]^{5}$ &
      $\mathbf{0}$ \; (origin) & $0$ \\
    \bottomrule
  \end{tabular*}
\end{table}

\subsection{Full results for synthetic optimization tasks}
\label{app:synth_optim}

Tables~\ref{tab:affine-detail-plain} and~\ref{tab:affine-detail-disguised} contain full comparisons of the frontier reasoning LLMs against GP-EI and a random baseline across all batch sizes $q=1,2,4$ and both single- and multi-turn protocols for the LLMs. We choose to show the regret values in scientific notation rather than rounding because for the Ackley and Rastrigin functions, the LLMs are capable of emitting the exact analytical optimum, resulting in a regret that is truly 0 up to machine precision. For an example of the prompt used for optimizing with the synthetic functions with an LLM, see Prompt~\ref{lst:prompt_synthetic}.

\iflandscapewide\begin{landscape}\fi
\begin{table}[!h]
\centering
\scriptsize
\setlength{\tabcolsep}{3pt}
\caption{\textbf{Plain synthetic tasks.} Final simple regret (mean $\pm$1 SE, $n=5$ seeds) in scientific notation, for every batch size $q$, arm, and protocol. Budget-matched: $n_\text{batches}=120/q$. Lower is better; \textbf{bold} = best arm for that task at that $q$.}
\label{tab:affine-detail-plain}
\begin{tabular}{@{}ll l l *{5}{r}@{}}
\toprule
$q$ & rounds & Model & Protocol & Branin & Hartmann-3 & Hartmann-6 & Ackley & Rastrigin \\
\midrule
\multirow{8}{*}{1} & \multirow{8}{*}{120} & GP-EI & --- & $ 4.42\times10^{-4} \pm 1.2\times10^{-4} $ & $ 1.61\times10^{-3} \pm 1.6\times10^{-3} $ & $ 5.39\times10^{-2} \pm 4.9\times10^{-2} $ & $ 3.41\times10^{0} \pm 9.5\times10^{-2} $ & $ 1.93\times10^{1} \pm 3.2\times10^{0} $ \\
 &  & Random & --- & $ 6.66\times10^{-1} \pm 2.9\times10^{-1} $ & $ 2.48\times10^{-1} \pm 6.7\times10^{-2} $ & $ 1.40\times10^{0} \pm 1.9\times10^{-1} $ & $ 1.56\times10^{1} \pm 1.6\times10^{0} $ & $ 3.49\times10^{1} \pm 2.4\times10^{0} $ \\
 &  & \multirow{2}{*}{Sonnet 4.6} & single-turn & $ 8.21\times10^{-1} \pm 8.2\times10^{-1} $ & $ 1.00\times10^{-6} \pm 4.5\times10^{-7} $ & $ 1.00\times10^{0} \pm 4.1\times10^{-1} $ & $ 1.46\times10^{1} \pm 3.7\times10^{0} $ & $ 2.82\times10^{1} \pm 6.7\times10^{0} $ \\
 &  &  & multi-turn & $ 3.71\times10^{-7} \pm 7.7\times10^{-9} $ & $ 1.55\times10^{-1} \pm 1.5\times10^{-1} $ & $ 2.25\times10^{-3} \pm 2.2\times10^{-3} $ & {\boldmath$ 0 $} & {\boldmath$ 0 $} \\
 &  & \multirow{2}{*}{Opus 4.8} & single-turn & $ 2.87\times10^{-3} \pm 2.8\times10^{-3} $ & $ 4.00\times10^{-2} \pm 2.2\times10^{-2} $ & $ 7.93\times10^{-1} \pm 2.4\times10^{-1} $ & $ 1.49\times10^{1} \pm 2.2\times10^{0} $ & $ 2.84\times10^{1} \pm 6.7\times10^{0} $ \\
 &  &  & multi-turn & $ 9.33\times10^{-1} \pm 9.3\times10^{-1} $ & $ 7.04\times10^{-4} \pm 3.0\times10^{-4} $ & $ 8.89\times10^{-4} \pm 4.7\times10^{-4} $ & {\boldmath$ 0 $} & $ 1.79\times10^{0} \pm 1.8\times10^{0} $ \\
 &  & \multirow{2}{*}{Opus 5} & single-turn & {\boldmath$ 3.58\times10^{-7} \pm 0.0\times10^{0} $} & $ 1.59\times10^{-1} \pm 1.3\times10^{-1} $ & $ 3.31\times10^{-1} \pm 2.5\times10^{-1} $ & $ 8.79\times10^{0} \pm 3.7\times10^{0} $ & $ 2.49\times10^{1} \pm 1.0\times10^{1} $ \\
 &  &  & multi-turn & $ 3.58\times10^{-7} \pm 1.3\times10^{-11} $ & {\boldmath$ 1.00\times10^{-6} \pm 4.5\times10^{-7} $} & {\boldmath$ 2.00\times10^{-6} \pm 0.0\times10^{0} $} & {\boldmath$ 0 $} & {\boldmath$ 0 $} \\
\midrule
\multirow{8}{*}{2} & \multirow{8}{*}{60} & GP-EI & --- & $ 5.96\times10^{-4} \pm 2.9\times10^{-4} $ & $ 7.60\times10^{-5} \pm 1.2\times10^{-5} $ & $ 1.44\times10^{-1} \pm 1.4\times10^{-1} $ & $ 3.66\times10^{0} \pm 5.6\times10^{-1} $ & $ 1.77\times10^{1} \pm 1.9\times10^{0} $ \\
 &  & Random & --- & $ 4.12\times10^{-1} \pm 1.7\times10^{-1} $ & $ 2.12\times10^{-1} \pm 8.1\times10^{-2} $ & $ 1.32\times10^{0} \pm 1.6\times10^{-1} $ & $ 1.58\times10^{1} \pm 1.6\times10^{0} $ & $ 3.45\times10^{1} \pm 2.9\times10^{0} $ \\
 &  & \multirow{2}{*}{Sonnet 4.6} & single-turn & $ 1.00\times10^{-6} \pm 4.5\times10^{-7} $ & $ 5.14\times10^{-2} \pm 3.4\times10^{-2} $ & $ 7.73\times10^{-1} \pm 3.1\times10^{-1} $ & $ 7.08\times10^{0} \pm 4.3\times10^{0} $ & $ 3.21\times10^{1} \pm 9.4\times10^{0} $ \\
 &  &  & multi-turn & $ 6.35\times10^{-7} \pm 2.6\times10^{-7} $ & {\boldmath$ 5.40\times10^{-5} \pm 5.4\times10^{-5} $} & $ 7.72\times10^{-3} \pm 4.6\times10^{-3} $ & {\boldmath$ 0 $} & $ 1.45\times10^{1} \pm 6.2\times10^{0} $ \\
 &  & \multirow{2}{*}{Opus 4.8} & single-turn & $ 5.63\times10^{-1} \pm 5.5\times10^{-1} $ & $ 6.50\times10^{-2} \pm 2.7\times10^{-2} $ & $ 1.48\times10^{0} \pm 3.6\times10^{-1} $ & $ 1.53\times10^{1} \pm 1.7\times10^{0} $ & $ 2.39\times10^{1} \pm 7.3\times10^{0} $ \\
 &  &  & multi-turn & $ 1.00\times10^{-6} \pm 4.5\times10^{-7} $ & $ 4.71\times10^{-3} \pm 4.7\times10^{-3} $ & $ 5.18\times10^{-1} \pm 3.2\times10^{-1} $ & {\boldmath$ 0 $} & $ 1.15\times10^{1} \pm 4.4\times10^{0} $ \\
 &  & \multirow{2}{*}{Opus 5} & single-turn & $ 1.00\times10^{-6} \pm 4.5\times10^{-7} $ & $ 9.36\times10^{-2} \pm 2.6\times10^{-2} $ & $ 7.59\times10^{-1} \pm 3.8\times10^{-1} $ & $ 3.05\times10^{0} \pm 3.0\times10^{0} $ & $ 3.00\times10^{1} \pm 8.1\times10^{0} $ \\
 &  &  & multi-turn & {\boldmath$ 3.58\times10^{-7} \pm 0.0\times10^{0} $} & $ 2.96\times10^{-3} \pm 3.0\times10^{-3} $ & {\boldmath$ 2.80\times10^{-5} \pm 2.6\times10^{-5} $} & {\boldmath$ 0 $} & {\boldmath$ 3.19\times10^{0} \pm 3.2\times10^{0} $} \\
\midrule
\multirow{8}{*}{4} & \multirow{8}{*}{30} & GP-EI & --- & $ 6.53\times10^{-4} \pm 1.3\times10^{-4} $ & $ 8.00\times10^{-5} \pm 1.7\times10^{-5} $ & $ 4.58\times10^{-2} \pm 3.8\times10^{-2} $ & $ 4.75\times10^{0} \pm 3.0\times10^{-1} $ & $ 1.63\times10^{1} \pm 2.3\times10^{0} $ \\
 &  & Random & --- & $ 2.82\times10^{-1} \pm 6.2\times10^{-2} $ & $ 2.12\times10^{-1} \pm 1.2\times10^{-1} $ & $ 1.24\times10^{0} \pm 1.9\times10^{-1} $ & $ 1.47\times10^{1} \pm 1.4\times10^{0} $ & $ 4.24\times10^{1} \pm 1.8\times10^{0} $ \\
 &  & \multirow{2}{*}{Sonnet 4.6} & single-turn & $ 2.57\times10^{-2} \pm 2.6\times10^{-2} $ & $ 1.19\times10^{-2} \pm 1.2\times10^{-2} $ & $ 7.39\times10^{-1} \pm 3.9\times10^{-1} $ & $ 1.02\times10^{1} \pm 4.2\times10^{0} $ & $ 2.93\times10^{1} \pm 6.8\times10^{0} $ \\
 &  &  & multi-turn & $ 5.64\times10^{-7} \pm 1.6\times10^{-7} $ & {\boldmath$ 3.19\times10^{-7} \pm 7.4\times10^{-8} $} & $ 4.07\times10^{-1} \pm 3.1\times10^{-1} $ & $ 3.32\times10^{0} \pm 3.3\times10^{0} $ & $ 2.51\times10^{1} \pm 9.1\times10^{0} $ \\
 &  & \multirow{2}{*}{Opus 4.8} & single-turn & $ 3.00\times10^{-6} \pm 2.7\times10^{-6} $ & $ 5.58\times10^{-2} \pm 2.2\times10^{-2} $ & $ 9.50\times10^{-1} \pm 3.8\times10^{-1} $ & $ 9.86\times10^{0} \pm 3.1\times10^{0} $ & $ 2.79\times10^{1} \pm 8.6\times10^{0} $ \\
 &  &  & multi-turn & $ 5.26\times10^{-7} \pm 7.2\times10^{-8} $ & $ 2.63\times10^{-3} \pm 2.4\times10^{-3} $ & $ 3.25\times10^{-1} \pm 1.3\times10^{-1} $ & $ 9.33\times10^{-1} \pm 9.3\times10^{-1} $ & $ 1.12\times10^{1} \pm 2.0\times10^{0} $ \\
 &  & \multirow{2}{*}{Opus 5} & single-turn & $ 3.94\times10^{-7} \pm 3.6\times10^{-8} $ & $ 5.92\times10^{-2} \pm 2.9\times10^{-2} $ & $ 7.33\times10^{-1} \pm 2.8\times10^{-1} $ & $ 6.20\times10^{0} \pm 3.8\times10^{0} $ & $ 3.04\times10^{1} \pm 7.7\times10^{0} $ \\
 &  &  & multi-turn & {\boldmath$ 3.58\times10^{-7} \pm 5.2\times10^{-12} $} & $ 1.57\times10^{-3} \pm 9.8\times10^{-4} $ & {\boldmath$ 1.50\times10^{-2} \pm 7.0\times10^{-3} $} & {\boldmath$ 0 $} & {\boldmath$ 0 $} \\
\bottomrule
\end{tabular}
\end{table}
\iflandscapewide\end{landscape}\fi

\iflandscapewide\begin{landscape}\fi
\begin{table}[!h]
\centering
\scriptsize
\setlength{\tabcolsep}{3pt}
\caption{\textbf{Disguised synthetic tasks} (domain bijection $+$ affine output map $y'=ay+b$, $a>0$; regret recovered by dividing by $a$ so it is comparable to the plain tasks). Final simple regret (mean $\pm$1 SE, $n=5$ seeds) in scientific notation, for every batch size $q$, arm, and protocol. Budget-matched: $n_\text{batches}=120/q$. Lower is better; \textbf{bold} = best arm for that task at that $q$.}
\label{tab:affine-detail-disguised}
\begin{tabular}{@{}ll l l *{5}{r}@{}}
\toprule
$q$ & rounds & Model & Protocol & Branin & Hartmann-3 & Hartmann-6 & Ackley & Rastrigin \\
\midrule
\multirow{8}{*}{1} & \multirow{8}{*}{120} & GP-EI & --- & $ 1.18\times10^{-3} \pm 3.0\times10^{-4} $ & $ 3.99\times10^{-4} \pm 1.3\times10^{-4} $ & $ 7.45\times10^{-2} \pm 2.7\times10^{-2} $ & $ 3.48\times10^{0} \pm 2.2\times10^{-1} $ & $ 2.03\times10^{1} \pm 4.2\times10^{0} $ \\
 &  & Random & --- & $ 4.48\times10^{-1} \pm 1.5\times10^{-1} $ & $ 2.76\times10^{-1} \pm 8.2\times10^{-2} $ & $ 1.27\times10^{0} \pm 2.0\times10^{-1} $ & $ 1.86\times10^{1} \pm 2.9\times10^{-1} $ & $ 3.94\times10^{1} \pm 4.8\times10^{0} $ \\
 &  & \multirow{2}{*}{Sonnet 4.6} & single-turn & $ 7.02\times10^{-4} \pm 7.0\times10^{-4} $ & $ 1.30\times10^{0} \pm 6.5\times10^{-1} $ & $ 7.55\times10^{-1} \pm 2.7\times10^{-1} $ & $ 1.77\times10^{1} \pm 6.9\times10^{-1} $ & $ 3.35\times10^{1} \pm 7.9\times10^{0} $ \\
 &  &  & multi-turn & $ 5.00\times10^{-6} \pm 3.1\times10^{-6} $ & $ 1.14\times10^{0} \pm 7.0\times10^{-1} $ & $ 3.53\times10^{-1} \pm 3.2\times10^{-1} $ & $ 4.81\times10^{0} \pm 3.6\times10^{0} $ & $ 2.98\times10^{1} \pm 4.3\times10^{0} $ \\
 &  & \multirow{2}{*}{Opus 4.8} & single-turn & $ 3.09\times10^{-1} \pm 3.1\times10^{-1} $ & $ 1.88\times10^{0} \pm 6.2\times10^{-1} $ & $ 1.18\times10^{0} \pm 3.3\times10^{-1} $ & $ 1.83\times10^{1} \pm 4.3\times10^{-1} $ & $ 3.20\times10^{1} \pm 5.8\times10^{0} $ \\
 &  &  & multi-turn & $ 3.75\times10^{-3} \pm 3.3\times10^{-3} $ & $ 7.29\times10^{-1} \pm 5.5\times10^{-1} $ & $ 2.88\times10^{-2} \pm 2.6\times10^{-2} $ & $ 6.42\times10^{0} \pm 4.0\times10^{0} $ & $ 2.46\times10^{1} \pm 3.4\times10^{0} $ \\
 &  & \multirow{2}{*}{Opus 5} & single-turn & $ 1.86\times10^{-1} \pm 1.8\times10^{-1} $ & $ 7.31\times10^{-1} \pm 5.5\times10^{-1} $ & $ 9.06\times10^{-1} \pm 3.5\times10^{-1} $ & $ 1.60\times10^{1} \pm 7.5\times10^{-1} $ & $ 2.39\times10^{1} \pm 8.2\times10^{0} $ \\
 &  &  & multi-turn & {\boldmath$ 1.00\times10^{-6} \pm 4.5\times10^{-7} $} & {\boldmath$ 1.00\times10^{-6} \pm 4.5\times10^{-7} $} & {\boldmath$ 2.40\times10^{-2} \pm 2.4\times10^{-2} $} & {\boldmath$ 3.48\times10^{-1} \pm 3.2\times10^{-1} $} & {\boldmath$ 8.40\times10^{0} \pm 2.3\times10^{0} $} \\
\midrule
\multirow{8}{*}{2} & \multirow{8}{*}{60} & GP-EI & --- & $ 7.46\times10^{-3} \pm 3.2\times10^{-3} $ & {\boldmath$ 7.96\times10^{-4} \pm 3.6\times10^{-4} $} & $ 1.75\times10^{-1} \pm 8.5\times10^{-2} $ & $ 3.23\times10^{0} \pm 2.2\times10^{-1} $ & $ 2.16\times10^{1} \pm 3.1\times10^{0} $ \\
 &  & Random & --- & $ 3.11\times10^{-1} \pm 6.1\times10^{-2} $ & $ 3.82\times10^{-1} \pm 1.2\times10^{-1} $ & $ 1.39\times10^{0} \pm 2.2\times10^{-1} $ & $ 1.86\times10^{1} \pm 2.9\times10^{-1} $ & $ 4.11\times10^{1} \pm 5.4\times10^{0} $ \\
 &  & \multirow{2}{*}{Sonnet 4.6} & single-turn & $ 1.35\times10^{-1} \pm 1.3\times10^{-1} $ & $ 1.35\times10^{0} \pm 6.3\times10^{-1} $ & $ 1.25\times10^{0} \pm 4.3\times10^{-1} $ & $ 1.79\times10^{1} \pm 5.9\times10^{-1} $ & $ 3.34\times10^{1} \pm 5.5\times10^{0} $ \\
 &  &  & multi-turn & $ 2.00\times10^{-6} \pm 1.8\times10^{-6} $ & $ 1.30\times10^{0} \pm 6.5\times10^{-1} $ & $ 2.33\times10^{-1} \pm 2.0\times10^{-1} $ & $ 8.17\times10^{0} \pm 3.7\times10^{0} $ & $ 2.73\times10^{1} \pm 5.1\times10^{0} $ \\
 &  & \multirow{2}{*}{Opus 4.8} & single-turn & $ 4.16\times10^{-1} \pm 4.2\times10^{-1} $ & $ 1.36\times10^{0} \pm 6.5\times10^{-1} $ & $ 1.09\times10^{0} \pm 3.4\times10^{-1} $ & $ 1.68\times10^{1} \pm 7.7\times10^{-1} $ & $ 3.69\times10^{1} \pm 4.2\times10^{0} $ \\
 &  &  & multi-turn & $ 4.00\times10^{-6} \pm 2.2\times10^{-6} $ & $ 5.73\times10^{-1} \pm 5.7\times10^{-1} $ & $ 4.47\times10^{-1} \pm 3.0\times10^{-1} $ & $ 1.04\times10^{1} \pm 3.7\times10^{0} $ & $ 2.45\times10^{1} \pm 4.9\times10^{0} $ \\
 &  & \multirow{2}{*}{Opus 5} & single-turn & $ 2.34\times10^{-3} \pm 2.3\times10^{-3} $ & $ 1.31\times10^{0} \pm 6.5\times10^{-1} $ & $ 7.60\times10^{-1} \pm 2.6\times10^{-1} $ & $ 1.48\times10^{1} \pm 1.5\times10^{0} $ & $ 2.22\times10^{1} \pm 4.1\times10^{0} $ \\
 &  &  & multi-turn & {\boldmath$ 4.69\times10^{-7} \pm 6.7\times10^{-8} $} & $ 4.57\times10^{-3} \pm 3.0\times10^{-3} $ & {\boldmath$ 5.07\times10^{-2} \pm 2.9\times10^{-2} $} & {\boldmath$ 1.60\times10^{0} \pm 1.0\times10^{0} $} & {\boldmath$ 8.36\times10^{0} \pm 1.1\times10^{0} $} \\
\midrule
\multirow{8}{*}{4} & \multirow{8}{*}{30} & GP-EI & --- & $ 2.55\times10^{-2} \pm 7.9\times10^{-3} $ & {\boldmath$ 1.21\times10^{-3} \pm 5.3\times10^{-4} $} & {\boldmath$ 1.38\times10^{-1} \pm 3.7\times10^{-2} $} & {\boldmath$ 3.44\times10^{0} \pm 5.1\times10^{-1} $} & $ 1.99\times10^{1} \pm 2.4\times10^{0} $ \\
 &  & Random & --- & $ 2.49\times10^{-1} \pm 6.1\times10^{-2} $ & $ 2.15\times10^{-1} \pm 5.0\times10^{-2} $ & $ 1.27\times10^{0} \pm 1.3\times10^{-1} $ & $ 1.85\times10^{1} \pm 3.5\times10^{-1} $ & $ 4.84\times10^{1} \pm 3.2\times10^{0} $ \\
 &  & \multirow{2}{*}{Sonnet 4.6} & single-turn & $ 3.00\times10^{-6} \pm 2.7\times10^{-6} $ & $ 1.30\times10^{0} \pm 6.5\times10^{-1} $ & $ 6.48\times10^{-1} \pm 3.0\times10^{-1} $ & $ 1.47\times10^{1} \pm 3.1\times10^{0} $ & $ 3.24\times10^{1} \pm 5.4\times10^{0} $ \\
 &  &  & multi-turn & {\boldmath$ 4.86\times10^{-7} \pm 9.6\times10^{-8} $} & $ 1.30\times10^{0} \pm 6.5\times10^{-1} $ & $ 3.83\times10^{-1} \pm 3.4\times10^{-1} $ & $ 1.37\times10^{1} \pm 3.4\times10^{0} $ & $ 2.14\times10^{1} \pm 4.8\times10^{0} $ \\
 &  & \multirow{2}{*}{Opus 4.8} & single-turn & $ 1.83\times10^{-2} \pm 1.8\times10^{-2} $ & $ 1.61\times10^{0} \pm 6.1\times10^{-1} $ & $ 1.01\times10^{0} \pm 3.6\times10^{-1} $ & $ 1.69\times10^{1} \pm 5.7\times10^{-1} $ & $ 2.90\times10^{1} \pm 5.4\times10^{0} $ \\
 &  &  & multi-turn & $ 3.00\times10^{-6} \pm 8.9\times10^{-7} $ & $ 1.30\times10^{0} \pm 6.5\times10^{-1} $ & $ 4.48\times10^{-1} \pm 3.0\times10^{-1} $ & $ 1.17\times10^{1} \pm 3.1\times10^{0} $ & $ 1.34\times10^{1} \pm 2.2\times10^{0} $ \\
 &  & \multirow{2}{*}{Opus 5} & single-turn & $ 9.46\times10^{-2} \pm 9.4\times10^{-2} $ & $ 1.30\times10^{0} \pm 6.5\times10^{-1} $ & $ 8.13\times10^{-1} \pm 4.0\times10^{-1} $ & $ 9.59\times10^{0} \pm 2.6\times10^{0} $ & $ 2.15\times10^{1} \pm 6.1\times10^{0} $ \\
 &  &  & multi-turn & $ 1.00\times10^{-6} \pm 4.5\times10^{-7} $ & $ 1.14\times10^{-2} \pm 8.9\times10^{-3} $ & $ 4.04\times10^{-1} \pm 1.3\times10^{-1} $ & $ 4.21\times10^{0} \pm 2.3\times10^{0} $ & {\boldmath$ 1.17\times10^{1} \pm 3.3\times10^{0} $} \\
\bottomrule
\end{tabular}
\end{table}
\iflandscapewide\end{landscape}\fi

\subsection{Analysis of LLM memorization on synthetic tasks}\label{app:synth_mem_analysis}

As noted in Section~\ref{sec:synth_tasks} for the synthetic test functions, examination of the reasoning traces for the model showed that the model at different points along the optimization trajectory recalled the specific function that was being optimized, leveraging this information to refine its proposals. Figure~\ref{fig:mem_breakdown} shows the effect of function recall on the regret for both the disguised and plain variations of the five synthetic test functions pooled across all three models in multi-turn mode with $q=1$, with the y-axis expressing the regret ratio between a specific turn and the turn immediately preceding the recall event. We see that in the plain case, when the function name is recalled, the regret relative to the pre-recall regret tends to drop by several orders of magnitude around that point, indicating that the model is exploiting information about the function to improve its proposals, moving closer to the optima. However, once the domain bijection and range affine transformation are applied, the model's behavior changes dramatically, with there being a slower decline of the regret after function name recall. For simple tasks like Branin, it is clear that function recall still has some benefit since the regret ratio decreases, but not to the same extent as in the plain case. This indicates that disguising the task forces the model to genuinely optimize, even if it can recall some helpful information about the synthetic task from its pretraining. 

\begin{figure}[!h]
    \centering
    \includegraphics[width=\textwidth]{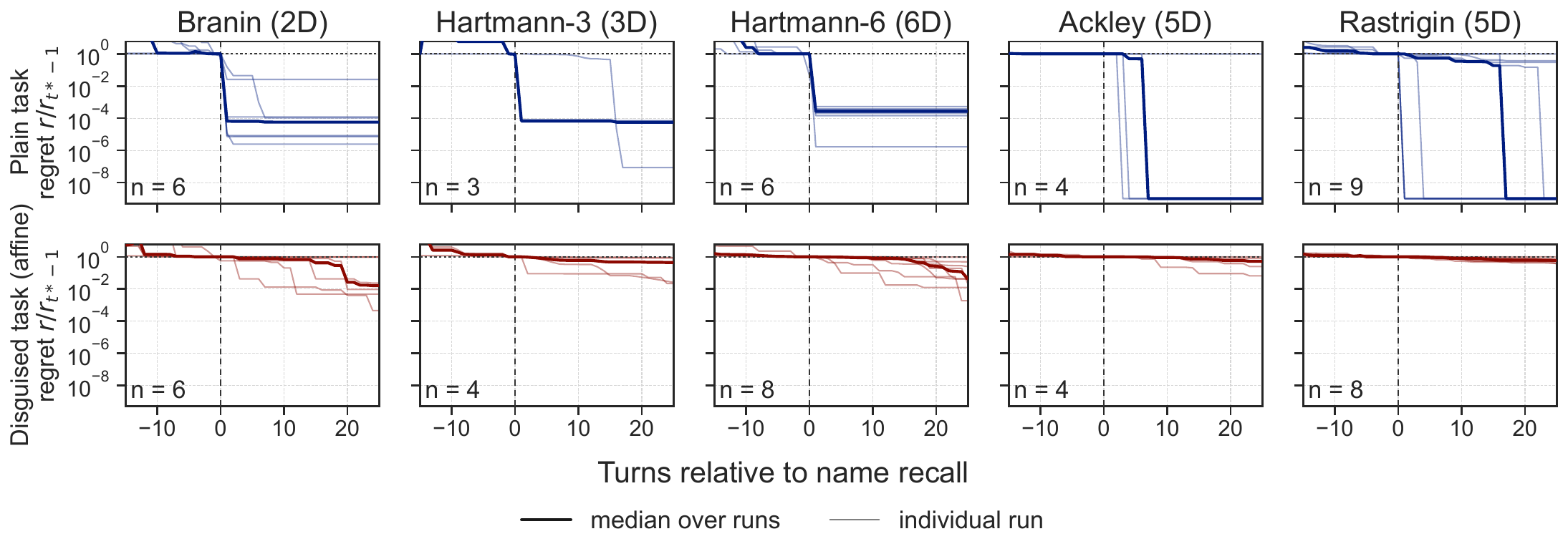}
    \caption{The ratio of the regret at a given offset to the regret at the turn immediately preceding the recall event. The top row is the plain variation of the five test functions and the bottom row is the disguised variation, where the disguised variation applies both the domain bijection and the range affine transformation. The lowercase $n$ indicates the number of recall events detected from analysis of the model's reasoning traces.}
    \label{fig:mem_breakdown}
\end{figure}

\section{Results on PMO evaluations}
\subsection{Comparison to non-reproducible methods}
\label{app:non-repro-comparison}
Tables~\ref{tab:cmp-lico} and~\ref{tab:cmp-gpmolformer} compare the Opus 5 multi-turn model performance in terms of Top-10 AUC against both LICO and GP-MoLFormer-SIM. Because these methods did not provide functional code or complete data on their optimization trajectories, we compare the performance of the LLM at a budget of 210 to those methods' performance at respective budgets of 1000 for LICO and 10000 for GP-MoLFormer-SIM. We note that on average across the 23 PMO oracles, the LLM achieves 66\% of the performance of LICO with only 21\% of the oracle budget, and 64\% of the performance of GP-MoLFormer-SIM with only 2.1\% of the oracle budget, again highlighting the strength of frontier reasoning LLMs as priors for optimization.

\begin{table}[!h]
\centering
\small
\setlength{\tabcolsep}{6pt}
\caption{Comparison of Opus 5 multi-turn performance to LICO in terms of Top-10 AUC. The error bars are $\pm$1 standard deviation over the five seeds.}
\label{tab:cmp-lico}
\begin{tabular}{@{}l r r r@{}}
\toprule
Oracle & Opus 5 multi-turn & LICO & Opus 5 / LICO (\%) \\
\midrule
albuterol\_similarity & $ 0.739 \pm 0.045 $ & $ 0.885 \pm 0.019 $ & 83.5 \\
amlodipine\_mpo & $ 0.485 \pm 0.023 $ & $ 0.679 \pm 0.027 $ & 71.5 \\
celecoxib\_rediscovery & $ 0.594 \pm 0.169 $ & $ 0.664 \pm 0.122 $ & 89.5 \\
deco\_hop & $ 0.583 \pm 0.059 $ & $ 0.619 \pm 0.015 $ & 94.1 \\
drd2 & $ 0.524 \pm 0.204 $ & $ 0.928 \pm 0.018 $ & 56.5 \\
fexofenadine\_mpo & $ 0.680 \pm 0.031 $ & $ 0.772 \pm 0.023 $ & 88.1 \\
gsk3b & $ 0.609 \pm 0.174 $ & $ 0.876 \pm 0.045 $ & 69.5 \\
isomers\_c7h8n2o2 & $ 0.158 \pm 0.100 $ & $ 0.939 \pm 0.022 $ & 16.8 \\
isomers\_c9h10n2o2pf2cl & $ 0.097 \pm 0.082 $ & $ 0.819 \pm 0.039 $ & 11.8 \\
jnk3 & $ 0.308 \pm 0.133 $ & $ 0.731 \pm 0.037 $ & 42.1 \\
median1 & $ 0.183 \pm 0.030 $ & $ 0.291 \pm 0.016 $ & 62.9 \\
median2 & $ 0.195 \pm 0.028 $ & $ 0.280 \pm 0.019 $ & 69.8 \\
mestranol\_similarity & $ 0.345 \pm 0.049 $ & $ 0.614 \pm 0.064 $ & 56.1 \\
osimertinib\_mpo & $ 0.722 \pm 0.032 $ & $ 0.820 \pm 0.012 $ & 88.0 \\
perindopril\_mpo & $ 0.436 \pm 0.014 $ & $ 0.557 \pm 0.028 $ & 78.2 \\
qed & $ 0.712 \pm 0.043 $ & $ 0.936 \pm 0.001 $ & 76.1 \\
ranolazine\_mpo & $ 0.606 \pm 0.026 $ & $ 0.774 \pm 0.008 $ & 78.3 \\
scaffold\_hop & $ 0.456 \pm 0.019 $ & $ 0.547 \pm 0.026 $ & 83.4 \\
sitagliptin\_mpo & $ 0.053 \pm 0.026 $ & $ 0.567 \pm 0.034 $ & 9.4 \\
thiothixene\_rediscovery & $ 0.377 \pm 0.024 $ & $ 0.514 \pm 0.037 $ & 73.3 \\
troglitazone\_rediscovery & $ 0.394 \pm 0.217 $ & $ 0.380 \pm 0.026 $ & 103.8 \\
valsartan\_smarts & $ 0.363 \pm 0.117 $ & $ 0.000 \pm 0.000 $ & -- \\
zaleplon\_mpo & $ 0.084 \pm 0.050 $ & $ 0.515 \pm 0.017 $ & 16.3 \\
\midrule
Sum (23 oracles) & $ 9.701 $ & $ 14.707 $ & 66.0 \\
\bottomrule
\end{tabular}
\end{table}

\begin{table}[!h]
\centering
\small
\setlength{\tabcolsep}{6pt}
\caption{Comparison of Opus 5 multi-turn performance to GP-MoLFormer-SIM in terms of Top-10 AUC. The error bars are $\pm$1 standard deviation over the five seeds.}
\label{tab:cmp-gpmolformer}
\begin{tabular}{@{}l r r r@{}}
\toprule
Oracle & Opus 5 multi-turn & GP-MoLFormer-SIM & Opus 5 / GP-MoLFormer-SIM (\%) \\
\midrule
albuterol\_similarity & $ 0.739 \pm 0.045 $ & $ 0.824 \pm 0.071 $ & 89.6 \\
amlodipine\_mpo & $ 0.485 \pm 0.023 $ & $ 0.680 \pm 0.064 $ & 71.4 \\
celecoxib\_rediscovery & $ 0.594 \pm 0.169 $ & $ 0.716 \pm 0.067 $ & 83.0 \\
deco\_hop & $ 0.583 \pm 0.059 $ & $ 0.710 \pm 0.058 $ & 82.1 \\
drd2 & $ 0.524 \pm 0.204 $ & $ 0.956 \pm 0.010 $ & 54.8 \\
fexofenadine\_mpo & $ 0.680 \pm 0.031 $ & $ 0.798 \pm 0.028 $ & 85.2 \\
gsk3b & $ 0.609 \pm 0.174 $ & $ 0.896 \pm 0.035 $ & 67.9 \\
isomers\_c7h8n2o2 & $ 0.158 \pm 0.100 $ & $ 0.932 \pm 0.011 $ & 16.9 \\
isomers\_c9h10n2o2pf2cl & $ 0.097 \pm 0.082 $ & $ 0.864 \pm 0.016 $ & 11.2 \\
jnk3 & $ 0.308 \pm 0.133 $ & $ 0.806 \pm 0.087 $ & 38.2 \\
median1 & $ 0.183 \pm 0.030 $ & $ 0.340 \pm 0.034 $ & 53.8 \\
median2 & $ 0.195 \pm 0.028 $ & $ 0.255 \pm 0.031 $ & 76.6 \\
mestranol\_similarity & $ 0.345 \pm 0.049 $ & $ 0.658 \pm 0.118 $ & 52.4 \\
osimertinib\_mpo & $ 0.722 \pm 0.032 $ & $ 0.819 \pm 0.004 $ & 88.1 \\
perindopril\_mpo & $ 0.436 \pm 0.014 $ & $ 0.584 \pm 0.042 $ & 74.6 \\
qed & $ 0.712 \pm 0.043 $ & $ 0.940 \pm 0.001 $ & 75.7 \\
ranolazine\_mpo & $ 0.606 \pm 0.026 $ & $ 0.812 \pm 0.024 $ & 74.6 \\
scaffold\_hop & $ 0.456 \pm 0.019 $ & $ 0.531 \pm 0.016 $ & 85.9 \\
sitagliptin\_mpo & $ 0.053 \pm 0.026 $ & $ 0.501 \pm 0.081 $ & 10.6 \\
thiothixene\_rediscovery & $ 0.377 \pm 0.024 $ & $ 0.504 \pm 0.033 $ & 74.8 \\
troglitazone\_rediscovery & $ 0.394 \pm 0.217 $ & $ 0.437 \pm 0.067 $ & 90.3 \\
valsartan\_smarts & $ 0.363 \pm 0.117 $ & $ 0.158 \pm 0.317 $ & 229.6 \\
zaleplon\_mpo & $ 0.084 \pm 0.050 $ & $ 0.504 \pm 0.022 $ & 16.6 \\
\midrule
Sum (23 oracles) & $ 9.701 $ & $ 15.225 $ & 63.7 \\
\bottomrule
\end{tabular}
\end{table}

\subsection{Comparison between LLMs on PMO tasks}\label{app:inter_llm_on_pmo}

Tables~\ref{tab:frontier-pmo-auc} and~\ref{tab:frontier-pmo-mean} report the full per-oracle comparison of the three frontier LLMs (each in single- and multi-turn mode) across all 23 PMO oracles---the complete, tabular version of Figure~\ref{fig:inter_llm_comp_pmo}.

\begin{table*}[htbp]
\centering
\scriptsize
\setlength{\tabcolsep}{4pt}
\caption{\textbf{Top-10 AUC at a 210-call budget on all 23 PMO oracles.} Full-benchmark, tabular form of Figure~\ref{fig:inter_llm_comp_pmo} (top row): the three frontier LLMs, each in single- and multi-turn mode, scored as the area under the running Top-10 mean curve out to 210 oracle calls. Mean $\pm$1 SE over five seeds; higher is better; \textbf{bold} = best arm for that oracle. The final row counts, for each arm, how many of the 23 oracles it wins (co-winners on an exact tie each count).}
\label{tab:frontier-pmo-auc}
\begin{tabular}{@{}l *{6}{c}@{}}
\toprule
Oracle & \multicolumn{2}{c}{Sonnet 4.6} & \multicolumn{2}{c}{Opus 4.8} & \multicolumn{2}{c}{Opus 5} \\
\cmidrule(lr){2-3}\cmidrule(lr){4-5}\cmidrule(lr){6-7}
 & single-turn & multi-turn & single-turn & multi-turn & single-turn & multi-turn \\
\midrule
albuterol\_similarity & $ 0.136 \pm 0.025 $ & $ 0.460 \pm 0.022 $ & $ 0.449 \pm 0.026 $ & $ 0.355 \pm 0.060 $ & $ 0.560 \pm 0.054 $ & {\boldmath$ 0.739 \pm 0.020 $} \\
amlodipine\_mpo & $ 0.308 \pm 0.038 $ & $ 0.447 \pm 0.005 $ & $ 0.457 \pm 0.008 $ & $ 0.471 \pm 0.010 $ & $ 0.473 \pm 0.009 $ & {\boldmath$ 0.485 \pm 0.010 $} \\
celecoxib\_rediscovery & $ 0.106 \pm 0.010 $ & $ 0.386 \pm 0.009 $ & $ 0.395 \pm 0.021 $ & $ 0.401 \pm 0.033 $ & $ 0.487 \pm 0.036 $ & {\boldmath$ 0.594 \pm 0.076 $} \\
deco\_hop & $ 0.424 \pm 0.050 $ & $ 0.560 \pm 0.006 $ & $ 0.551 \pm 0.005 $ & $ 0.540 \pm 0.012 $ & $ 0.556 \pm 0.006 $ & {\boldmath$ 0.583 \pm 0.026 $} \\
drd2 & $ 0.227 \pm 0.103 $ & $ 0.337 \pm 0.121 $ & $ 0.320 \pm 0.094 $ & $ 0.393 \pm 0.116 $ & $ 0.288 \pm 0.119 $ & {\boldmath$ 0.524 \pm 0.091 $} \\
fexofenadine\_mpo & $ 0.343 \pm 0.093 $ & $ 0.633 \pm 0.013 $ & $ 0.642 \pm 0.013 $ & $ 0.624 \pm 0.017 $ & $ 0.654 \pm 0.018 $ & {\boldmath$ 0.680 \pm 0.014 $} \\
gsk3b & $ 0.088 \pm 0.052 $ & $ 0.296 \pm 0.044 $ & $ 0.266 \pm 0.037 $ & $ 0.275 \pm 0.051 $ & $ 0.356 \pm 0.077 $ & {\boldmath$ 0.609 \pm 0.078 $} \\
isomers\_c7h8n2o2 & $ 0.005 \pm 0.003 $ & $ 0.158 \pm 0.078 $ & $ 0.052 \pm 0.032 $ & {\boldmath$ 0.210 \pm 0.058 $} & $ 0.101 \pm 0.036 $ & $ 0.158 \pm 0.045 $ \\
isomers\_c9h10n2o2pf2cl & $ 0.005 \pm 0.002 $ & {\boldmath$ 0.122 \pm 0.044 $} & $ 0.043 \pm 0.025 $ & $ 0.085 \pm 0.036 $ & $ 0.089 \pm 0.037 $ & $ 0.097 \pm 0.037 $ \\
jnk3 & $ 0.066 \pm 0.042 $ & $ 0.181 \pm 0.042 $ & $ 0.253 \pm 0.042 $ & $ 0.179 \pm 0.040 $ & $ 0.212 \pm 0.042 $ & {\boldmath$ 0.308 \pm 0.060 $} \\
median1 & $ 0.059 \pm 0.018 $ & $ 0.152 \pm 0.008 $ & $ 0.141 \pm 0.012 $ & $ 0.150 \pm 0.011 $ & $ 0.165 \pm 0.012 $ & {\boldmath$ 0.183 \pm 0.013 $} \\
median2 & $ 0.064 \pm 0.024 $ & $ 0.182 \pm 0.008 $ & $ 0.190 \pm 0.012 $ & $ 0.185 \pm 0.011 $ & {\boldmath$ 0.195 \pm 0.011 $} & $ 0.195 \pm 0.013 $ \\
mestranol\_similarity & $ 0.126 \pm 0.038 $ & $ 0.271 \pm 0.061 $ & $ 0.318 \pm 0.024 $ & $ 0.313 \pm 0.023 $ & $ 0.314 \pm 0.018 $ & {\boldmath$ 0.345 \pm 0.022 $} \\
osimertinib\_mpo & $ 0.395 \pm 0.112 $ & $ 0.704 \pm 0.016 $ & $ 0.708 \pm 0.015 $ & $ 0.676 \pm 0.028 $ & $ 0.705 \pm 0.019 $ & {\boldmath$ 0.722 \pm 0.014 $} \\
perindopril\_mpo & $ 0.180 \pm 0.068 $ & $ 0.431 \pm 0.010 $ & $ 0.392 \pm 0.041 $ & $ 0.367 \pm 0.056 $ & $ 0.427 \pm 0.006 $ & {\boldmath$ 0.436 \pm 0.006 $} \\
qed & $ 0.288 \pm 0.060 $ & $ 0.860 \pm 0.011 $ & $ 0.637 \pm 0.127 $ & $ 0.625 \pm 0.067 $ & {\boldmath$ 0.886 \pm 0.006 $} & $ 0.712 \pm 0.019 $ \\
ranolazine\_mpo & $ 0.241 \pm 0.093 $ & $ 0.538 \pm 0.037 $ & $ 0.495 \pm 0.046 $ & $ 0.506 \pm 0.035 $ & $ 0.590 \pm 0.024 $ & {\boldmath$ 0.606 \pm 0.011 $} \\
scaffold\_hop & $ 0.378 \pm 0.052 $ & $ 0.356 \pm 0.068 $ & $ 0.433 \pm 0.009 $ & $ 0.433 \pm 0.010 $ & $ 0.393 \pm 0.041 $ & {\boldmath$ 0.456 \pm 0.009 $} \\
sitagliptin\_mpo & $ 0.012 \pm 0.003 $ & $ 0.062 \pm 0.024 $ & $ 0.048 \pm 0.024 $ & $ 0.072 \pm 0.019 $ & {\boldmath$ 0.098 \pm 0.030 $} & $ 0.053 \pm 0.012 $ \\
thiothixene\_rediscovery & $ 0.113 \pm 0.035 $ & $ 0.280 \pm 0.014 $ & $ 0.257 \pm 0.027 $ & $ 0.262 \pm 0.009 $ & $ 0.294 \pm 0.005 $ & {\boldmath$ 0.377 \pm 0.011 $} \\
troglitazone\_rediscovery & $ 0.082 \pm 0.009 $ & $ 0.266 \pm 0.005 $ & $ 0.216 \pm 0.023 $ & $ 0.232 \pm 0.015 $ & $ 0.266 \pm 0.008 $ & {\boldmath$ 0.394 \pm 0.097 $} \\
valsartan\_smarts & $ 0.000 \pm 0.000 $ & $ 0.000 \pm 0.000 $ & $ 0.000 \pm 0.000 $ & $ 0.000 \pm 0.000 $ & $ 0.000 \pm 0.000 $ & {\boldmath$ 0.363 \pm 0.052 $} \\
zaleplon\_mpo & $ 0.042 \pm 0.012 $ & $ 0.086 \pm 0.012 $ & $ 0.084 \pm 0.027 $ & {\boldmath$ 0.136 \pm 0.014 $} & $ 0.106 \pm 0.032 $ & $ 0.084 \pm 0.022 $ \\
\midrule
Mean (23 oracles) & $ 0.160 $ & $ 0.338 $ & $ 0.319 $ & $ 0.326 $ & $ 0.357 $ & $ 0.422 $ \\
Oracles won (best of 23) & 0 & 1 & 0 & 2 & 3 & 17 \\
\bottomrule
\end{tabular}
\end{table*}

\begin{table*}[htbp]
\centering
\scriptsize
\setlength{\tabcolsep}{4pt}
\caption{\textbf{Top-10 mean at a 210-call budget on all 23 PMO oracles.} Full-benchmark, tabular form of Figure~\ref{fig:inter_llm_comp_pmo} (bottom row): the mean oracle score of the 10 best molecules found within 210 oracle calls, for the three frontier LLMs in single- and multi-turn mode. Mean $\pm$1 SE over five seeds; higher is better; \textbf{bold} = best arm for that oracle. The final row counts, for each arm, how many of the 23 oracles it wins (co-winners on an exact tie each count).}
\label{tab:frontier-pmo-mean}
\begin{tabular}{@{}l *{6}{c}@{}}
\toprule
Oracle & \multicolumn{2}{c}{Sonnet 4.6} & \multicolumn{2}{c}{Opus 4.8} & \multicolumn{2}{c}{Opus 5} \\
\cmidrule(lr){2-3}\cmidrule(lr){4-5}\cmidrule(lr){6-7}
 & single-turn & multi-turn & single-turn & multi-turn & single-turn & multi-turn \\
\midrule
albuterol\_similarity & $ 0.448 \pm 0.021 $ & $ 0.659 \pm 0.037 $ & $ 0.549 \pm 0.042 $ & $ 0.511 \pm 0.033 $ & $ 0.775 \pm 0.082 $ & {\boldmath$ 1.000 \pm 0.000 $} \\
amlodipine\_mpo & $ 0.515 \pm 0.011 $ & $ 0.524 \pm 0.011 $ & $ 0.515 \pm 0.011 $ & $ 0.554 \pm 0.010 $ & $ 0.547 \pm 0.009 $ & {\boldmath$ 0.564 \pm 0.011 $} \\
celecoxib\_rediscovery & $ 0.363 \pm 0.014 $ & $ 0.537 \pm 0.041 $ & $ 0.476 \pm 0.027 $ & $ 0.578 \pm 0.069 $ & $ 0.767 \pm 0.087 $ & {\boldmath$ 1.000 \pm 0.000 $} \\
deco\_hop & $ 0.585 \pm 0.005 $ & $ 0.605 \pm 0.004 $ & $ 0.587 \pm 0.006 $ & $ 0.602 \pm 0.009 $ & $ 0.597 \pm 0.007 $ & {\boldmath$ 0.661 \pm 0.056 $} \\
drd2 & $ 0.477 \pm 0.159 $ & $ 0.514 \pm 0.133 $ & $ 0.568 \pm 0.137 $ & $ 0.683 \pm 0.171 $ & $ 0.395 \pm 0.137 $ & {\boldmath$ 0.945 \pm 0.028 $} \\
fexofenadine\_mpo & $ 0.674 \pm 0.024 $ & $ 0.714 \pm 0.016 $ & $ 0.700 \pm 0.016 $ & $ 0.732 \pm 0.015 $ & $ 0.733 \pm 0.021 $ & {\boldmath$ 0.770 \pm 0.015 $} \\
gsk3b & $ 0.287 \pm 0.064 $ & $ 0.413 \pm 0.060 $ & $ 0.355 \pm 0.048 $ & $ 0.452 \pm 0.042 $ & $ 0.474 \pm 0.095 $ & {\boldmath$ 0.775 \pm 0.068 $} \\
isomers\_c7h8n2o2 & $ 0.036 \pm 0.024 $ & $ 0.375 \pm 0.149 $ & $ 0.119 \pm 0.060 $ & $ 0.487 \pm 0.135 $ & $ 0.404 \pm 0.147 $ & {\boldmath$ 0.522 \pm 0.120 $} \\
isomers\_c9h10n2o2pf2cl & $ 0.026 \pm 0.014 $ & $ 0.300 \pm 0.104 $ & $ 0.064 \pm 0.037 $ & $ 0.171 \pm 0.064 $ & $ 0.184 \pm 0.075 $ & {\boldmath$ 0.328 \pm 0.110 $} \\
jnk3 & $ 0.151 \pm 0.037 $ & $ 0.247 \pm 0.048 $ & $ 0.366 \pm 0.087 $ & $ 0.245 \pm 0.057 $ & $ 0.290 \pm 0.052 $ & {\boldmath$ 0.495 \pm 0.098 $} \\
median1 & $ 0.158 \pm 0.009 $ & $ 0.193 \pm 0.010 $ & $ 0.162 \pm 0.015 $ & $ 0.196 \pm 0.016 $ & $ 0.212 \pm 0.022 $ & {\boldmath$ 0.260 \pm 0.029 $} \\
median2 & $ 0.200 \pm 0.010 $ & $ 0.228 \pm 0.009 $ & $ 0.213 \pm 0.014 $ & $ 0.228 \pm 0.010 $ & $ 0.228 \pm 0.010 $ & {\boldmath$ 0.239 \pm 0.016 $} \\
mestranol\_similarity & $ 0.322 \pm 0.018 $ & $ 0.362 \pm 0.051 $ & $ 0.372 \pm 0.033 $ & $ 0.410 \pm 0.036 $ & $ 0.377 \pm 0.022 $ & {\boldmath$ 0.439 \pm 0.034 $} \\
osimertinib\_mpo & $ 0.741 \pm 0.024 $ & $ 0.781 \pm 0.022 $ & $ 0.771 \pm 0.021 $ & $ 0.802 \pm 0.022 $ & $ 0.781 \pm 0.026 $ & {\boldmath$ 0.830 \pm 0.015 $} \\
perindopril\_mpo & $ 0.444 \pm 0.031 $ & $ 0.501 \pm 0.007 $ & $ 0.487 \pm 0.015 $ & $ 0.488 \pm 0.020 $ & $ 0.488 \pm 0.004 $ & {\boldmath$ 0.535 \pm 0.017 $} \\
qed & $ 0.916 \pm 0.012 $ & {\boldmath$ 0.945 \pm 0.002 $} & $ 0.917 \pm 0.011 $ & $ 0.935 \pm 0.008 $ & $ 0.939 \pm 0.007 $ & $ 0.729 \pm 0.020 $ \\
ranolazine\_mpo & $ 0.595 \pm 0.044 $ & $ 0.767 \pm 0.010 $ & $ 0.628 \pm 0.040 $ & $ 0.759 \pm 0.022 $ & $ 0.749 \pm 0.018 $ & {\boldmath$ 0.781 \pm 0.012 $} \\
scaffold\_hop & $ 0.471 \pm 0.008 $ & $ 0.476 \pm 0.009 $ & $ 0.468 \pm 0.011 $ & $ 0.482 \pm 0.011 $ & $ 0.478 \pm 0.011 $ & {\boldmath$ 0.512 \pm 0.011 $} \\
sitagliptin\_mpo & $ 0.065 \pm 0.015 $ & $ 0.158 \pm 0.039 $ & $ 0.082 \pm 0.038 $ & {\boldmath$ 0.255 \pm 0.042 $} & $ 0.213 \pm 0.055 $ & $ 0.147 \pm 0.037 $ \\
thiothixene\_rediscovery & $ 0.327 \pm 0.032 $ & $ 0.359 \pm 0.009 $ & $ 0.286 \pm 0.034 $ & $ 0.331 \pm 0.012 $ & $ 0.382 \pm 0.009 $ & {\boldmath$ 0.503 \pm 0.030 $} \\
troglitazone\_rediscovery & $ 0.264 \pm 0.007 $ & $ 0.336 \pm 0.010 $ & $ 0.284 \pm 0.012 $ & $ 0.303 \pm 0.012 $ & $ 0.437 \pm 0.108 $ & {\boldmath$ 0.578 \pm 0.107 $} \\
valsartan\_smarts & $ 0.000 \pm 0.000 $ & $ 0.000 \pm 0.000 $ & $ 0.000 \pm 0.000 $ & $ 0.000 \pm 0.000 $ & $ 0.000 \pm 0.000 $ & {\boldmath$ 0.891 \pm 0.026 $} \\
zaleplon\_mpo & $ 0.131 \pm 0.025 $ & $ 0.190 \pm 0.021 $ & $ 0.150 \pm 0.047 $ & {\boldmath$ 0.234 \pm 0.018 $} & $ 0.220 \pm 0.034 $ & $ 0.212 \pm 0.037 $ \\
\midrule
Mean (23 oracles) & $ 0.356 $ & $ 0.443 $ & $ 0.396 $ & $ 0.454 $ & $ 0.464 $ & $ 0.596 $ \\
Oracles won (best of 23) & 0 & 1 & 0 & 2 & 0 & 20 \\
\bottomrule
\end{tabular}
\end{table*}

\subsection{Full results for molecular optimization tasks in PMO}
\label{app:molec_optim}

Tables~\ref{tab:opus-panel-auc-a}--\ref{tab:opus-panel-fold-c} give the full comparison of Opus 5 multi-turn against the entire panel of PMO methods across all 23 oracles: the 25 published PMO specialists together with GenMol, ExLLM, MolLEO, and the five back-filled specialists (SMILES Augmented Memory, SMILES AHC, SMILES BAR, MolGAN, and REINVENT-Transformer). Because the 34-way panel is too wide for a single table, each metric is partitioned into three parts of 11 methods (ranked by macro-average Top-10 AUC, so a given part contains the same methods in every metric), with the Opus 5 multi-turn column repeated in each Top-10 AUC and Top-10 mean part. We report Top-10 AUC@210 (Tables~\ref{tab:opus-panel-auc-a}--\ref{tab:opus-panel-auc-c}), Top-10 mean@210 (Tables~\ref{tab:opus-panel-mean-a}--\ref{tab:opus-panel-mean-c}), and the fold oracle-call budget each method requires to reach Opus 5 multi-turn's Top-10 mean@210 (Tables~\ref{tab:opus-panel-fold-a}--\ref{tab:opus-panel-fold-c}). Error bars are $\pm$1 SE over seeds. For an example of the prompts used for the LLMs in the molecular generation tasks, see Prompt~\ref{lst:prompt_molgen}.


\begin{landscape}
\begin{table}[htbp]
\centering
\scriptsize
\setlength{\tabcolsep}{3pt}
\caption{\textbf{Top-10 AUC at a 210-call budget, Opus 5 multi-turn versus the full method panel (part a of 3).} Rows are the 23 PMO oracles; columns are Opus 5 multi-turn (reference, repeated in every part and set off by a vertical rule) and methods 1--11 of 33, ranked by macro-average Top-10 AUC. Each cell is mean $\pm$1 SE over seeds; higher is better; \textbf{bold} = best over the full 34-way panel for that oracle. The footer gives the per-column macro-average and the number of oracles won over the full panel.}
\label{tab:opus-panel-auc-a}
\begin{tabular}{@{}l c!{\vrule} *{11}{c}@{}}
\toprule
Oracle & \rotatebox{90}{Opus 5 (multi-turn)} & \rotatebox{90}{GenMol} & \rotatebox{90}{ExLLM (Gemini 2.5 Flash)} & \rotatebox{90}{MolLEO (BioT5)} & \rotatebox{90}{synnet} & \rotatebox{90}{dog\_gen} & \rotatebox{90}{dog\_ae} & \rotatebox{90}{graph\_ga} & \rotatebox{90}{SMILES Aug. Memory} & \rotatebox{90}{reinvent} & \rotatebox{90}{REINVENT-Transformer} & \rotatebox{90}{smiles\_lstm\_hc} \\
\midrule
albuterol\_similarity & \textbf{0.739}\,{\tiny$\pm$0.020} & 0.567\,{\tiny$\pm$0.006} & 0.638\,{\tiny$\pm$0.000} & 0.357\,{\tiny$\pm$0.003} & 0.341\,{\tiny$\pm$0.002} & 0.384\,{\tiny$\pm$0.000} & 0.368\,{\tiny$\pm$0.008} & 0.322\,{\tiny$\pm$0.000} & 0.344\,{\tiny$\pm$0.002} & 0.342\,{\tiny$\pm$0.002} & 0.350\,{\tiny$\pm$0.004} & 0.322\,{\tiny$\pm$0.001} \\
amlodipine\_mpo & 0.485\,{\tiny$\pm$0.010} & 0.424\,{\tiny$\pm$0.051} & \textbf{0.611}\,{\tiny$\pm$0.005} & 0.437\,{\tiny$\pm$0.004} & 0.417\,{\tiny$\pm$0.001} & 0.388\,{\tiny$\pm$0.000} & 0.398\,{\tiny$\pm$0.003} & 0.434\,{\tiny$\pm$0.001} & 0.416\,{\tiny$\pm$0.001} & 0.411\,{\tiny$\pm$0.000} & 0.412\,{\tiny$\pm$0.005} & 0.434\,{\tiny$\pm$0.001} \\
celecoxib\_rediscovery & \textbf{0.594}\,{\tiny$\pm$0.076} & 0.463\,{\tiny$\pm$0.007} & 0.493\,{\tiny$\pm$0.004} & 0.242\,{\tiny$\pm$0.004} & 0.246\,{\tiny$\pm$0.001} & 0.223\,{\tiny$\pm$0.000} & 0.253\,{\tiny$\pm$0.005} & 0.233\,{\tiny$\pm$0.001} & 0.230\,{\tiny$\pm$0.001} & 0.260\,{\tiny$\pm$0.000} & 0.232\,{\tiny$\pm$0.003} & 0.233\,{\tiny$\pm$0.001} \\
deco\_hop & 0.583\,{\tiny$\pm$0.026} & \textbf{0.793}\,{\tiny$\pm$0.008} & 0.691\,{\tiny$\pm$0.002} & 0.532\,{\tiny$\pm$0.001} & 0.542\,{\tiny$\pm$0.001} & 0.532\,{\tiny$\pm$0.000} & 0.538\,{\tiny$\pm$0.004} & 0.530\,{\tiny$\pm$0.000} & 0.531\,{\tiny$\pm$0.000} & 0.532\,{\tiny$\pm$0.000} & 0.527\,{\tiny$\pm$0.001} & 0.529\,{\tiny$\pm$0.000} \\
drd2 & 0.524\,{\tiny$\pm$0.091} & \textbf{0.924}\,{\tiny$\pm$0.004} & 0.344\,{\tiny$\pm$0.018} & 0.164\,{\tiny$\pm$0.012} & 0.196\,{\tiny$\pm$0.018} & 0.261\,{\tiny$\pm$0.000} & 0.160\,{\tiny$\pm$0.021} & 0.226\,{\tiny$\pm$0.017} & 0.136\,{\tiny$\pm$0.014} & 0.074\,{\tiny$\pm$0.001} & 0.069\,{\tiny$\pm$0.008} & 0.050\,{\tiny$\pm$0.001} \\
fexofenadine\_mpo & 0.680\,{\tiny$\pm$0.014} & \textbf{0.714}\,{\tiny$\pm$0.003} & 0.650\,{\tiny$\pm$0.002} & 0.555\,{\tiny$\pm$0.007} & 0.588\,{\tiny$\pm$0.001} & 0.498\,{\tiny$\pm$0.000} & 0.510\,{\tiny$\pm$0.012} & 0.549\,{\tiny$\pm$0.000} & 0.550\,{\tiny$\pm$0.001} & 0.537\,{\tiny$\pm$0.002} & 0.539\,{\tiny$\pm$0.007} & 0.549\,{\tiny$\pm$0.001} \\
gsk3b & 0.609\,{\tiny$\pm$0.078} & \textbf{0.782}\,{\tiny$\pm$0.006} & 0.176\,{\tiny$\pm$0.016} & 0.194\,{\tiny$\pm$0.024} & 0.160\,{\tiny$\pm$0.003} & 0.240\,{\tiny$\pm$0.000} & 0.217\,{\tiny$\pm$0.006} & 0.148\,{\tiny$\pm$0.011} & 0.114\,{\tiny$\pm$0.004} & 0.086\,{\tiny$\pm$0.005} & 0.132\,{\tiny$\pm$0.011} & 0.143\,{\tiny$\pm$0.005} \\
isomers\_c7h8n2o2 & 0.158\,{\tiny$\pm$0.045} & 0.197\,{\tiny$\pm$0.026} & \textbf{0.502}\,{\tiny$\pm$0.000} & 0.181\,{\tiny$\pm$0.019} & 0.014\,{\tiny$\pm$0.001} & 0.028\,{\tiny$\pm$0.000} & 0.038\,{\tiny$\pm$0.006} & 0.020\,{\tiny$\pm$0.007} & 0.016\,{\tiny$\pm$0.001} & 0.013\,{\tiny$\pm$0.001} & 0.008\,{\tiny$\pm$0.003} & 0.012\,{\tiny$\pm$0.000} \\
isomers\_c9h10n2o2pf2cl & 0.097\,{\tiny$\pm$0.037} & 0.183\,{\tiny$\pm$0.006} & \textbf{0.551}\,{\tiny$\pm$0.002} & 0.249\,{\tiny$\pm$0.014} & 0.003\,{\tiny$\pm$0.000} & 0.009\,{\tiny$\pm$0.000} & 0.010\,{\tiny$\pm$0.001} & 0.024\,{\tiny$\pm$0.008} & 0.007\,{\tiny$\pm$0.001} & 0.007\,{\tiny$\pm$0.001} & 0.004\,{\tiny$\pm$0.001} & 0.015\,{\tiny$\pm$0.000} \\
jnk3 & 0.308\,{\tiny$\pm$0.060} & \textbf{0.653}\,{\tiny$\pm$0.016} & 0.493\,{\tiny$\pm$0.002} & 0.086\,{\tiny$\pm$0.008} & 0.113\,{\tiny$\pm$0.002} & 0.135\,{\tiny$\pm$0.000} & 0.124\,{\tiny$\pm$0.007} & 0.071\,{\tiny$\pm$0.003} & 0.077\,{\tiny$\pm$0.003} & 0.073\,{\tiny$\pm$0.002} & 0.069\,{\tiny$\pm$0.003} & 0.069\,{\tiny$\pm$0.001} \\
median1 & 0.183\,{\tiny$\pm$0.013} & \textbf{0.275}\,{\tiny$\pm$0.004} & 0.197\,{\tiny$\pm$0.001} & 0.130\,{\tiny$\pm$0.003} & 0.128\,{\tiny$\pm$0.002} & 0.108\,{\tiny$\pm$0.000} & 0.109\,{\tiny$\pm$0.004} & 0.115\,{\tiny$\pm$0.001} & 0.126\,{\tiny$\pm$0.000} & 0.131\,{\tiny$\pm$0.001} & 0.132\,{\tiny$\pm$0.002} & 0.116\,{\tiny$\pm$0.000} \\
median2 & 0.195\,{\tiny$\pm$0.013} & \textbf{0.285}\,{\tiny$\pm$0.004} & 0.253\,{\tiny$\pm$0.002} & 0.146\,{\tiny$\pm$0.001} & 0.150\,{\tiny$\pm$0.001} & 0.131\,{\tiny$\pm$0.000} & 0.136\,{\tiny$\pm$0.002} & 0.143\,{\tiny$\pm$0.000} & 0.148\,{\tiny$\pm$0.000} & 0.135\,{\tiny$\pm$0.000} & 0.143\,{\tiny$\pm$0.001} & 0.143\,{\tiny$\pm$0.000} \\
mestranol\_similarity & 0.345\,{\tiny$\pm$0.022} & \textbf{0.535}\,{\tiny$\pm$0.013} & 0.500\,{\tiny$\pm$0.008} & 0.273\,{\tiny$\pm$0.007} & 0.249\,{\tiny$\pm$0.001} & 0.242\,{\tiny$\pm$0.000} & 0.246\,{\tiny$\pm$0.006} & 0.244\,{\tiny$\pm$0.001} & 0.249\,{\tiny$\pm$0.001} & 0.249\,{\tiny$\pm$0.000} & 0.271\,{\tiny$\pm$0.006} & 0.247\,{\tiny$\pm$0.002} \\
osimertinib\_mpo & 0.722\,{\tiny$\pm$0.014} & \textbf{0.787}\,{\tiny$\pm$0.003} & 0.460\,{\tiny$\pm$0.010} & 0.595\,{\tiny$\pm$0.012} & 0.531\,{\tiny$\pm$0.003} & 0.574\,{\tiny$\pm$0.000} & 0.496\,{\tiny$\pm$0.014} & 0.607\,{\tiny$\pm$0.002} & 0.590\,{\tiny$\pm$0.001} & 0.629\,{\tiny$\pm$0.000} & 0.594\,{\tiny$\pm$0.007} & 0.605\,{\tiny$\pm$0.000} \\
perindopril\_mpo & 0.436\,{\tiny$\pm$0.006} & \textbf{0.550}\,{\tiny$\pm$0.007} & 0.503\,{\tiny$\pm$0.006} & 0.399\,{\tiny$\pm$0.005} & 0.362\,{\tiny$\pm$0.001} & 0.335\,{\tiny$\pm$0.000} & 0.314\,{\tiny$\pm$0.006} & 0.353\,{\tiny$\pm$0.001} & 0.340\,{\tiny$\pm$0.000} & 0.331\,{\tiny$\pm$0.001} & 0.348\,{\tiny$\pm$0.001} & 0.353\,{\tiny$\pm$0.001} \\
qed & 0.712\,{\tiny$\pm$0.019} & 0.850\,{\tiny$\pm$0.003} & \textbf{0.891}\,{\tiny$\pm$0.000} & 0.855\,{\tiny$\pm$0.003} & 0.870\,{\tiny$\pm$0.000} & 0.812\,{\tiny$\pm$0.000} & 0.816\,{\tiny$\pm$0.006} & 0.850\,{\tiny$\pm$0.002} & 0.867\,{\tiny$\pm$0.000} & 0.844\,{\tiny$\pm$0.000} & 0.857\,{\tiny$\pm$0.003} & 0.855\,{\tiny$\pm$0.000} \\
ranolazine\_mpo & 0.606\,{\tiny$\pm$0.011} & \textbf{0.637}\,{\tiny$\pm$0.003} & 0.417\,{\tiny$\pm$0.006} & 0.245\,{\tiny$\pm$0.016} & 0.374\,{\tiny$\pm$0.003} & 0.293\,{\tiny$\pm$0.000} & 0.375\,{\tiny$\pm$0.009} & 0.198\,{\tiny$\pm$0.006} & 0.229\,{\tiny$\pm$0.011} & 0.294\,{\tiny$\pm$0.003} & 0.224\,{\tiny$\pm$0.015} & 0.200\,{\tiny$\pm$0.007} \\
scaffold\_hop & 0.456\,{\tiny$\pm$0.009} & 0.497\,{\tiny$\pm$0.003} & \textbf{0.580}\,{\tiny$\pm$0.005} & 0.404\,{\tiny$\pm$0.002} & 0.417\,{\tiny$\pm$0.001} & 0.404\,{\tiny$\pm$0.000} & 0.408\,{\tiny$\pm$0.002} & 0.400\,{\tiny$\pm$0.000} & 0.400\,{\tiny$\pm$0.000} & 0.403\,{\tiny$\pm$0.000} & 0.397\,{\tiny$\pm$0.001} & 0.399\,{\tiny$\pm$0.000} \\
sitagliptin\_mpo & 0.053\,{\tiny$\pm$0.012} & \textbf{0.181}\,{\tiny$\pm$0.011} & 0.098\,{\tiny$\pm$0.005} & 0.096\,{\tiny$\pm$0.013} & 0.002\,{\tiny$\pm$0.000} & 0.001\,{\tiny$\pm$0.000} & 0.001\,{\tiny$\pm$0.000} & 0.002\,{\tiny$\pm$0.001} & 0.027\,{\tiny$\pm$0.002} & 0.001\,{\tiny$\pm$0.000} & 0.028\,{\tiny$\pm$0.002} & 0.001\,{\tiny$\pm$0.000} \\
thiothixene\_rediscovery & 0.377\,{\tiny$\pm$0.011} & 0.392\,{\tiny$\pm$0.007} & \textbf{0.505}\,{\tiny$\pm$0.006} & 0.233\,{\tiny$\pm$0.003} & 0.253\,{\tiny$\pm$0.001} & 0.210\,{\tiny$\pm$0.000} & 0.223\,{\tiny$\pm$0.004} & 0.229\,{\tiny$\pm$0.001} & 0.239\,{\tiny$\pm$0.000} & 0.219\,{\tiny$\pm$0.000} & 0.231\,{\tiny$\pm$0.002} & 0.229\,{\tiny$\pm$0.001} \\
troglitazone\_rediscovery & 0.394\,{\tiny$\pm$0.097} & \textbf{0.465}\,{\tiny$\pm$0.012} & 0.310\,{\tiny$\pm$0.001} & 0.190\,{\tiny$\pm$0.002} & 0.185\,{\tiny$\pm$0.001} & 0.178\,{\tiny$\pm$0.000} & 0.180\,{\tiny$\pm$0.004} & 0.186\,{\tiny$\pm$0.001} & 0.187\,{\tiny$\pm$0.000} & 0.178\,{\tiny$\pm$0.000} & 0.184\,{\tiny$\pm$0.004} & 0.186\,{\tiny$\pm$0.000} \\
valsartan\_smarts & \textbf{0.363}\,{\tiny$\pm$0.052} & 0.015\,{\tiny$\pm$0.004} & 0.002\,{\tiny$\pm$0.001} & 0.000\,{\tiny$\pm$0.000} & 0.000\,{\tiny$\pm$0.000} & 0.000\,{\tiny$\pm$0.000} & 0.000\,{\tiny$\pm$0.000} & 0.000\,{\tiny$\pm$0.000} & 0.000\,{\tiny$\pm$0.000} & 0.002\,{\tiny$\pm$0.002} & 0.000\,{\tiny$\pm$0.000} & 0.000\,{\tiny$\pm$0.000} \\
zaleplon\_mpo & 0.084\,{\tiny$\pm$0.022} & 0.320\,{\tiny$\pm$0.004} & \textbf{0.476}\,{\tiny$\pm$0.003} & 0.274\,{\tiny$\pm$0.006} & 0.007\,{\tiny$\pm$0.002} & 0.001\,{\tiny$\pm$0.000} & 0.000\,{\tiny$\pm$0.000} & 0.010\,{\tiny$\pm$0.004} & 0.001\,{\tiny$\pm$0.000} & 0.001\,{\tiny$\pm$0.000} & 0.001\,{\tiny$\pm$0.000} & 0.006\,{\tiny$\pm$0.000} \\
\midrule
Mean & $0.422$ & $0.500$ & $0.450$ & $0.297$ & $0.267$ & $0.260$ & $0.257$ & $0.256$ & $0.253$ & $0.250$ & $0.250$ & $0.248$ \\
Oracles won & 3 & 13 & 7 & 0 & 0 & 0 & 0 & 0 & 0 & 0 & 0 & 0 \\
\bottomrule
\end{tabular}
\end{table}
\end{landscape}

\begin{landscape}
\begin{table}[htbp]
\centering
\scriptsize
\setlength{\tabcolsep}{3pt}
\caption{\textbf{Top-10 AUC at a 210-call budget, Opus 5 multi-turn versus the full method panel (part b of 3).} Rows are the 23 PMO oracles; columns are Opus 5 multi-turn (reference, repeated in every part and set off by a vertical rule) and methods 12--22 of 33, ranked by macro-average Top-10 AUC. Each cell is mean $\pm$1 SE over seeds; higher is better; \textbf{bold} = best over the full 34-way panel for that oracle. The footer gives the per-column macro-average and the number of oracles won over the full panel.}
\label{tab:opus-panel-auc-b}
\begin{tabular}{@{}l c!{\vrule} *{11}{c}@{}}
\toprule
Oracle & \rotatebox{90}{Opus 5 (multi-turn)} & \rotatebox{90}{smiles\_ga} & \rotatebox{90}{SMILES AHC} & \rotatebox{90}{gflownet} & \rotatebox{90}{pasithea} & \rotatebox{90}{stoned} & \rotatebox{90}{selfies\_lstm\_hc} & \rotatebox{90}{reinvent\_selfies} & \rotatebox{90}{mimosa} & \rotatebox{90}{gflownet\_al} & \rotatebox{90}{smiles\_vae\_bo} & \rotatebox{90}{jt\_vae\_bo} \\
\midrule
albuterol\_similarity & \textbf{0.739}\,{\tiny$\pm$0.020} & 0.321\,{\tiny$\pm$0.000} & 0.322\,{\tiny$\pm$0.001} & 0.244\,{\tiny$\pm$0.001} & 0.323\,{\tiny$\pm$0.000} & 0.322\,{\tiny$\pm$0.000} & 0.324\,{\tiny$\pm$0.002} & 0.349\,{\tiny$\pm$0.001} & 0.331\,{\tiny$\pm$0.001} & 0.260\,{\tiny$\pm$0.000} & 0.331\,{\tiny$\pm$0.001} & 0.333\,{\tiny$\pm$0.000} \\
amlodipine\_mpo & 0.485\,{\tiny$\pm$0.010} & 0.434\,{\tiny$\pm$0.001} & 0.415\,{\tiny$\pm$0.000} & 0.291\,{\tiny$\pm$0.002} & 0.432\,{\tiny$\pm$0.001} & 0.435\,{\tiny$\pm$0.001} & 0.434\,{\tiny$\pm$0.001} & 0.395\,{\tiny$\pm$0.001} & 0.426\,{\tiny$\pm$0.000} & 0.258\,{\tiny$\pm$0.000} & 0.426\,{\tiny$\pm$0.000} & 0.424\,{\tiny$\pm$0.001} \\
celecoxib\_rediscovery & \textbf{0.594}\,{\tiny$\pm$0.076} & 0.233\,{\tiny$\pm$0.001} & 0.222\,{\tiny$\pm$0.001} & 0.169\,{\tiny$\pm$0.001} & 0.218\,{\tiny$\pm$0.001} & 0.233\,{\tiny$\pm$0.001} & 0.232\,{\tiny$\pm$0.001} & 0.222\,{\tiny$\pm$0.001} & 0.216\,{\tiny$\pm$0.000} & 0.172\,{\tiny$\pm$0.001} & 0.216\,{\tiny$\pm$0.000} & 0.212\,{\tiny$\pm$0.003} \\
deco\_hop & 0.583\,{\tiny$\pm$0.026} & 0.529\,{\tiny$\pm$0.000} & 0.525\,{\tiny$\pm$0.000} & 0.531\,{\tiny$\pm$0.000} & 0.532\,{\tiny$\pm$0.000} & 0.530\,{\tiny$\pm$0.000} & 0.529\,{\tiny$\pm$0.000} & 0.520\,{\tiny$\pm$0.001} & 0.532\,{\tiny$\pm$0.001} & 0.533\,{\tiny$\pm$0.000} & 0.531\,{\tiny$\pm$0.000} & 0.526\,{\tiny$\pm$0.001} \\
drd2 & 0.524\,{\tiny$\pm$0.091} & 0.058\,{\tiny$\pm$0.008} & 0.108\,{\tiny$\pm$0.002} & 0.111\,{\tiny$\pm$0.004} & 0.060\,{\tiny$\pm$0.000} & 0.048\,{\tiny$\pm$0.000} & 0.048\,{\tiny$\pm$0.000} & 0.076\,{\tiny$\pm$0.002} & 0.059\,{\tiny$\pm$0.001} & 0.058\,{\tiny$\pm$0.001} & 0.059\,{\tiny$\pm$0.001} & 0.061\,{\tiny$\pm$0.001} \\
fexofenadine\_mpo & 0.680\,{\tiny$\pm$0.014} & 0.547\,{\tiny$\pm$0.001} & 0.518\,{\tiny$\pm$0.000} & 0.576\,{\tiny$\pm$0.002} & 0.556\,{\tiny$\pm$0.000} & 0.548\,{\tiny$\pm$0.001} & 0.547\,{\tiny$\pm$0.001} & 0.545\,{\tiny$\pm$0.000} & 0.531\,{\tiny$\pm$0.009} & 0.590\,{\tiny$\pm$0.000} & 0.548\,{\tiny$\pm$0.000} & 0.548\,{\tiny$\pm$0.000} \\
gsk3b & 0.609\,{\tiny$\pm$0.078} & 0.135\,{\tiny$\pm$0.000} & 0.114\,{\tiny$\pm$0.003} & 0.363\,{\tiny$\pm$0.004} & 0.121\,{\tiny$\pm$0.001} & 0.135\,{\tiny$\pm$0.001} & 0.135\,{\tiny$\pm$0.000} & 0.127\,{\tiny$\pm$0.002} & 0.123\,{\tiny$\pm$0.005} & 0.398\,{\tiny$\pm$0.001} & 0.097\,{\tiny$\pm$0.003} & 0.098\,{\tiny$\pm$0.003} \\
isomers\_c7h8n2o2 & 0.158\,{\tiny$\pm$0.045} & 0.013\,{\tiny$\pm$0.001} & 0.019\,{\tiny$\pm$0.000} & 0.033\,{\tiny$\pm$0.004} & 0.015\,{\tiny$\pm$0.000} & 0.012\,{\tiny$\pm$0.000} & 0.012\,{\tiny$\pm$0.000} & 0.071\,{\tiny$\pm$0.000} & 0.016\,{\tiny$\pm$0.003} & 0.022\,{\tiny$\pm$0.000} & 0.020\,{\tiny$\pm$0.000} & 0.025\,{\tiny$\pm$0.000} \\
isomers\_c9h10n2o2pf2cl & 0.097\,{\tiny$\pm$0.037} & 0.020\,{\tiny$\pm$0.004} & 0.010\,{\tiny$\pm$0.001} & 0.004\,{\tiny$\pm$0.001} & 0.005\,{\tiny$\pm$0.000} & 0.015\,{\tiny$\pm$0.000} & 0.015\,{\tiny$\pm$0.000} & 0.044\,{\tiny$\pm$0.001} & 0.010\,{\tiny$\pm$0.000} & 0.006\,{\tiny$\pm$0.000} & 0.010\,{\tiny$\pm$0.000} & 0.011\,{\tiny$\pm$0.000} \\
jnk3 & 0.308\,{\tiny$\pm$0.060} & 0.069\,{\tiny$\pm$0.001} & 0.060\,{\tiny$\pm$0.002} & 0.185\,{\tiny$\pm$0.001} & 0.079\,{\tiny$\pm$0.001} & 0.068\,{\tiny$\pm$0.001} & 0.067\,{\tiny$\pm$0.001} & 0.060\,{\tiny$\pm$0.001} & 0.057\,{\tiny$\pm$0.001} & 0.179\,{\tiny$\pm$0.000} & 0.057\,{\tiny$\pm$0.001} & 0.058\,{\tiny$\pm$0.001} \\
median1 & 0.183\,{\tiny$\pm$0.013} & 0.115\,{\tiny$\pm$0.001} & 0.139\,{\tiny$\pm$0.000} & 0.104\,{\tiny$\pm$0.001} & 0.114\,{\tiny$\pm$0.001} & 0.115\,{\tiny$\pm$0.001} & 0.115\,{\tiny$\pm$0.001} & 0.120\,{\tiny$\pm$0.000} & 0.109\,{\tiny$\pm$0.000} & 0.107\,{\tiny$\pm$0.000} & 0.109\,{\tiny$\pm$0.000} & 0.110\,{\tiny$\pm$0.000} \\
median2 & 0.195\,{\tiny$\pm$0.013} & 0.143\,{\tiny$\pm$0.001} & 0.143\,{\tiny$\pm$0.001} & 0.139\,{\tiny$\pm$0.000} & 0.144\,{\tiny$\pm$0.000} & 0.143\,{\tiny$\pm$0.000} & 0.142\,{\tiny$\pm$0.000} & 0.137\,{\tiny$\pm$0.001} & 0.143\,{\tiny$\pm$0.000} & 0.133\,{\tiny$\pm$0.000} & 0.143\,{\tiny$\pm$0.000} & 0.144\,{\tiny$\pm$0.000} \\
mestranol\_similarity & 0.345\,{\tiny$\pm$0.022} & 0.243\,{\tiny$\pm$0.002} & 0.260\,{\tiny$\pm$0.000} & 0.223\,{\tiny$\pm$0.001} & 0.263\,{\tiny$\pm$0.002} & 0.244\,{\tiny$\pm$0.001} & 0.245\,{\tiny$\pm$0.000} & 0.265\,{\tiny$\pm$0.001} & 0.241\,{\tiny$\pm$0.000} & 0.200\,{\tiny$\pm$0.004} & 0.241\,{\tiny$\pm$0.000} & 0.241\,{\tiny$\pm$0.000} \\
osimertinib\_mpo & 0.722\,{\tiny$\pm$0.014} & 0.607\,{\tiny$\pm$0.002} & 0.620\,{\tiny$\pm$0.001} & 0.686\,{\tiny$\pm$0.000} & 0.626\,{\tiny$\pm$0.000} & 0.605\,{\tiny$\pm$0.000} & 0.605\,{\tiny$\pm$0.001} & 0.608\,{\tiny$\pm$0.001} & 0.623\,{\tiny$\pm$0.001} & 0.705\,{\tiny$\pm$0.000} & 0.623\,{\tiny$\pm$0.001} & 0.612\,{\tiny$\pm$0.003} \\
perindopril\_mpo & 0.436\,{\tiny$\pm$0.006} & 0.353\,{\tiny$\pm$0.001} & 0.338\,{\tiny$\pm$0.000} & 0.305\,{\tiny$\pm$0.000} & 0.356\,{\tiny$\pm$0.000} & 0.353\,{\tiny$\pm$0.000} & 0.353\,{\tiny$\pm$0.001} & 0.325\,{\tiny$\pm$0.002} & 0.358\,{\tiny$\pm$0.000} & 0.293\,{\tiny$\pm$0.003} & 0.358\,{\tiny$\pm$0.000} & 0.351\,{\tiny$\pm$0.000} \\
qed & 0.712\,{\tiny$\pm$0.019} & 0.856\,{\tiny$\pm$0.000} & 0.854\,{\tiny$\pm$0.001} & 0.698\,{\tiny$\pm$0.001} & 0.840\,{\tiny$\pm$0.000} & 0.856\,{\tiny$\pm$0.000} & 0.855\,{\tiny$\pm$0.000} & 0.844\,{\tiny$\pm$0.000} & 0.855\,{\tiny$\pm$0.006} & 0.679\,{\tiny$\pm$0.001} & 0.843\,{\tiny$\pm$0.000} & 0.842\,{\tiny$\pm$0.001} \\
ranolazine\_mpo & 0.606\,{\tiny$\pm$0.011} & 0.197\,{\tiny$\pm$0.005} & 0.185\,{\tiny$\pm$0.001} & 0.277\,{\tiny$\pm$0.020} & 0.182\,{\tiny$\pm$0.000} & 0.191\,{\tiny$\pm$0.001} & 0.188\,{\tiny$\pm$0.003} & 0.181\,{\tiny$\pm$0.005} & 0.177\,{\tiny$\pm$0.001} & 0.304\,{\tiny$\pm$0.002} & 0.177\,{\tiny$\pm$0.001} & 0.198\,{\tiny$\pm$0.001} \\
scaffold\_hop & 0.456\,{\tiny$\pm$0.009} & 0.399\,{\tiny$\pm$0.000} & 0.393\,{\tiny$\pm$0.000} & 0.404\,{\tiny$\pm$0.001} & 0.402\,{\tiny$\pm$0.000} & 0.400\,{\tiny$\pm$0.000} & 0.399\,{\tiny$\pm$0.000} & 0.384\,{\tiny$\pm$0.001} & 0.406\,{\tiny$\pm$0.002} & 0.404\,{\tiny$\pm$0.000} & 0.402\,{\tiny$\pm$0.001} & 0.394\,{\tiny$\pm$0.000} \\
sitagliptin\_mpo & 0.053\,{\tiny$\pm$0.012} & 0.001\,{\tiny$\pm$0.000} & 0.032\,{\tiny$\pm$0.006} & 0.000\,{\tiny$\pm$0.000} & 0.002\,{\tiny$\pm$0.000} & 0.001\,{\tiny$\pm$0.000} & 0.001\,{\tiny$\pm$0.000} & 0.003\,{\tiny$\pm$0.000} & 0.003\,{\tiny$\pm$0.000} & 0.000\,{\tiny$\pm$0.000} & 0.003\,{\tiny$\pm$0.000} & 0.003\,{\tiny$\pm$0.000} \\
thiothixene\_rediscovery & 0.377\,{\tiny$\pm$0.011} & 0.229\,{\tiny$\pm$0.000} & 0.232\,{\tiny$\pm$0.000} & 0.197\,{\tiny$\pm$0.000} & 0.218\,{\tiny$\pm$0.000} & 0.230\,{\tiny$\pm$0.000} & 0.229\,{\tiny$\pm$0.001} & 0.213\,{\tiny$\pm$0.001} & 0.225\,{\tiny$\pm$0.003} & 0.186\,{\tiny$\pm$0.000} & 0.217\,{\tiny$\pm$0.000} & 0.215\,{\tiny$\pm$0.000} \\
troglitazone\_rediscovery & 0.394\,{\tiny$\pm$0.097} & 0.186\,{\tiny$\pm$0.000} & 0.181\,{\tiny$\pm$0.000} & 0.148\,{\tiny$\pm$0.000} & 0.195\,{\tiny$\pm$0.000} & 0.186\,{\tiny$\pm$0.000} & 0.186\,{\tiny$\pm$0.001} & 0.166\,{\tiny$\pm$0.000} & 0.194\,{\tiny$\pm$0.001} & 0.148\,{\tiny$\pm$0.000} & 0.194\,{\tiny$\pm$0.001} & 0.194\,{\tiny$\pm$0.000} \\
valsartan\_smarts & \textbf{0.363}\,{\tiny$\pm$0.052} & 0.000\,{\tiny$\pm$0.000} & 0.000\,{\tiny$\pm$0.000} & 0.000\,{\tiny$\pm$0.000} & 0.000\,{\tiny$\pm$0.000} & 0.000\,{\tiny$\pm$0.000} & 0.000\,{\tiny$\pm$0.000} & 0.000\,{\tiny$\pm$0.000} & 0.000\,{\tiny$\pm$0.000} & 0.000\,{\tiny$\pm$0.000} & 0.000\,{\tiny$\pm$0.000} & 0.000\,{\tiny$\pm$0.000} \\
zaleplon\_mpo & 0.084\,{\tiny$\pm$0.022} & 0.006\,{\tiny$\pm$0.001} & 0.001\,{\tiny$\pm$0.000} & 0.001\,{\tiny$\pm$0.000} & 0.002\,{\tiny$\pm$0.000} & 0.005\,{\tiny$\pm$0.000} & 0.007\,{\tiny$\pm$0.001} & 0.008\,{\tiny$\pm$0.001} & 0.002\,{\tiny$\pm$0.000} & 0.001\,{\tiny$\pm$0.000} & 0.002\,{\tiny$\pm$0.000} & 0.002\,{\tiny$\pm$0.000} \\
\midrule
Mean & $0.422$ & $0.248$ & $0.247$ & $0.247$ & $0.247$ & $0.247$ & $0.247$ & $0.246$ & $0.245$ & $0.245$ & $0.244$ & $0.244$ \\
Oracles won & 3 & 0 & 0 & 0 & 0 & 0 & 0 & 0 & 0 & 0 & 0 & 0 \\
\bottomrule
\end{tabular}
\end{table}
\end{landscape}

\begin{landscape}
\begin{table}[htbp]
\centering
\scriptsize
\setlength{\tabcolsep}{3pt}
\caption{\textbf{Top-10 AUC at a 210-call budget, Opus 5 multi-turn versus the full method panel (part c of 3).} Rows are the 23 PMO oracles; columns are Opus 5 multi-turn (reference, repeated in every part and set off by a vertical rule) and methods 23--33 of 33, ranked by macro-average Top-10 AUC. Each cell is mean $\pm$1 SE over seeds; higher is better; \textbf{bold} = best over the full 34-way panel for that oracle. The footer gives the per-column macro-average and the number of oracles won over the full panel.}
\label{tab:opus-panel-auc-c}
\begin{tabular}{@{}l c!{\vrule} *{11}{c}@{}}
\toprule
Oracle & \rotatebox{90}{Opus 5 (multi-turn)} & \rotatebox{90}{SMILES BAR} & \rotatebox{90}{screening} & \rotatebox{90}{selfies\_vae\_bo} & \rotatebox{90}{gp\_bo} & \rotatebox{90}{dst} & \rotatebox{90}{mol\_pal} & \rotatebox{90}{mars} & \rotatebox{90}{graph\_mcts} & \rotatebox{90}{MolGAN} & \rotatebox{90}{moldqn} & \rotatebox{90}{selfies\_ga} \\
\midrule
albuterol\_similarity & \textbf{0.739}\,{\tiny$\pm$0.020} & 0.330\,{\tiny$\pm$0.001} & 0.348\,{\tiny$\pm$0.001} & 0.331\,{\tiny$\pm$0.001} & 0.333\,{\tiny$\pm$0.001} & 0.338\,{\tiny$\pm$0.003} & 0.353\,{\tiny$\pm$0.000} & 0.405\,{\tiny$\pm$0.002} & 0.417\,{\tiny$\pm$0.003} & 0.267\,{\tiny$\pm$0.005} & 0.206\,{\tiny$\pm$0.001} & 0.178\,{\tiny$\pm$0.001} \\
amlodipine\_mpo & 0.485\,{\tiny$\pm$0.010} & 0.417\,{\tiny$\pm$0.000} & 0.411\,{\tiny$\pm$0.000} & 0.418\,{\tiny$\pm$0.007} & 0.410\,{\tiny$\pm$0.001} & 0.425\,{\tiny$\pm$0.002} & 0.396\,{\tiny$\pm$0.000} & 0.359\,{\tiny$\pm$0.002} & 0.321\,{\tiny$\pm$0.002} & 0.171\,{\tiny$\pm$0.001} & 0.154\,{\tiny$\pm$0.003} & 0.001\,{\tiny$\pm$0.001} \\
celecoxib\_rediscovery & \textbf{0.594}\,{\tiny$\pm$0.076} & 0.209\,{\tiny$\pm$0.001} & 0.231\,{\tiny$\pm$0.002} & 0.216\,{\tiny$\pm$0.001} & 0.223\,{\tiny$\pm$0.001} & 0.222\,{\tiny$\pm$0.002} & 0.245\,{\tiny$\pm$0.000} & 0.198\,{\tiny$\pm$0.002} & 0.131\,{\tiny$\pm$0.002} & 0.076\,{\tiny$\pm$0.002} & 0.069\,{\tiny$\pm$0.000} & 0.030\,{\tiny$\pm$0.002} \\
deco\_hop & 0.583\,{\tiny$\pm$0.026} & 0.529\,{\tiny$\pm$0.000} & 0.530\,{\tiny$\pm$0.000} & 0.531\,{\tiny$\pm$0.000} & 0.534\,{\tiny$\pm$0.000} & 0.531\,{\tiny$\pm$0.002} & 0.527\,{\tiny$\pm$0.000} & 0.532\,{\tiny$\pm$0.000} & 0.494\,{\tiny$\pm$0.002} & 0.482\,{\tiny$\pm$0.000} & 0.494\,{\tiny$\pm$0.000} & 0.476\,{\tiny$\pm$0.000} \\
drd2 & 0.524\,{\tiny$\pm$0.091} & 0.134\,{\tiny$\pm$0.006} & 0.030\,{\tiny$\pm$0.001} & 0.059\,{\tiny$\pm$0.001} & 0.059\,{\tiny$\pm$0.003} & 0.064\,{\tiny$\pm$0.004} & 0.044\,{\tiny$\pm$0.003} & 0.059\,{\tiny$\pm$0.010} & 0.029\,{\tiny$\pm$0.000} & 0.017\,{\tiny$\pm$0.001} & 0.009\,{\tiny$\pm$0.000} & 0.013\,{\tiny$\pm$0.000} \\
fexofenadine\_mpo & 0.680\,{\tiny$\pm$0.014} & 0.524\,{\tiny$\pm$0.000} & 0.543\,{\tiny$\pm$0.000} & 0.548\,{\tiny$\pm$0.000} & 0.514\,{\tiny$\pm$0.000} & 0.546\,{\tiny$\pm$0.002} & 0.499\,{\tiny$\pm$0.001} & 0.447\,{\tiny$\pm$0.003} & 0.339\,{\tiny$\pm$0.004} & 0.152\,{\tiny$\pm$0.004} & 0.301\,{\tiny$\pm$0.005} & 0.002\,{\tiny$\pm$0.000} \\
gsk3b & 0.609\,{\tiny$\pm$0.078} & 0.100\,{\tiny$\pm$0.002} & 0.139\,{\tiny$\pm$0.000} & 0.097\,{\tiny$\pm$0.003} & 0.110\,{\tiny$\pm$0.001} & 0.115\,{\tiny$\pm$0.016} & 0.081\,{\tiny$\pm$0.001} & 0.167\,{\tiny$\pm$0.002} & 0.081\,{\tiny$\pm$0.003} & 0.072\,{\tiny$\pm$0.003} & 0.094\,{\tiny$\pm$0.001} & 0.058\,{\tiny$\pm$0.000} \\
isomers\_c7h8n2o2 & 0.158\,{\tiny$\pm$0.045} & 0.002\,{\tiny$\pm$0.000} & 0.003\,{\tiny$\pm$0.000} & 0.020\,{\tiny$\pm$0.000} & 0.015\,{\tiny$\pm$0.000} & 0.009\,{\tiny$\pm$0.002} & 0.009\,{\tiny$\pm$0.000} & 0.012\,{\tiny$\pm$0.001} & 0.049\,{\tiny$\pm$0.010} & 0.472\,{\tiny$\pm$0.010} & 0.067\,{\tiny$\pm$0.003} & 0.023\,{\tiny$\pm$0.001} \\
isomers\_c9h10n2o2pf2cl & 0.097\,{\tiny$\pm$0.037} & 0.004\,{\tiny$\pm$0.000} & 0.016\,{\tiny$\pm$0.000} & 0.010\,{\tiny$\pm$0.000} & 0.007\,{\tiny$\pm$0.003} & 0.007\,{\tiny$\pm$0.001} & 0.004\,{\tiny$\pm$0.000} & 0.001\,{\tiny$\pm$0.000} & 0.045\,{\tiny$\pm$0.003} & 0.264\,{\tiny$\pm$0.009} & 0.028\,{\tiny$\pm$0.003} & 0.003\,{\tiny$\pm$0.000} \\
jnk3 & 0.308\,{\tiny$\pm$0.060} & 0.055\,{\tiny$\pm$0.000} & 0.069\,{\tiny$\pm$0.000} & 0.057\,{\tiny$\pm$0.001} & 0.074\,{\tiny$\pm$0.000} & 0.061\,{\tiny$\pm$0.005} & 0.063\,{\tiny$\pm$0.000} & 0.080\,{\tiny$\pm$0.002} & 0.026\,{\tiny$\pm$0.002} & 0.024\,{\tiny$\pm$0.002} & 0.026\,{\tiny$\pm$0.001} & 0.012\,{\tiny$\pm$0.000} \\
median1 & 0.183\,{\tiny$\pm$0.013} & 0.142\,{\tiny$\pm$0.000} & 0.127\,{\tiny$\pm$0.000} & 0.109\,{\tiny$\pm$0.000} & 0.118\,{\tiny$\pm$0.000} & 0.135\,{\tiny$\pm$0.005} & 0.118\,{\tiny$\pm$0.000} & 0.144\,{\tiny$\pm$0.002} & 0.100\,{\tiny$\pm$0.002} & 0.157\,{\tiny$\pm$0.001} & 0.058\,{\tiny$\pm$0.001} & 0.048\,{\tiny$\pm$0.000} \\
median2 & 0.195\,{\tiny$\pm$0.013} & 0.138\,{\tiny$\pm$0.000} & 0.145\,{\tiny$\pm$0.000} & 0.143\,{\tiny$\pm$0.000} & 0.146\,{\tiny$\pm$0.000} & 0.133\,{\tiny$\pm$0.011} & 0.141\,{\tiny$\pm$0.000} & 0.128\,{\tiny$\pm$0.001} & 0.088\,{\tiny$\pm$0.001} & 0.064\,{\tiny$\pm$0.001} & 0.067\,{\tiny$\pm$0.000} & 0.020\,{\tiny$\pm$0.001} \\
mestranol\_similarity & 0.345\,{\tiny$\pm$0.022} & 0.257\,{\tiny$\pm$0.000} & 0.262\,{\tiny$\pm$0.000} & 0.241\,{\tiny$\pm$0.000} & 0.243\,{\tiny$\pm$0.001} & 0.214\,{\tiny$\pm$0.015} & 0.267\,{\tiny$\pm$0.001} & 0.240\,{\tiny$\pm$0.002} & 0.142\,{\tiny$\pm$0.001} & 0.115\,{\tiny$\pm$0.001} & 0.069\,{\tiny$\pm$0.000} & 0.048\,{\tiny$\pm$0.000} \\
osimertinib\_mpo & 0.722\,{\tiny$\pm$0.014} & 0.586\,{\tiny$\pm$0.001} & 0.582\,{\tiny$\pm$0.001} & 0.623\,{\tiny$\pm$0.001} & 0.572\,{\tiny$\pm$0.000} & 0.593\,{\tiny$\pm$0.011} & 0.515\,{\tiny$\pm$0.006} & 0.409\,{\tiny$\pm$0.007} & 0.579\,{\tiny$\pm$0.002} & 0.240\,{\tiny$\pm$0.007} & 0.565\,{\tiny$\pm$0.002} & 0.020\,{\tiny$\pm$0.005} \\
perindopril\_mpo & 0.436\,{\tiny$\pm$0.006} & 0.325\,{\tiny$\pm$0.000} & 0.332\,{\tiny$\pm$0.000} & 0.358\,{\tiny$\pm$0.000} & 0.349\,{\tiny$\pm$0.000} & 0.346\,{\tiny$\pm$0.005} & 0.344\,{\tiny$\pm$0.001} & 0.298\,{\tiny$\pm$0.004} & 0.091\,{\tiny$\pm$0.003} & 0.064\,{\tiny$\pm$0.003} & 0.016\,{\tiny$\pm$0.000} & 0.004\,{\tiny$\pm$0.000} \\
qed & 0.712\,{\tiny$\pm$0.019} & 0.844\,{\tiny$\pm$0.001} & 0.863\,{\tiny$\pm$0.000} & 0.843\,{\tiny$\pm$0.000} & 0.856\,{\tiny$\pm$0.000} & 0.845\,{\tiny$\pm$0.002} & 0.858\,{\tiny$\pm$0.000} & 0.803\,{\tiny$\pm$0.001} & 0.699\,{\tiny$\pm$0.004} & 0.551\,{\tiny$\pm$0.003} & 0.448\,{\tiny$\pm$0.013} & 0.406\,{\tiny$\pm$0.001} \\
ranolazine\_mpo & 0.606\,{\tiny$\pm$0.011} & 0.159\,{\tiny$\pm$0.001} & 0.170\,{\tiny$\pm$0.000} & 0.177\,{\tiny$\pm$0.001} & 0.197\,{\tiny$\pm$0.002} & 0.165\,{\tiny$\pm$0.036} & 0.156\,{\tiny$\pm$0.001} & 0.270\,{\tiny$\pm$0.012} & 0.044\,{\tiny$\pm$0.003} & 0.003\,{\tiny$\pm$0.000} & 0.003\,{\tiny$\pm$0.000} & 0.002\,{\tiny$\pm$0.000} \\
scaffold\_hop & 0.456\,{\tiny$\pm$0.009} & 0.398\,{\tiny$\pm$0.000} & 0.400\,{\tiny$\pm$0.000} & 0.402\,{\tiny$\pm$0.001} & 0.392\,{\tiny$\pm$0.000} & 0.400\,{\tiny$\pm$0.002} & 0.396\,{\tiny$\pm$0.000} & 0.403\,{\tiny$\pm$0.001} & 0.346\,{\tiny$\pm$0.001} & 0.327\,{\tiny$\pm$0.000} & 0.342\,{\tiny$\pm$0.000} & 0.318\,{\tiny$\pm$0.000} \\
sitagliptin\_mpo & 0.053\,{\tiny$\pm$0.012} & 0.040\,{\tiny$\pm$0.004} & 0.002\,{\tiny$\pm$0.000} & 0.003\,{\tiny$\pm$0.000} & 0.001\,{\tiny$\pm$0.000} & 0.003\,{\tiny$\pm$0.000} & 0.004\,{\tiny$\pm$0.000} & 0.002\,{\tiny$\pm$0.000} & 0.000\,{\tiny$\pm$0.000} & 0.012\,{\tiny$\pm$0.003} & 0.000\,{\tiny$\pm$0.000} & 0.000\,{\tiny$\pm$0.000} \\
thiothixene\_rediscovery & 0.377\,{\tiny$\pm$0.011} & 0.215\,{\tiny$\pm$0.001} & 0.220\,{\tiny$\pm$0.001} & 0.217\,{\tiny$\pm$0.000} & 0.235\,{\tiny$\pm$0.000} & 0.219\,{\tiny$\pm$0.001} & 0.208\,{\tiny$\pm$0.001} & 0.191\,{\tiny$\pm$0.001} & 0.144\,{\tiny$\pm$0.002} & 0.072\,{\tiny$\pm$0.001} & 0.039\,{\tiny$\pm$0.002} & 0.037\,{\tiny$\pm$0.001} \\
troglitazone\_rediscovery & 0.394\,{\tiny$\pm$0.097} & 0.186\,{\tiny$\pm$0.001} & 0.178\,{\tiny$\pm$0.000} & 0.194\,{\tiny$\pm$0.001} & 0.187\,{\tiny$\pm$0.001} & 0.178\,{\tiny$\pm$0.014} & 0.186\,{\tiny$\pm$0.001} & 0.170\,{\tiny$\pm$0.001} & 0.161\,{\tiny$\pm$0.001} & 0.096\,{\tiny$\pm$0.001} & 0.050\,{\tiny$\pm$0.001} & 0.032\,{\tiny$\pm$0.002} \\
valsartan\_smarts & \textbf{0.363}\,{\tiny$\pm$0.052} & 0.000\,{\tiny$\pm$0.000} & 0.000\,{\tiny$\pm$0.000} & 0.000\,{\tiny$\pm$0.000} & 0.000\,{\tiny$\pm$0.000} & 0.000\,{\tiny$\pm$0.000} & 0.000\,{\tiny$\pm$0.000} & 0.000\,{\tiny$\pm$0.000} & 0.000\,{\tiny$\pm$0.000} & 0.000\,{\tiny$\pm$0.000} & 0.000\,{\tiny$\pm$0.000} & 0.000\,{\tiny$\pm$0.000} \\
zaleplon\_mpo & 0.084\,{\tiny$\pm$0.022} & 0.006\,{\tiny$\pm$0.003} & 0.001\,{\tiny$\pm$0.000} & 0.002\,{\tiny$\pm$0.000} & 0.007\,{\tiny$\pm$0.000} & 0.001\,{\tiny$\pm$0.000} & 0.002\,{\tiny$\pm$0.000} & 0.026\,{\tiny$\pm$0.000} & 0.002\,{\tiny$\pm$0.001} & 0.000\,{\tiny$\pm$0.000} & 0.000\,{\tiny$\pm$0.000} & 0.000\,{\tiny$\pm$0.000} \\
\midrule
Mean & $0.422$ & $0.243$ & $0.243$ & $0.243$ & $0.243$ & $0.241$ & $0.235$ & $0.232$ & $0.188$ & $0.161$ & $0.135$ & $0.075$ \\
Oracles won & 3 & 0 & 0 & 0 & 0 & 0 & 0 & 0 & 0 & 0 & 0 & 0 \\
\bottomrule
\end{tabular}
\end{table}
\end{landscape}

\begin{landscape}
\begin{table}[htbp]
\centering
\scriptsize
\setlength{\tabcolsep}{3pt}
\caption{\textbf{Top-10 mean at a 210-call budget, Opus 5 multi-turn versus the full method panel (part a of 3).} Rows are the 23 PMO oracles; columns are Opus 5 multi-turn (reference, repeated in every part) and methods 1--11 of 33, in the same order as the Top-10 AUC tables (Opus 5 multi-turn set off by a vertical rule). Each cell is mean $\pm$1 SE over seeds; higher is better; \textbf{bold} = best over the full 34-way panel for that oracle. The footer gives the per-column macro-average and the number of oracles won over the full panel.}
\label{tab:opus-panel-mean-a}
\begin{tabular}{@{}l c!{\vrule} *{11}{c}@{}}
\toprule
Oracle & \rotatebox{90}{Opus 5 (multi-turn)} & \rotatebox{90}{GenMol} & \rotatebox{90}{ExLLM (Gemini 2.5 Flash)} & \rotatebox{90}{MolLEO (BioT5)} & \rotatebox{90}{synnet} & \rotatebox{90}{dog\_gen} & \rotatebox{90}{dog\_ae} & \rotatebox{90}{graph\_ga} & \rotatebox{90}{SMILES Aug. Memory} & \rotatebox{90}{reinvent} & \rotatebox{90}{REINVENT-Transformer} & \rotatebox{90}{smiles\_lstm\_hc} \\
\midrule
albuterol\_similarity & \textbf{1.000}\,{\tiny$\pm$0.000} & 0.653\,{\tiny$\pm$0.009} & \textbf{1.000}\,{\tiny$\pm$0.000} & 0.475\,{\tiny$\pm$0.016} & 0.420\,{\tiny$\pm$0.005} & 0.460\,{\tiny$\pm$0.000} & 0.419\,{\tiny$\pm$0.010} & 0.383\,{\tiny$\pm$0.002} & 0.408\,{\tiny$\pm$0.008} & 0.400\,{\tiny$\pm$0.005} & 0.408\,{\tiny$\pm$0.006} & 0.384\,{\tiny$\pm$0.003} \\
amlodipine\_mpo & 0.564\,{\tiny$\pm$0.011} & 0.665\,{\tiny$\pm$0.005} & \textbf{0.871}\,{\tiny$\pm$0.002} & 0.537\,{\tiny$\pm$0.006} & 0.503\,{\tiny$\pm$0.002} & 0.472\,{\tiny$\pm$0.000} & 0.461\,{\tiny$\pm$0.002} & 0.483\,{\tiny$\pm$0.002} & 0.472\,{\tiny$\pm$0.003} & 0.458\,{\tiny$\pm$0.002} & 0.472\,{\tiny$\pm$0.004} & 0.482\,{\tiny$\pm$0.003} \\
celecoxib\_rediscovery & \textbf{1.000}\,{\tiny$\pm$0.000} & 0.537\,{\tiny$\pm$0.013} & 0.834\,{\tiny$\pm$0.003} & 0.294\,{\tiny$\pm$0.008} & 0.302\,{\tiny$\pm$0.004} & 0.273\,{\tiny$\pm$0.000} & 0.304\,{\tiny$\pm$0.008} & 0.275\,{\tiny$\pm$0.001} & 0.268\,{\tiny$\pm$0.003} & 0.304\,{\tiny$\pm$0.001} & 0.275\,{\tiny$\pm$0.004} & 0.272\,{\tiny$\pm$0.004} \\
deco\_hop & 0.661\,{\tiny$\pm$0.056} & 0.888\,{\tiny$\pm$0.002} & \textbf{0.928}\,{\tiny$\pm$0.008} & 0.572\,{\tiny$\pm$0.002} & 0.588\,{\tiny$\pm$0.002} & 0.576\,{\tiny$\pm$0.000} & 0.583\,{\tiny$\pm$0.010} & 0.567\,{\tiny$\pm$0.001} & 0.568\,{\tiny$\pm$0.001} & 0.569\,{\tiny$\pm$0.000} & 0.565\,{\tiny$\pm$0.001} & 0.567\,{\tiny$\pm$0.001} \\
drd2 & 0.945\,{\tiny$\pm$0.028} & \textbf{0.997}\,{\tiny$\pm$0.001} & 0.966\,{\tiny$\pm$0.019} & 0.625\,{\tiny$\pm$0.058} & 0.502\,{\tiny$\pm$0.050} & 0.391\,{\tiny$\pm$0.000} & 0.328\,{\tiny$\pm$0.037} & 0.417\,{\tiny$\pm$0.076} & 0.257\,{\tiny$\pm$0.068} & 0.140\,{\tiny$\pm$0.016} & 0.141\,{\tiny$\pm$0.025} & 0.080\,{\tiny$\pm$0.003} \\
fexofenadine\_mpo & 0.770\,{\tiny$\pm$0.015} & 0.772\,{\tiny$\pm$0.005} & \textbf{0.985}\,{\tiny$\pm$0.001} & 0.646\,{\tiny$\pm$0.005} & 0.701\,{\tiny$\pm$0.002} & 0.583\,{\tiny$\pm$0.000} & 0.615\,{\tiny$\pm$0.007} & 0.619\,{\tiny$\pm$0.002} & 0.632\,{\tiny$\pm$0.004} & 0.602\,{\tiny$\pm$0.004} & 0.623\,{\tiny$\pm$0.007} & 0.621\,{\tiny$\pm$0.003} \\
gsk3b & 0.775\,{\tiny$\pm$0.068} & \textbf{0.878}\,{\tiny$\pm$0.012} & 0.561\,{\tiny$\pm$0.044} & 0.386\,{\tiny$\pm$0.046} & 0.301\,{\tiny$\pm$0.017} & 0.409\,{\tiny$\pm$0.000} & 0.335\,{\tiny$\pm$0.010} & 0.237\,{\tiny$\pm$0.045} & 0.171\,{\tiny$\pm$0.011} & 0.155\,{\tiny$\pm$0.016} & 0.218\,{\tiny$\pm$0.022} & 0.221\,{\tiny$\pm$0.021} \\
isomers\_c7h8n2o2 & 0.522\,{\tiny$\pm$0.120} & 0.324\,{\tiny$\pm$0.037} & \textbf{1.000}\,{\tiny$\pm$0.000} & 0.540\,{\tiny$\pm$0.034} & 0.038\,{\tiny$\pm$0.003} & 0.039\,{\tiny$\pm$0.000} & 0.088\,{\tiny$\pm$0.013} & 0.056\,{\tiny$\pm$0.030} & 0.031\,{\tiny$\pm$0.005} & 0.021\,{\tiny$\pm$0.004} & 0.018\,{\tiny$\pm$0.004} & 0.021\,{\tiny$\pm$0.001} \\
isomers\_c9h10n2o2pf2cl & 0.328\,{\tiny$\pm$0.110} & 0.366\,{\tiny$\pm$0.016} & \textbf{1.000}\,{\tiny$\pm$0.000} & 0.506\,{\tiny$\pm$0.020} & 0.009\,{\tiny$\pm$0.003} & 0.022\,{\tiny$\pm$0.000} & 0.016\,{\tiny$\pm$0.002} & 0.074\,{\tiny$\pm$0.044} & 0.018\,{\tiny$\pm$0.005} & 0.013\,{\tiny$\pm$0.004} & 0.009\,{\tiny$\pm$0.002} & 0.023\,{\tiny$\pm$0.001} \\
jnk3 & 0.495\,{\tiny$\pm$0.098} & \textbf{0.762}\,{\tiny$\pm$0.016} & 0.568\,{\tiny$\pm$0.011} & 0.166\,{\tiny$\pm$0.015} & 0.182\,{\tiny$\pm$0.005} & 0.217\,{\tiny$\pm$0.000} & 0.186\,{\tiny$\pm$0.016} & 0.107\,{\tiny$\pm$0.006} & 0.111\,{\tiny$\pm$0.008} & 0.102\,{\tiny$\pm$0.007} & 0.116\,{\tiny$\pm$0.007} & 0.114\,{\tiny$\pm$0.011} \\
median1 & 0.260\,{\tiny$\pm$0.029} & 0.321\,{\tiny$\pm$0.003} & \textbf{0.350}\,{\tiny$\pm$0.002} & 0.168\,{\tiny$\pm$0.006} & 0.161\,{\tiny$\pm$0.004} & 0.121\,{\tiny$\pm$0.000} & 0.139\,{\tiny$\pm$0.008} & 0.147\,{\tiny$\pm$0.003} & 0.155\,{\tiny$\pm$0.002} & 0.159\,{\tiny$\pm$0.002} & 0.160\,{\tiny$\pm$0.004} & 0.150\,{\tiny$\pm$0.001} \\
median2 & 0.239\,{\tiny$\pm$0.016} & 0.320\,{\tiny$\pm$0.006} & \textbf{0.403}\,{\tiny$\pm$0.004} & 0.167\,{\tiny$\pm$0.003} & 0.183\,{\tiny$\pm$0.001} & 0.154\,{\tiny$\pm$0.000} & 0.158\,{\tiny$\pm$0.003} & 0.161\,{\tiny$\pm$0.000} & 0.165\,{\tiny$\pm$0.000} & 0.154\,{\tiny$\pm$0.001} & 0.162\,{\tiny$\pm$0.001} & 0.161\,{\tiny$\pm$0.000} \\
mestranol\_similarity & 0.439\,{\tiny$\pm$0.034} & 0.697\,{\tiny$\pm$0.023} & \textbf{0.974}\,{\tiny$\pm$0.014} & 0.340\,{\tiny$\pm$0.013} & 0.305\,{\tiny$\pm$0.003} & 0.288\,{\tiny$\pm$0.000} & 0.288\,{\tiny$\pm$0.009} & 0.310\,{\tiny$\pm$0.009} & 0.290\,{\tiny$\pm$0.002} & 0.294\,{\tiny$\pm$0.004} & 0.327\,{\tiny$\pm$0.008} & 0.321\,{\tiny$\pm$0.001} \\
osimertinib\_mpo & 0.830\,{\tiny$\pm$0.015} & \textbf{0.844}\,{\tiny$\pm$0.004} & 0.832\,{\tiny$\pm$0.019} & 0.740\,{\tiny$\pm$0.004} & 0.733\,{\tiny$\pm$0.003} & 0.677\,{\tiny$\pm$0.000} & 0.666\,{\tiny$\pm$0.013} & 0.694\,{\tiny$\pm$0.008} & 0.738\,{\tiny$\pm$0.004} & 0.726\,{\tiny$\pm$0.001} & 0.726\,{\tiny$\pm$0.006} & 0.687\,{\tiny$\pm$0.002} \\
perindopril\_mpo & 0.535\,{\tiny$\pm$0.017} & 0.612\,{\tiny$\pm$0.009} & \textbf{0.770}\,{\tiny$\pm$0.007} & 0.596\,{\tiny$\pm$0.012} & 0.443\,{\tiny$\pm$0.004} & 0.394\,{\tiny$\pm$0.000} & 0.388\,{\tiny$\pm$0.002} & 0.384\,{\tiny$\pm$0.001} & 0.388\,{\tiny$\pm$0.002} & 0.389\,{\tiny$\pm$0.003} & 0.392\,{\tiny$\pm$0.002} & 0.386\,{\tiny$\pm$0.000} \\
qed & 0.729\,{\tiny$\pm$0.020} & 0.917\,{\tiny$\pm$0.001} & \textbf{0.946}\,{\tiny$\pm$0.000} & 0.922\,{\tiny$\pm$0.002} & 0.935\,{\tiny$\pm$0.000} & 0.892\,{\tiny$\pm$0.000} & 0.888\,{\tiny$\pm$0.003} & 0.918\,{\tiny$\pm$0.004} & 0.928\,{\tiny$\pm$0.000} & 0.919\,{\tiny$\pm$0.001} & 0.926\,{\tiny$\pm$0.002} & 0.923\,{\tiny$\pm$0.001} \\
ranolazine\_mpo & 0.781\,{\tiny$\pm$0.012} & 0.705\,{\tiny$\pm$0.006} & \textbf{0.790}\,{\tiny$\pm$0.006} & 0.496\,{\tiny$\pm$0.028} & 0.544\,{\tiny$\pm$0.015} & 0.388\,{\tiny$\pm$0.000} & 0.548\,{\tiny$\pm$0.009} & 0.322\,{\tiny$\pm$0.027} & 0.329\,{\tiny$\pm$0.034} & 0.370\,{\tiny$\pm$0.009} & 0.328\,{\tiny$\pm$0.014} & 0.323\,{\tiny$\pm$0.020} \\
scaffold\_hop & 0.512\,{\tiny$\pm$0.011} & 0.533\,{\tiny$\pm$0.003} & \textbf{0.880}\,{\tiny$\pm$0.012} & 0.441\,{\tiny$\pm$0.003} & 0.466\,{\tiny$\pm$0.003} & 0.449\,{\tiny$\pm$0.000} & 0.445\,{\tiny$\pm$0.004} & 0.434\,{\tiny$\pm$0.001} & 0.435\,{\tiny$\pm$0.001} & 0.438\,{\tiny$\pm$0.001} & 0.435\,{\tiny$\pm$0.001} & 0.435\,{\tiny$\pm$0.001} \\
sitagliptin\_mpo & 0.147\,{\tiny$\pm$0.037} & \textbf{0.266}\,{\tiny$\pm$0.013} & 0.184\,{\tiny$\pm$0.017} & 0.181\,{\tiny$\pm$0.019} & 0.004\,{\tiny$\pm$0.001} & 0.002\,{\tiny$\pm$0.000} & 0.001\,{\tiny$\pm$0.000} & 0.006\,{\tiny$\pm$0.004} & 0.054\,{\tiny$\pm$0.013} & 0.003\,{\tiny$\pm$0.000} & 0.052\,{\tiny$\pm$0.005} & 0.002\,{\tiny$\pm$0.000} \\
thiothixene\_rediscovery & 0.503\,{\tiny$\pm$0.030} & 0.456\,{\tiny$\pm$0.018} & \textbf{0.902}\,{\tiny$\pm$0.010} & 0.282\,{\tiny$\pm$0.007} & 0.319\,{\tiny$\pm$0.003} & 0.242\,{\tiny$\pm$0.000} & 0.260\,{\tiny$\pm$0.007} & 0.257\,{\tiny$\pm$0.001} & 0.271\,{\tiny$\pm$0.002} & 0.256\,{\tiny$\pm$0.003} & 0.266\,{\tiny$\pm$0.004} & 0.257\,{\tiny$\pm$0.001} \\
troglitazone\_rediscovery & 0.578\,{\tiny$\pm$0.107} & \textbf{0.631}\,{\tiny$\pm$0.012} & 0.523\,{\tiny$\pm$0.010} & 0.216\,{\tiny$\pm$0.005} & 0.223\,{\tiny$\pm$0.003} & 0.203\,{\tiny$\pm$0.000} & 0.209\,{\tiny$\pm$0.005} & 0.213\,{\tiny$\pm$0.001} & 0.211\,{\tiny$\pm$0.001} & 0.203\,{\tiny$\pm$0.001} & 0.213\,{\tiny$\pm$0.003} & 0.213\,{\tiny$\pm$0.001} \\
valsartan\_smarts & \textbf{0.891}\,{\tiny$\pm$0.026} & 0.046\,{\tiny$\pm$0.014} & 0.014\,{\tiny$\pm$0.010} & 0.000\,{\tiny$\pm$0.000} & 0.000\,{\tiny$\pm$0.000} & 0.000\,{\tiny$\pm$0.000} & 0.000\,{\tiny$\pm$0.000} & 0.000\,{\tiny$\pm$0.000} & 0.000\,{\tiny$\pm$0.000} & 0.010\,{\tiny$\pm$0.009} & 0.000\,{\tiny$\pm$0.000} & 0.000\,{\tiny$\pm$0.000} \\
zaleplon\_mpo & 0.212\,{\tiny$\pm$0.037} & 0.389\,{\tiny$\pm$0.002} & \textbf{0.689}\,{\tiny$\pm$0.003} & 0.371\,{\tiny$\pm$0.007} & 0.026\,{\tiny$\pm$0.005} & 0.002\,{\tiny$\pm$0.000} & 0.001\,{\tiny$\pm$0.000} & 0.027\,{\tiny$\pm$0.018} & 0.004\,{\tiny$\pm$0.002} & 0.001\,{\tiny$\pm$0.000} & 0.002\,{\tiny$\pm$0.001} & 0.007\,{\tiny$\pm$0.001} \\
\midrule
Mean & $0.596$ & $0.590$ & $0.738$ & $0.420$ & $0.343$ & $0.315$ & $0.319$ & $0.308$ & $0.300$ & $0.291$ & $0.297$ & $0.289$ \\
Oracles won & 3 & 6 & 15 & 0 & 0 & 0 & 0 & 0 & 0 & 0 & 0 & 0 \\
\bottomrule
\end{tabular}
\end{table}
\end{landscape}

\begin{landscape}
\begin{table}[htbp]
\centering
\scriptsize
\setlength{\tabcolsep}{3pt}
\caption{\textbf{Top-10 mean at a 210-call budget, Opus 5 multi-turn versus the full method panel (part b of 3).} Rows are the 23 PMO oracles; columns are Opus 5 multi-turn (reference, repeated in every part) and methods 12--22 of 33, in the same order as the Top-10 AUC tables (Opus 5 multi-turn set off by a vertical rule). Each cell is mean $\pm$1 SE over seeds; higher is better; \textbf{bold} = best over the full 34-way panel for that oracle. The footer gives the per-column macro-average and the number of oracles won over the full panel.}
\label{tab:opus-panel-mean-b}
\begin{tabular}{@{}l c!{\vrule} *{11}{c}@{}}
\toprule
Oracle & \rotatebox{90}{Opus 5 (multi-turn)} & \rotatebox{90}{smiles\_ga} & \rotatebox{90}{SMILES AHC} & \rotatebox{90}{gflownet} & \rotatebox{90}{pasithea} & \rotatebox{90}{stoned} & \rotatebox{90}{selfies\_lstm\_hc} & \rotatebox{90}{reinvent\_selfies} & \rotatebox{90}{mimosa} & \rotatebox{90}{gflownet\_al} & \rotatebox{90}{smiles\_vae\_bo} & \rotatebox{90}{jt\_vae\_bo} \\
\midrule
albuterol\_similarity & \textbf{1.000}\,{\tiny$\pm$0.000} & 0.380\,{\tiny$\pm$0.000} & 0.364\,{\tiny$\pm$0.001} & 0.290\,{\tiny$\pm$0.004} & 0.364\,{\tiny$\pm$0.002} & 0.380\,{\tiny$\pm$0.002} & 0.387\,{\tiny$\pm$0.005} & 0.398\,{\tiny$\pm$0.004} & 0.386\,{\tiny$\pm$0.002} & 0.304\,{\tiny$\pm$0.001} & 0.386\,{\tiny$\pm$0.002} & 0.391\,{\tiny$\pm$0.003} \\
amlodipine\_mpo & 0.564\,{\tiny$\pm$0.011} & 0.484\,{\tiny$\pm$0.001} & 0.467\,{\tiny$\pm$0.001} & 0.380\,{\tiny$\pm$0.004} & 0.479\,{\tiny$\pm$0.001} & 0.484\,{\tiny$\pm$0.001} & 0.482\,{\tiny$\pm$0.002} & 0.460\,{\tiny$\pm$0.003} & 0.478\,{\tiny$\pm$0.000} & 0.375\,{\tiny$\pm$0.001} & 0.478\,{\tiny$\pm$0.000} & 0.473\,{\tiny$\pm$0.003} \\
celecoxib\_rediscovery & \textbf{1.000}\,{\tiny$\pm$0.000} & 0.274\,{\tiny$\pm$0.002} & 0.276\,{\tiny$\pm$0.002} & 0.203\,{\tiny$\pm$0.002} & 0.263\,{\tiny$\pm$0.004} & 0.274\,{\tiny$\pm$0.004} & 0.271\,{\tiny$\pm$0.005} & 0.268\,{\tiny$\pm$0.003} & 0.270\,{\tiny$\pm$0.001} & 0.196\,{\tiny$\pm$0.002} & 0.270\,{\tiny$\pm$0.001} & 0.260\,{\tiny$\pm$0.009} \\
deco\_hop & 0.661\,{\tiny$\pm$0.056} & 0.565\,{\tiny$\pm$0.000} & 0.565\,{\tiny$\pm$0.001} & 0.569\,{\tiny$\pm$0.001} & 0.570\,{\tiny$\pm$0.000} & 0.567\,{\tiny$\pm$0.001} & 0.566\,{\tiny$\pm$0.000} & 0.559\,{\tiny$\pm$0.001} & 0.568\,{\tiny$\pm$0.001} & 0.567\,{\tiny$\pm$0.000} & 0.570\,{\tiny$\pm$0.001} & 0.566\,{\tiny$\pm$0.001} \\
drd2 & 0.945\,{\tiny$\pm$0.028} & 0.105\,{\tiny$\pm$0.028} & 0.185\,{\tiny$\pm$0.007} & 0.173\,{\tiny$\pm$0.012} & 0.078\,{\tiny$\pm$0.001} & 0.070\,{\tiny$\pm$0.002} & 0.071\,{\tiny$\pm$0.003} & 0.135\,{\tiny$\pm$0.013} & 0.077\,{\tiny$\pm$0.001} & 0.082\,{\tiny$\pm$0.003} & 0.077\,{\tiny$\pm$0.001} & 0.086\,{\tiny$\pm$0.006} \\
fexofenadine\_mpo & 0.770\,{\tiny$\pm$0.015} & 0.614\,{\tiny$\pm$0.002} & 0.597\,{\tiny$\pm$0.001} & 0.636\,{\tiny$\pm$0.005} & 0.633\,{\tiny$\pm$0.001} & 0.615\,{\tiny$\pm$0.001} & 0.614\,{\tiny$\pm$0.002} & 0.638\,{\tiny$\pm$0.002} & 0.615\,{\tiny$\pm$0.010} & 0.645\,{\tiny$\pm$0.001} & 0.633\,{\tiny$\pm$0.002} & 0.634\,{\tiny$\pm$0.001} \\
gsk3b & 0.775\,{\tiny$\pm$0.068} & 0.179\,{\tiny$\pm$0.003} & 0.163\,{\tiny$\pm$0.013} & 0.448\,{\tiny$\pm$0.014} & 0.173\,{\tiny$\pm$0.001} & 0.179\,{\tiny$\pm$0.004} & 0.178\,{\tiny$\pm$0.003} & 0.174\,{\tiny$\pm$0.007} & 0.201\,{\tiny$\pm$0.013} & 0.467\,{\tiny$\pm$0.003} & 0.132\,{\tiny$\pm$0.008} & 0.131\,{\tiny$\pm$0.007} \\
isomers\_c7h8n2o2 & 0.522\,{\tiny$\pm$0.120} & 0.024\,{\tiny$\pm$0.002} & 0.024\,{\tiny$\pm$0.000} & 0.065\,{\tiny$\pm$0.012} & 0.018\,{\tiny$\pm$0.000} & 0.021\,{\tiny$\pm$0.001} & 0.023\,{\tiny$\pm$0.000} & 0.091\,{\tiny$\pm$0.001} & 0.023\,{\tiny$\pm$0.003} & 0.038\,{\tiny$\pm$0.000} & 0.026\,{\tiny$\pm$0.001} & 0.036\,{\tiny$\pm$0.002} \\
isomers\_c9h10n2o2pf2cl & 0.328\,{\tiny$\pm$0.110} & 0.043\,{\tiny$\pm$0.016} & 0.023\,{\tiny$\pm$0.003} & 0.012\,{\tiny$\pm$0.002} & 0.011\,{\tiny$\pm$0.000} & 0.023\,{\tiny$\pm$0.001} & 0.025\,{\tiny$\pm$0.001} & 0.063\,{\tiny$\pm$0.002} & 0.014\,{\tiny$\pm$0.001} & 0.013\,{\tiny$\pm$0.000} & 0.014\,{\tiny$\pm$0.000} & 0.017\,{\tiny$\pm$0.001} \\
jnk3 & 0.495\,{\tiny$\pm$0.098} & 0.106\,{\tiny$\pm$0.004} & 0.101\,{\tiny$\pm$0.005} & 0.239\,{\tiny$\pm$0.004} & 0.113\,{\tiny$\pm$0.004} & 0.102\,{\tiny$\pm$0.001} & 0.097\,{\tiny$\pm$0.004} & 0.081\,{\tiny$\pm$0.002} & 0.078\,{\tiny$\pm$0.003} & 0.224\,{\tiny$\pm$0.002} & 0.078\,{\tiny$\pm$0.003} & 0.080\,{\tiny$\pm$0.004} \\
median1 & 0.260\,{\tiny$\pm$0.029} & 0.146\,{\tiny$\pm$0.004} & 0.159\,{\tiny$\pm$0.001} & 0.127\,{\tiny$\pm$0.003} & 0.137\,{\tiny$\pm$0.002} & 0.149\,{\tiny$\pm$0.001} & 0.150\,{\tiny$\pm$0.000} & 0.142\,{\tiny$\pm$0.001} & 0.142\,{\tiny$\pm$0.000} & 0.135\,{\tiny$\pm$0.003} & 0.142\,{\tiny$\pm$0.000} & 0.142\,{\tiny$\pm$0.001} \\
median2 & 0.239\,{\tiny$\pm$0.016} & 0.164\,{\tiny$\pm$0.002} & 0.162\,{\tiny$\pm$0.001} & 0.154\,{\tiny$\pm$0.000} & 0.165\,{\tiny$\pm$0.002} & 0.162\,{\tiny$\pm$0.001} & 0.161\,{\tiny$\pm$0.001} & 0.160\,{\tiny$\pm$0.002} & 0.168\,{\tiny$\pm$0.000} & 0.148\,{\tiny$\pm$0.000} & 0.167\,{\tiny$\pm$0.000} & 0.166\,{\tiny$\pm$0.001} \\
mestranol\_similarity & 0.439\,{\tiny$\pm$0.034} & 0.310\,{\tiny$\pm$0.008} & 0.299\,{\tiny$\pm$0.001} & 0.249\,{\tiny$\pm$0.002} & 0.306\,{\tiny$\pm$0.003} & 0.312\,{\tiny$\pm$0.007} & 0.313\,{\tiny$\pm$0.007} & 0.302\,{\tiny$\pm$0.004} & 0.277\,{\tiny$\pm$0.001} & 0.231\,{\tiny$\pm$0.003} & 0.277\,{\tiny$\pm$0.001} & 0.277\,{\tiny$\pm$0.001} \\
osimertinib\_mpo & 0.830\,{\tiny$\pm$0.015} & 0.693\,{\tiny$\pm$0.006} & 0.725\,{\tiny$\pm$0.000} & 0.761\,{\tiny$\pm$0.002} & 0.707\,{\tiny$\pm$0.002} & 0.691\,{\tiny$\pm$0.004} & 0.688\,{\tiny$\pm$0.003} & 0.730\,{\tiny$\pm$0.003} & 0.726\,{\tiny$\pm$0.001} & 0.763\,{\tiny$\pm$0.001} & 0.728\,{\tiny$\pm$0.000} & 0.723\,{\tiny$\pm$0.005} \\
perindopril\_mpo & 0.535\,{\tiny$\pm$0.017} & 0.384\,{\tiny$\pm$0.001} & 0.389\,{\tiny$\pm$0.001} & 0.363\,{\tiny$\pm$0.002} & 0.406\,{\tiny$\pm$0.001} & 0.386\,{\tiny$\pm$0.000} & 0.387\,{\tiny$\pm$0.001} & 0.369\,{\tiny$\pm$0.003} & 0.401\,{\tiny$\pm$0.000} & 0.363\,{\tiny$\pm$0.004} & 0.401\,{\tiny$\pm$0.000} & 0.395\,{\tiny$\pm$0.000} \\
qed & 0.729\,{\tiny$\pm$0.020} & 0.923\,{\tiny$\pm$0.001} & 0.917\,{\tiny$\pm$0.002} & 0.814\,{\tiny$\pm$0.005} & 0.905\,{\tiny$\pm$0.002} & 0.924\,{\tiny$\pm$0.001} & 0.922\,{\tiny$\pm$0.000} & 0.909\,{\tiny$\pm$0.002} & 0.924\,{\tiny$\pm$0.002} & 0.803\,{\tiny$\pm$0.005} & 0.920\,{\tiny$\pm$0.001} & 0.913\,{\tiny$\pm$0.005} \\
ranolazine\_mpo & 0.781\,{\tiny$\pm$0.012} & 0.314\,{\tiny$\pm$0.021} & 0.276\,{\tiny$\pm$0.009} & 0.486\,{\tiny$\pm$0.028} & 0.260\,{\tiny$\pm$0.003} & 0.289\,{\tiny$\pm$0.002} & 0.287\,{\tiny$\pm$0.012} & 0.266\,{\tiny$\pm$0.018} & 0.237\,{\tiny$\pm$0.003} & 0.434\,{\tiny$\pm$0.008} & 0.238\,{\tiny$\pm$0.003} & 0.281\,{\tiny$\pm$0.005} \\
scaffold\_hop & 0.512\,{\tiny$\pm$0.011} & 0.435\,{\tiny$\pm$0.001} & 0.431\,{\tiny$\pm$0.002} & 0.437\,{\tiny$\pm$0.001} & 0.440\,{\tiny$\pm$0.001} & 0.435\,{\tiny$\pm$0.001} & 0.433\,{\tiny$\pm$0.001} & 0.423\,{\tiny$\pm$0.001} & 0.436\,{\tiny$\pm$0.001} & 0.435\,{\tiny$\pm$0.000} & 0.439\,{\tiny$\pm$0.002} & 0.434\,{\tiny$\pm$0.001} \\
sitagliptin\_mpo & 0.147\,{\tiny$\pm$0.037} & 0.004\,{\tiny$\pm$0.001} & 0.048\,{\tiny$\pm$0.011} & 0.000\,{\tiny$\pm$0.000} & 0.004\,{\tiny$\pm$0.000} & 0.002\,{\tiny$\pm$0.000} & 0.003\,{\tiny$\pm$0.001} & 0.006\,{\tiny$\pm$0.000} & 0.004\,{\tiny$\pm$0.000} & 0.001\,{\tiny$\pm$0.000} & 0.004\,{\tiny$\pm$0.000} & 0.005\,{\tiny$\pm$0.001} \\
thiothixene\_rediscovery & 0.503\,{\tiny$\pm$0.030} & 0.258\,{\tiny$\pm$0.000} & 0.269\,{\tiny$\pm$0.001} & 0.218\,{\tiny$\pm$0.002} & 0.249\,{\tiny$\pm$0.001} & 0.260\,{\tiny$\pm$0.002} & 0.255\,{\tiny$\pm$0.002} & 0.245\,{\tiny$\pm$0.003} & 0.258\,{\tiny$\pm$0.003} & 0.213\,{\tiny$\pm$0.000} & 0.250\,{\tiny$\pm$0.000} & 0.247\,{\tiny$\pm$0.000} \\
troglitazone\_rediscovery & 0.578\,{\tiny$\pm$0.107} & 0.213\,{\tiny$\pm$0.000} & 0.207\,{\tiny$\pm$0.000} & 0.166\,{\tiny$\pm$0.001} & 0.219\,{\tiny$\pm$0.001} & 0.213\,{\tiny$\pm$0.001} & 0.211\,{\tiny$\pm$0.002} & 0.192\,{\tiny$\pm$0.001} & 0.218\,{\tiny$\pm$0.002} & 0.165\,{\tiny$\pm$0.000} & 0.221\,{\tiny$\pm$0.001} & 0.218\,{\tiny$\pm$0.001} \\
valsartan\_smarts & \textbf{0.891}\,{\tiny$\pm$0.026} & 0.000\,{\tiny$\pm$0.000} & 0.000\,{\tiny$\pm$0.000} & 0.000\,{\tiny$\pm$0.000} & 0.000\,{\tiny$\pm$0.000} & 0.000\,{\tiny$\pm$0.000} & 0.000\,{\tiny$\pm$0.000} & 0.000\,{\tiny$\pm$0.000} & 0.000\,{\tiny$\pm$0.000} & 0.000\,{\tiny$\pm$0.000} & 0.000\,{\tiny$\pm$0.000} & 0.000\,{\tiny$\pm$0.000} \\
zaleplon\_mpo & 0.212\,{\tiny$\pm$0.037} & 0.012\,{\tiny$\pm$0.005} & 0.002\,{\tiny$\pm$0.000} & 0.001\,{\tiny$\pm$0.000} & 0.003\,{\tiny$\pm$0.000} & 0.006\,{\tiny$\pm$0.000} & 0.010\,{\tiny$\pm$0.003} & 0.012\,{\tiny$\pm$0.002} & 0.004\,{\tiny$\pm$0.000} & 0.001\,{\tiny$\pm$0.000} & 0.004\,{\tiny$\pm$0.000} & 0.009\,{\tiny$\pm$0.005} \\
\midrule
Mean & $0.596$ & $0.288$ & $0.289$ & $0.295$ & $0.283$ & $0.285$ & $0.284$ & $0.288$ & $0.283$ & $0.287$ & $0.281$ & $0.282$ \\
Oracles won & 3 & 0 & 0 & 0 & 0 & 0 & 0 & 0 & 0 & 0 & 0 & 0 \\
\bottomrule
\end{tabular}
\end{table}
\end{landscape}

\begin{landscape}
\begin{table}[htbp]
\centering
\scriptsize
\setlength{\tabcolsep}{3pt}
\caption{\textbf{Top-10 mean at a 210-call budget, Opus 5 multi-turn versus the full method panel (part c of 3).} Rows are the 23 PMO oracles; columns are Opus 5 multi-turn (reference, repeated in every part) and methods 23--33 of 33, in the same order as the Top-10 AUC tables (Opus 5 multi-turn set off by a vertical rule). Each cell is mean $\pm$1 SE over seeds; higher is better; \textbf{bold} = best over the full 34-way panel for that oracle. The footer gives the per-column macro-average and the number of oracles won over the full panel.}
\label{tab:opus-panel-mean-c}
\begin{tabular}{@{}l c!{\vrule} *{11}{c}@{}}
\toprule
Oracle & \rotatebox{90}{Opus 5 (multi-turn)} & \rotatebox{90}{SMILES BAR} & \rotatebox{90}{screening} & \rotatebox{90}{selfies\_vae\_bo} & \rotatebox{90}{gp\_bo} & \rotatebox{90}{dst} & \rotatebox{90}{mol\_pal} & \rotatebox{90}{mars} & \rotatebox{90}{graph\_mcts} & \rotatebox{90}{MolGAN} & \rotatebox{90}{moldqn} & \rotatebox{90}{selfies\_ga} \\
\midrule
albuterol\_similarity & \textbf{1.000}\,{\tiny$\pm$0.000} & 0.377\,{\tiny$\pm$0.003} & 0.403\,{\tiny$\pm$0.002} & 0.386\,{\tiny$\pm$0.002} & 0.379\,{\tiny$\pm$0.003} & 0.396\,{\tiny$\pm$0.006} & 0.409\,{\tiny$\pm$0.000} & 0.473\,{\tiny$\pm$0.005} & 0.514\,{\tiny$\pm$0.010} & 0.302\,{\tiny$\pm$0.004} & 0.237\,{\tiny$\pm$0.000} & 0.226\,{\tiny$\pm$0.005} \\
amlodipine\_mpo & 0.564\,{\tiny$\pm$0.011} & 0.458\,{\tiny$\pm$0.002} & 0.467\,{\tiny$\pm$0.001} & 0.473\,{\tiny$\pm$0.004} & 0.469\,{\tiny$\pm$0.000} & 0.469\,{\tiny$\pm$0.003} & 0.482\,{\tiny$\pm$0.001} & 0.427\,{\tiny$\pm$0.006} & 0.404\,{\tiny$\pm$0.005} & 0.213\,{\tiny$\pm$0.003} & 0.197\,{\tiny$\pm$0.009} & 0.006\,{\tiny$\pm$0.003} \\
celecoxib\_rediscovery & \textbf{1.000}\,{\tiny$\pm$0.000} & 0.247\,{\tiny$\pm$0.004} & 0.273\,{\tiny$\pm$0.005} & 0.271\,{\tiny$\pm$0.001} & 0.274\,{\tiny$\pm$0.004} & 0.266\,{\tiny$\pm$0.002} & 0.280\,{\tiny$\pm$0.000} & 0.230\,{\tiny$\pm$0.005} & 0.163\,{\tiny$\pm$0.006} & 0.095\,{\tiny$\pm$0.001} & 0.085\,{\tiny$\pm$0.002} & 0.049\,{\tiny$\pm$0.006} \\
deco\_hop & 0.661\,{\tiny$\pm$0.056} & 0.564\,{\tiny$\pm$0.000} & 0.568\,{\tiny$\pm$0.000} & 0.570\,{\tiny$\pm$0.001} & 0.570\,{\tiny$\pm$0.000} & 0.566\,{\tiny$\pm$0.002} & 0.562\,{\tiny$\pm$0.000} & 0.569\,{\tiny$\pm$0.001} & 0.536\,{\tiny$\pm$0.004} & 0.510\,{\tiny$\pm$0.000} & 0.526\,{\tiny$\pm$0.001} & 0.502\,{\tiny$\pm$0.000} \\
drd2 & 0.945\,{\tiny$\pm$0.028} & 0.213\,{\tiny$\pm$0.019} & 0.051\,{\tiny$\pm$0.003} & 0.077\,{\tiny$\pm$0.001} & 0.099\,{\tiny$\pm$0.006} & 0.100\,{\tiny$\pm$0.009} & 0.080\,{\tiny$\pm$0.007} & 0.134\,{\tiny$\pm$0.042} & 0.053\,{\tiny$\pm$0.006} & 0.024\,{\tiny$\pm$0.001} & 0.011\,{\tiny$\pm$0.000} & 0.018\,{\tiny$\pm$0.001} \\
fexofenadine\_mpo & 0.770\,{\tiny$\pm$0.015} & 0.608\,{\tiny$\pm$0.003} & 0.616\,{\tiny$\pm$0.001} & 0.633\,{\tiny$\pm$0.002} & 0.607\,{\tiny$\pm$0.000} & 0.629\,{\tiny$\pm$0.005} & 0.612\,{\tiny$\pm$0.001} & 0.549\,{\tiny$\pm$0.004} & 0.465\,{\tiny$\pm$0.006} & 0.221\,{\tiny$\pm$0.007} & 0.356\,{\tiny$\pm$0.023} & 0.005\,{\tiny$\pm$0.001} \\
gsk3b & 0.775\,{\tiny$\pm$0.068} & 0.151\,{\tiny$\pm$0.002} & 0.188\,{\tiny$\pm$0.001} & 0.132\,{\tiny$\pm$0.008} & 0.178\,{\tiny$\pm$0.003} & 0.168\,{\tiny$\pm$0.032} & 0.126\,{\tiny$\pm$0.003} & 0.212\,{\tiny$\pm$0.006} & 0.130\,{\tiny$\pm$0.007} & 0.101\,{\tiny$\pm$0.006} & 0.119\,{\tiny$\pm$0.001} & 0.087\,{\tiny$\pm$0.002} \\
isomers\_c7h8n2o2 & 0.522\,{\tiny$\pm$0.120} & 0.007\,{\tiny$\pm$0.002} & 0.006\,{\tiny$\pm$0.000} & 0.026\,{\tiny$\pm$0.001} & 0.028\,{\tiny$\pm$0.001} & 0.022\,{\tiny$\pm$0.006} & 0.016\,{\tiny$\pm$0.002} & 0.029\,{\tiny$\pm$0.007} & 0.119\,{\tiny$\pm$0.032} & 0.619\,{\tiny$\pm$0.015} & 0.104\,{\tiny$\pm$0.011} & 0.043\,{\tiny$\pm$0.003} \\
isomers\_c9h10n2o2pf2cl & 0.328\,{\tiny$\pm$0.110} & 0.013\,{\tiny$\pm$0.001} & 0.026\,{\tiny$\pm$0.001} & 0.014\,{\tiny$\pm$0.001} & 0.016\,{\tiny$\pm$0.007} & 0.012\,{\tiny$\pm$0.002} & 0.011\,{\tiny$\pm$0.001} & 0.004\,{\tiny$\pm$0.001} & 0.078\,{\tiny$\pm$0.014} & 0.343\,{\tiny$\pm$0.007} & 0.045\,{\tiny$\pm$0.009} & 0.007\,{\tiny$\pm$0.000} \\
jnk3 & 0.495\,{\tiny$\pm$0.098} & 0.094\,{\tiny$\pm$0.002} & 0.094\,{\tiny$\pm$0.001} & 0.078\,{\tiny$\pm$0.003} & 0.100\,{\tiny$\pm$0.001} & 0.086\,{\tiny$\pm$0.007} & 0.083\,{\tiny$\pm$0.003} & 0.113\,{\tiny$\pm$0.006} & 0.041\,{\tiny$\pm$0.004} & 0.033\,{\tiny$\pm$0.002} & 0.046\,{\tiny$\pm$0.002} & 0.021\,{\tiny$\pm$0.001} \\
median1 & 0.260\,{\tiny$\pm$0.029} & 0.166\,{\tiny$\pm$0.001} & 0.151\,{\tiny$\pm$0.001} & 0.143\,{\tiny$\pm$0.001} & 0.141\,{\tiny$\pm$0.001} & 0.164\,{\tiny$\pm$0.007} & 0.143\,{\tiny$\pm$0.000} & 0.170\,{\tiny$\pm$0.005} & 0.133\,{\tiny$\pm$0.005} & 0.180\,{\tiny$\pm$0.003} & 0.071\,{\tiny$\pm$0.002} & 0.059\,{\tiny$\pm$0.001} \\
median2 & 0.239\,{\tiny$\pm$0.016} & 0.159\,{\tiny$\pm$0.001} & 0.169\,{\tiny$\pm$0.001} & 0.167\,{\tiny$\pm$0.000} & 0.166\,{\tiny$\pm$0.000} & 0.153\,{\tiny$\pm$0.012} & 0.158\,{\tiny$\pm$0.001} & 0.145\,{\tiny$\pm$0.002} & 0.103\,{\tiny$\pm$0.003} & 0.075\,{\tiny$\pm$0.001} & 0.079\,{\tiny$\pm$0.000} & 0.034\,{\tiny$\pm$0.002} \\
mestranol\_similarity & 0.439\,{\tiny$\pm$0.034} & 0.305\,{\tiny$\pm$0.003} & 0.296\,{\tiny$\pm$0.001} & 0.277\,{\tiny$\pm$0.001} & 0.303\,{\tiny$\pm$0.002} & 0.271\,{\tiny$\pm$0.004} & 0.312\,{\tiny$\pm$0.005} & 0.292\,{\tiny$\pm$0.005} & 0.180\,{\tiny$\pm$0.005} & 0.134\,{\tiny$\pm$0.001} & 0.094\,{\tiny$\pm$0.001} & 0.072\,{\tiny$\pm$0.002} \\
osimertinib\_mpo & 0.830\,{\tiny$\pm$0.015} & 0.710\,{\tiny$\pm$0.002} & 0.690\,{\tiny$\pm$0.003} & 0.726\,{\tiny$\pm$0.001} & 0.710\,{\tiny$\pm$0.001} & 0.719\,{\tiny$\pm$0.010} & 0.691\,{\tiny$\pm$0.009} & 0.564\,{\tiny$\pm$0.017} & 0.661\,{\tiny$\pm$0.006} & 0.368\,{\tiny$\pm$0.006} & 0.617\,{\tiny$\pm$0.004} & 0.061\,{\tiny$\pm$0.017} \\
perindopril\_mpo & 0.535\,{\tiny$\pm$0.017} & 0.377\,{\tiny$\pm$0.001} & 0.391\,{\tiny$\pm$0.001} & 0.401\,{\tiny$\pm$0.000} & 0.388\,{\tiny$\pm$0.000} & 0.394\,{\tiny$\pm$0.003} & 0.387\,{\tiny$\pm$0.004} & 0.380\,{\tiny$\pm$0.008} & 0.142\,{\tiny$\pm$0.012} & 0.084\,{\tiny$\pm$0.002} & 0.018\,{\tiny$\pm$0.000} & 0.008\,{\tiny$\pm$0.002} \\
qed & 0.729\,{\tiny$\pm$0.020} & 0.922\,{\tiny$\pm$0.002} & 0.928\,{\tiny$\pm$0.001} & 0.921\,{\tiny$\pm$0.001} & 0.922\,{\tiny$\pm$0.001} & 0.924\,{\tiny$\pm$0.002} & 0.929\,{\tiny$\pm$0.001} & 0.883\,{\tiny$\pm$0.004} & 0.815\,{\tiny$\pm$0.008} & 0.609\,{\tiny$\pm$0.003} & 0.573\,{\tiny$\pm$0.027} & 0.476\,{\tiny$\pm$0.004} \\
ranolazine\_mpo & 0.781\,{\tiny$\pm$0.012} & 0.221\,{\tiny$\pm$0.004} & 0.263\,{\tiny$\pm$0.002} & 0.235\,{\tiny$\pm$0.001} & 0.274\,{\tiny$\pm$0.002} & 0.223\,{\tiny$\pm$0.047} & 0.209\,{\tiny$\pm$0.002} & 0.438\,{\tiny$\pm$0.037} & 0.061\,{\tiny$\pm$0.013} & 0.004\,{\tiny$\pm$0.000} & 0.004\,{\tiny$\pm$0.000} & 0.004\,{\tiny$\pm$0.000} \\
scaffold\_hop & 0.512\,{\tiny$\pm$0.011} & 0.430\,{\tiny$\pm$0.001} & 0.436\,{\tiny$\pm$0.000} & 0.439\,{\tiny$\pm$0.002} & 0.428\,{\tiny$\pm$0.001} & 0.434\,{\tiny$\pm$0.003} & 0.431\,{\tiny$\pm$0.001} & 0.438\,{\tiny$\pm$0.002} & 0.382\,{\tiny$\pm$0.001} & 0.349\,{\tiny$\pm$0.001} & 0.367\,{\tiny$\pm$0.001} & 0.337\,{\tiny$\pm$0.000} \\
sitagliptin\_mpo & 0.147\,{\tiny$\pm$0.037} & 0.067\,{\tiny$\pm$0.003} & 0.003\,{\tiny$\pm$0.000} & 0.004\,{\tiny$\pm$0.000} & 0.002\,{\tiny$\pm$0.000} & 0.005\,{\tiny$\pm$0.000} & 0.006\,{\tiny$\pm$0.000} & 0.004\,{\tiny$\pm$0.000} & 0.001\,{\tiny$\pm$0.000} & 0.018\,{\tiny$\pm$0.004} & 0.000\,{\tiny$\pm$0.000} & 0.000\,{\tiny$\pm$0.000} \\
thiothixene\_rediscovery & 0.503\,{\tiny$\pm$0.030} & 0.248\,{\tiny$\pm$0.002} & 0.256\,{\tiny$\pm$0.002} & 0.250\,{\tiny$\pm$0.000} & 0.265\,{\tiny$\pm$0.001} & 0.252\,{\tiny$\pm$0.002} & 0.243\,{\tiny$\pm$0.002} & 0.217\,{\tiny$\pm$0.003} & 0.190\,{\tiny$\pm$0.005} & 0.085\,{\tiny$\pm$0.001} & 0.061\,{\tiny$\pm$0.007} & 0.060\,{\tiny$\pm$0.003} \\
troglitazone\_rediscovery & 0.578\,{\tiny$\pm$0.107} & 0.218\,{\tiny$\pm$0.003} & 0.211\,{\tiny$\pm$0.001} & 0.221\,{\tiny$\pm$0.001} & 0.211\,{\tiny$\pm$0.001} & 0.205\,{\tiny$\pm$0.016} & 0.207\,{\tiny$\pm$0.004} & 0.200\,{\tiny$\pm$0.002} & 0.186\,{\tiny$\pm$0.003} & 0.110\,{\tiny$\pm$0.002} & 0.069\,{\tiny$\pm$0.004} & 0.052\,{\tiny$\pm$0.006} \\
valsartan\_smarts & \textbf{0.891}\,{\tiny$\pm$0.026} & 0.000\,{\tiny$\pm$0.000} & 0.000\,{\tiny$\pm$0.000} & 0.000\,{\tiny$\pm$0.000} & 0.000\,{\tiny$\pm$0.000} & 0.000\,{\tiny$\pm$0.000} & 0.000\,{\tiny$\pm$0.000} & 0.000\,{\tiny$\pm$0.000} & 0.000\,{\tiny$\pm$0.000} & 0.000\,{\tiny$\pm$0.000} & 0.000\,{\tiny$\pm$0.000} & 0.000\,{\tiny$\pm$0.000} \\
zaleplon\_mpo & 0.212\,{\tiny$\pm$0.037} & 0.011\,{\tiny$\pm$0.005} & 0.002\,{\tiny$\pm$0.000} & 0.004\,{\tiny$\pm$0.000} & 0.012\,{\tiny$\pm$0.001} & 0.003\,{\tiny$\pm$0.001} & 0.013\,{\tiny$\pm$0.003} & 0.028\,{\tiny$\pm$0.000} & 0.004\,{\tiny$\pm$0.002} & 0.000\,{\tiny$\pm$0.000} & 0.000\,{\tiny$\pm$0.000} & 0.000\,{\tiny$\pm$0.000} \\
\midrule
Mean & $0.596$ & $0.286$ & $0.282$ & $0.280$ & $0.284$ & $0.281$ & $0.278$ & $0.282$ & $0.233$ & $0.195$ & $0.160$ & $0.092$ \\
Oracles won & 3 & 0 & 0 & 0 & 0 & 0 & 0 & 0 & 0 & 0 & 0 & 0 \\
\bottomrule
\end{tabular}
\end{table}
\end{landscape}

\begin{table}[htbp]
\centering
\scriptsize
\setlength{\tabcolsep}{3pt}
\caption{\textbf{Fold oracle-call budget to reach Opus 5 multi-turn (part a of 3).} For each method (columns; methods 1--11 of 33, same order as the Top-10 AUC tables) and oracle (rows), the fold increase in oracle-call budget---relative to the LLM's 210 calls---needed for the method's running Top-10 mean to reach Opus 5 multi-turn's Top-10 mean@210, taken as the \emph{median over seeds} of the first step at which the bar is reached. Lower is more sample efficient; a value below $1\times$ reaches Opus 5's level in fewer calls (and $<0.01\times$ means it is reached at essentially the first oracle call); \texttt{--} = the method never reaches Opus 5's level within the run it actually did. The footer counts, per method, how many of the 23 oracles it beats.}
\label{tab:opus-panel-fold-a}
\begin{tabular}{@{}l *{11}{c}@{}}
\toprule
Oracle & \rotatebox{90}{GenMol} & \rotatebox{90}{ExLLM (Gemini 2.5 Flash)} & \rotatebox{90}{MolLEO (BioT5)} & \rotatebox{90}{synnet} & \rotatebox{90}{dog\_gen} & \rotatebox{90}{dog\_ae} & \rotatebox{90}{graph\_ga} & \rotatebox{90}{SMILES Aug. Memory} & \rotatebox{90}{reinvent} & \rotatebox{90}{REINVENT-Transformer} & \rotatebox{90}{smiles\_lstm\_hc} \\
\midrule
albuterol\_similarity & -- & $0.54\times$ & $12.17\times$ & -- & $44.83\times$ & -- & $41\times$ & $11.94\times$ & -- & $15.94\times$ & $43.91\times$ \\
amlodipine\_mpo & $0.41\times$ & $0.51\times$ & $1.37\times$ & $12.66\times$ & $32.56\times$ & -- & $10.05\times$ & $8.36\times$ & $11.23\times$ & $8.09\times$ & $18.9\times$ \\
celecoxib\_rediscovery & -- & $4.83\times$ & -- & -- & -- & -- & -- & -- & -- & -- & -- \\
deco\_hop & $0.12\times$ & $0.5\times$ & $8.02\times$ & -- & $14.54\times$ & $4.92\times$ & $42.84\times$ & $21.42\times$ & $28.65\times$ & $24.79\times$ & $6.86\times$ \\
drd2 & $0.16\times$ & $0.83\times$ & $1.54\times$ & $2.84\times$ & $6.94\times$ & $4.28\times$ & $4.23\times$ & $2.86\times$ & $4.33\times$ & $3.57\times$ & $9.89\times$ \\
fexofenadine\_mpo & $0.82\times$ & $0.54\times$ & $12.88\times$ & $16.92\times$ & $45.85\times$ & -- & $20.56\times$ & $9.31\times$ & $19.42\times$ & $14.3\times$ & $38.6\times$ \\
gsk3b & $0.16\times$ & $1.61\times$ & $6.73\times$ & $14.75\times$ & $18.4\times$ & -- & $20.89\times$ & $5.85\times$ & $8.91\times$ & $6.16\times$ & $12.3\times$ \\
isomers\_c7h8n2o2 & $2.68\times$ & $0.5\times$ & $0.9\times$ & $18.77\times$ & $26.1\times$ & -- & $2.51\times$ & $3.63\times$ & $7.31\times$ & $4.78\times$ & $25.82\times$ \\
isomers\_c9h10n2o2pf2cl & $0.99\times$ & $0.48\times$ & $0.61\times$ & $18.6\times$ & $34.9\times$ & -- & $3.5\times$ & $5.51\times$ & $8.22\times$ & $10.3\times$ & $22.99\times$ \\
jnk3 & $<0.01\times$ & $<0.01\times$ & $2.92\times$ & $6.8\times$ & $15.92\times$ & $4.35\times$ & $17.15\times$ & $5.81\times$ & $7.03\times$ & $6.53\times$ & $9.99\times$ \\
median1 & $0.23\times$ & $0.52\times$ & $2.32\times$ & -- & $33.17\times$ & -- & $8.13\times$ & $6.24\times$ & $6.94\times$ & $6.1\times$ & $23.34\times$ \\
median2 & $<0.01\times$ & $0.51\times$ & $4.65\times$ & $12.96\times$ & $33.4\times$ & -- & $10.15\times$ & $7.37\times$ & $8.35\times$ & $7.78\times$ & $19.63\times$ \\
mestranol\_similarity & $0.14\times$ & $0.51\times$ & $3.54\times$ & $17.38\times$ & $26.17\times$ & -- & $7.74\times$ & $5.89\times$ & $6.41\times$ & $6.81\times$ & $12.94\times$ \\
osimertinib\_mpo & $0.56\times$ & $1\times$ & $9.7\times$ & -- & $42.57\times$ & -- & $17.26\times$ & $9.59\times$ & $20.07\times$ & $16.2\times$ & $38.3\times$ \\
perindopril\_mpo & $0.08\times$ & $0.57\times$ & $0.89\times$ & $10.93\times$ & $39.09\times$ & -- & $18.32\times$ & $12.69\times$ & $19.24\times$ & $19.33\times$ & $40.79\times$ \\
qed & $4.2\times$ & $0.75\times$ & $5.82\times$ & $5.52\times$ & $21.56\times$ & -- & $10.06\times$ & $5.97\times$ & $7\times$ & $8.24\times$ & $15.31\times$ \\
ranolazine\_mpo & $2.97\times$ & $0.89\times$ & $12.96\times$ & $22.69\times$ & $35.48\times$ & -- & $33.3\times$ & $8.32\times$ & $17.33\times$ & $13.73\times$ & $38.09\times$ \\
scaffold\_hop & $0.22\times$ & $0.5\times$ & $7.19\times$ & $19.01\times$ & $24.1\times$ & -- & $14.16\times$ & $8.21\times$ & $14.1\times$ & $13.9\times$ & $19.32\times$ \\
sitagliptin\_mpo & $0.81\times$ & $1.37\times$ & $1.85\times$ & -- & $47.31\times$ & -- & $10.72\times$ & $1.89\times$ & -- & $3.47\times$ & -- \\
thiothixene\_rediscovery & $2.26\times$ & $0.51\times$ & $6.67\times$ & $36.19\times$ & $43.97\times$ & -- & $20.72\times$ & $11.7\times$ & $17.52\times$ & $16.6\times$ & $33.27\times$ \\
troglitazone\_rediscovery & $0.79\times$ & $1.33\times$ & -- & -- & $41.92\times$ & -- & -- & $18.71\times$ & $29.05\times$ & $27.3\times$ & $42.46\times$ \\
valsartan\_smarts & $32.24\times$ & $9.67\times$ & -- & -- & -- & -- & -- & -- & $6.13\times$ & -- & -- \\
zaleplon\_mpo & $0.12\times$ & $0.06\times$ & $0.24\times$ & $2.84\times$ & $36.74\times$ & -- & $4.56\times$ & $5.63\times$ & $7.13\times$ & $12.93\times$ & $25.08\times$ \\
\midrule
Oracles beaten & 21/23 & 23/23 & 20/23 & 15/23 & 21/23 & 3/23 & 20/23 & 21/23 & 20/23 & 21/23 & 20/23 \\
\bottomrule
\end{tabular}
\end{table}

\begin{table}[htbp]
\centering
\scriptsize
\setlength{\tabcolsep}{3pt}
\caption{\textbf{Fold oracle-call budget to reach Opus 5 multi-turn (part b of 3).} For each method (columns; methods 12--22 of 33, same order as the Top-10 AUC tables) and oracle (rows), the fold increase in oracle-call budget---relative to the LLM's 210 calls---needed for the method's running Top-10 mean to reach Opus 5 multi-turn's Top-10 mean@210, taken as the \emph{median over seeds} of the first step at which the bar is reached. Lower is more sample efficient; a value below $1\times$ reaches Opus 5's level in fewer calls (and $<0.01\times$ means it is reached at essentially the first oracle call); \texttt{--} = the method never reaches Opus 5's level within the run it actually did. The footer counts, per method, how many of the 23 oracles it beats.}
\label{tab:opus-panel-fold-b}
\begin{tabular}{@{}l *{11}{c}@{}}
\toprule
Oracle & \rotatebox{90}{smiles\_ga} & \rotatebox{90}{SMILES AHC} & \rotatebox{90}{gflownet} & \rotatebox{90}{pasithea} & \rotatebox{90}{stoned} & \rotatebox{90}{selfies\_lstm\_hc} & \rotatebox{90}{reinvent\_selfies} & \rotatebox{90}{mimosa} & \rotatebox{90}{gflownet\_al} & \rotatebox{90}{smiles\_vae\_bo} & \rotatebox{90}{jt\_vae\_bo} \\
\midrule
albuterol\_similarity & -- & -- & -- & -- & -- & $47.37\times$ & $23\times$ & -- & -- & -- & -- \\
amlodipine\_mpo & $10.68\times$ & $16.32\times$ & -- & -- & $4.79\times$ & $39.15\times$ & $13.84\times$ & $31.07\times$ & -- & $40.99\times$ & -- \\
celecoxib\_rediscovery & -- & -- & -- & -- & -- & -- & -- & -- & -- & -- & -- \\
deco\_hop & -- & $35.31\times$ & -- & -- & -- & -- & $34.8\times$ & -- & -- & -- & -- \\
drd2 & $7.37\times$ & $7.93\times$ & -- & -- & $6.67\times$ & $27.41\times$ & $4.75\times$ & $15.15\times$ & -- & -- & -- \\
fexofenadine\_mpo & $30.52\times$ & $26.07\times$ & -- & -- & $8.93\times$ & -- & $32.75\times$ & -- & -- & -- & -- \\
gsk3b & -- & $12.96\times$ & -- & -- & $30.49\times$ & -- & $13.72\times$ & -- & -- & -- & -- \\
isomers\_c7h8n2o2 & $3.1\times$ & $17.19\times$ & $35.2\times$ & $6.61\times$ & $4.12\times$ & $14.34\times$ & $6.1\times$ & $15.75\times$ & -- & -- & -- \\
isomers\_c9h10n2o2pf2cl & $3.81\times$ & $27.34\times$ & -- & $7.95\times$ & $3.9\times$ & $20.43\times$ & $6.76\times$ & $18.33\times$ & -- & -- & -- \\
jnk3 & -- & $18.78\times$ & $36.77\times$ & -- & $8.21\times$ & -- & $11.57\times$ & $18.71\times$ & -- & -- & -- \\
median1 & -- & $7.85\times$ & -- & -- & $6\times$ & $31.2\times$ & $7.98\times$ & $18.31\times$ & -- & -- & -- \\
median2 & -- & $10.79\times$ & -- & -- & $6.77\times$ & $36.95\times$ & $17.74\times$ & -- & -- & -- & -- \\
mestranol\_similarity & $10.61\times$ & $9.23\times$ & -- & -- & $4.82\times$ & $21.55\times$ & $8.04\times$ & $14.81\times$ & -- & $45.95\times$ & -- \\
osimertinib\_mpo & $9.08\times$ & $30.28\times$ & -- & -- & $13.86\times$ & $37.8\times$ & $26.89\times$ & $24.64\times$ & -- & -- & -- \\
perindopril\_mpo & -- & $37.03\times$ & -- & -- & -- & -- & $25.23\times$ & $31.02\times$ & -- & -- & -- \\
qed & $7.92\times$ & $8.73\times$ & -- & -- & $4.78\times$ & $27.23\times$ & $11.6\times$ & $12.94\times$ & -- & $13.47\times$ & $18.01\times$ \\
ranolazine\_mpo & $32.69\times$ & $25.12\times$ & -- & -- & $13.51\times$ & $46.36\times$ & $22.99\times$ & $24.8\times$ & -- & -- & -- \\
scaffold\_hop & $27.89\times$ & $15.27\times$ & -- & -- & $10.69\times$ & $44.04\times$ & $21.32\times$ & $13.79\times$ & -- & -- & -- \\
sitagliptin\_mpo & $7.43\times$ & -- & -- & -- & $7.97\times$ & -- & $31.47\times$ & $31.09\times$ & -- & -- & -- \\
thiothixene\_rediscovery & -- & $20.4\times$ & -- & -- & -- & -- & $22.38\times$ & -- & -- & -- & -- \\
troglitazone\_rediscovery & -- & $37.55\times$ & -- & -- & -- & -- & -- & -- & -- & -- & -- \\
valsartan\_smarts & -- & -- & -- & -- & -- & -- & -- & -- & -- & -- & -- \\
zaleplon\_mpo & $7.12\times$ & $34.98\times$ & -- & -- & $4.73\times$ & $17.89\times$ & $9.68\times$ & $21.65\times$ & -- & -- & -- \\
\midrule
Oracles beaten & 12/23 & 19/23 & 2/23 & 2/23 & 16/23 & 13/23 & 20/23 & 14/23 & 0/23 & 3/23 & 1/23 \\
\bottomrule
\end{tabular}
\end{table}

\begin{table}[htbp]
\centering
\scriptsize
\setlength{\tabcolsep}{3pt}
\caption{\textbf{Fold oracle-call budget to reach Opus 5 multi-turn (part c of 3).} For each method (columns; methods 23--33 of 33, same order as the Top-10 AUC tables) and oracle (rows), the fold increase in oracle-call budget---relative to the LLM's 210 calls---needed for the method's running Top-10 mean to reach Opus 5 multi-turn's Top-10 mean@210, taken as the \emph{median over seeds} of the first step at which the bar is reached. Lower is more sample efficient; a value below $1\times$ reaches Opus 5's level in fewer calls (and $<0.01\times$ means it is reached at essentially the first oracle call); \texttt{--} = the method never reaches Opus 5's level within the run it actually did. The footer counts, per method, how many of the 23 oracles it beats.}
\label{tab:opus-panel-fold-c}
\begin{tabular}{@{}l *{11}{c}@{}}
\toprule
Oracle & \rotatebox{90}{SMILES BAR} & \rotatebox{90}{screening} & \rotatebox{90}{selfies\_vae\_bo} & \rotatebox{90}{gp\_bo} & \rotatebox{90}{dst} & \rotatebox{90}{mol\_pal} & \rotatebox{90}{mars} & \rotatebox{90}{graph\_mcts} & \rotatebox{90}{MolGAN} & \rotatebox{90}{moldqn} & \rotatebox{90}{selfies\_ga} \\
\midrule
albuterol\_similarity & -- & -- & -- & $25.54\times$ & -- & -- & -- & -- & -- & -- & -- \\
amlodipine\_mpo & $29.52\times$ & $35.91\times$ & -- & $24.46\times$ & -- & $10.95\times$ & -- & -- & -- & -- & -- \\
celecoxib\_rediscovery & -- & -- & -- & -- & -- & -- & -- & -- & -- & -- & -- \\
deco\_hop & -- & -- & -- & $35.4\times$ & -- & $12.4\times$ & -- & -- & -- & -- & -- \\
drd2 & $8.62\times$ & -- & -- & $7.74\times$ & $16.04\times$ & -- & $14.94\times$ & -- & -- & -- & $28.89\times$ \\
fexofenadine\_mpo & $43.6\times$ & -- & -- & $36.82\times$ & $22.35\times$ & -- & -- & -- & -- & -- & -- \\
gsk3b & $20.1\times$ & -- & -- & $9.02\times$ & $19.66\times$ & -- & -- & -- & -- & -- & -- \\
isomers\_c7h8n2o2 & $10.02\times$ & -- & -- & $7.22\times$ & $18.16\times$ & $7.56\times$ & $9.76\times$ & $12.48\times$ & $0.53\times$ & $39.76\times$ & $6.26\times$ \\
isomers\_c9h10n2o2pf2cl & $10.95\times$ & -- & $41.99\times$ & $7.33\times$ & $18.2\times$ & -- & $11.43\times$ & $6.25\times$ & $0.74\times$ & $11.24\times$ & $7.5\times$ \\
jnk3 & $19.98\times$ & -- & -- & $11.49\times$ & $14.96\times$ & -- & $15.43\times$ & -- & -- & -- & -- \\
median1 & $16.39\times$ & -- & -- & $3.15\times$ & $25.52\times$ & -- & -- & -- & $38.05\times$ & -- & -- \\
median2 & $26.65\times$ & -- & -- & $3\times$ & -- & -- & -- & -- & -- & -- & -- \\
mestranol\_similarity & $14.12\times$ & $25.42\times$ & -- & $4.08\times$ & $13.23\times$ & $3.15\times$ & $29.21\times$ & -- & -- & -- & $30.72\times$ \\
osimertinib\_mpo & $44.01\times$ & -- & -- & $42.85\times$ & $38.28\times$ & -- & -- & -- & -- & -- & -- \\
perindopril\_mpo & -- & -- & -- & $41.02\times$ & -- & -- & -- & -- & -- & -- & -- \\
qed & $10.04\times$ & $22.16\times$ & $24.55\times$ & $25.73\times$ & $14.02\times$ & $9.95\times$ & -- & -- & -- & -- & $30.07\times$ \\
ranolazine\_mpo & $31.19\times$ & -- & -- & $26.55\times$ & $22.46\times$ & -- & $14.91\times$ & -- & -- & -- & $38.6\times$ \\
scaffold\_hop & $34.57\times$ & -- & -- & $13.4\times$ & $19.22\times$ & -- & -- & -- & -- & -- & -- \\
sitagliptin\_mpo & $3.49\times$ & -- & -- & $22.25\times$ & -- & -- & -- & -- & -- & -- & $17.16\times$ \\
thiothixene\_rediscovery & $40.14\times$ & -- & -- & $8.7\times$ & -- & -- & -- & -- & -- & -- & -- \\
troglitazone\_rediscovery & -- & -- & -- & -- & -- & -- & -- & -- & -- & -- & -- \\
valsartan\_smarts & -- & -- & -- & -- & -- & -- & -- & -- & -- & -- & -- \\
zaleplon\_mpo & $9.67\times$ & -- & $23.66\times$ & $11.03\times$ & $19.07\times$ & -- & $9.87\times$ & -- & -- & -- & $13.16\times$ \\
\midrule
Oracles beaten & 17/23 & 3/23 & 3/23 & 20/23 & 13/23 & 5/23 & 7/23 & 2/23 & 3/23 & 2/23 & 8/23 \\
\bottomrule
\end{tabular}
\end{table}

\clearpage

\captionof{lstlisting}{Synthetic function optimization: a representative rendered prompt for the multi-turn opening turn (here a two-dimensional task with batch size $q=1$). The system prompt fixes the objective-agnostic framing (the model is never told the identity of the test function, its optimum, or $f^\star$), the normalized $[0,1]^d$ search space, and the JSON response format; the appended clause establishes the stateful multi-turn session. The user prompt serializes the initial design as \texttt{\{"x": [...], "y": ...\}} records, states the current best and remaining batch budget, and requests the next batch. The initial-design list is illustrative and abbreviated.}
\label{lst:prompt_synthetic}
\begin{lstlisting}[nolol, escapechar=¤]
¤\tprompt{SYSTEM PROMPT}¤
You are an optimization assistant. Your task is to propose 1 candidate point in the search space described below.

Your job is to maximize the objective function.

The search space has 2 continuous dimensions, all bounded in [0, 1].

Respond with a JSON object containing a single key "candidates" whose value is a list of exactly 1 candidate. Each candidate is a list of 2 floating-point numbers in [0, 1].

Example response format:
{"candidates": [[0.5, 0.55]]}

This is an ongoing optimization session: after each proposal I will report the objective values of the points you proposed and ask you to propose again. Use the accumulating results to refine your search.

¤\tprompt{USER PROMPT}¤
Initial observations (4 total):
[
  {
    "x": [
      0.512345,
      0.876190
    ],
    "y": -1.834521
  },
  {
    "x": [
      0.113092,
      0.240187
    ],
    "y": -12.407882
  },
  {
    "x": [
      0.755130,
      0.098720
    ],
    "y": -0.774301
  },
  {
    "x": [
      0.334170,
      0.655240
    ],
    "y": -3.219145
  }
]

Current best objective value: -0.774301

Remaining evaluation batches: 120

Propose exactly 1 new candidate(s) to maximize the objective. Each candidate must have 2 values, each in [0, 1]. Respond with JSON only.
\end{lstlisting}


\captionof{lstlisting}{Molecular generation (PMO): a representative rendered prompt for the multi-turn generative opening turn (batch size $q=10$). As in the synthetic task, the objective is never named---the model sees only the measured (SMILES, value) history and is asked to invent new molecules that maximize the (unnamed) oracle. The system prompt sets the de novo generation task (no candidate pool), the validity/no-repeat requirements, and the JSON response format, and the appended clause establishes the stateful session. The user prompt lists the measured molecules as tab-separated SMILES/value rows (highest first), the current best, and the remaining batch budget. The seed history is illustrative and abbreviated.}
\label{lst:prompt_molgen}
\begin{lstlisting}[nolol, escapechar=¤]
¤\tprompt{SYSTEM PROMPT}¤
You are a molecular design assistant. Your task is to design 10 new molecules (given as SMILES) to measure next. There is no fixed pool ¤\textemdash{}¤ you invent the molecules.

Your job is to maximize the objective function.

Every molecule must be a valid, chemically reasonable structure, written as a SMILES string RDKit can parse. Do not repeat a molecule you have already proposed; each must be new.

Respond with a JSON object containing a single key "selections" whose value is a list of exactly 10 valid SMILES strings.

Example response format:
{"selections": ["CCO", "c1ccccc1"]}

This is an ongoing design session: after each batch I will report the measured values of the molecules you proposed. Use the accumulating results to refine your designs toward higher values.

¤\tprompt{USER PROMPT}¤
Measured molecules so far (10 total), SMILES<tab>value:
Cc1ccc(NC(=O)c2ccc(Cl)cc2)cc1	0.418732
COc1ccc(CN2CCN(C)CC2)cc1	0.392015
O=C(Nc1ccccc1)c1ccncc1	0.361498
CC(C)Cc1ccc(C(C)C(=O)O)cc1	0.287340
c1ccc2[nH]ccc2c1	0.213067
¤\ldots¤ [5 additional seed molecules omitted for brevity]

Current best measured value: 0.418732

Remaining measurement batches: 20

Respond with JSON only: generate exactly 10 new, valid molecule(s) as SMILES to maximize the objective.
\end{lstlisting}

\clearpage

\end{document}